\documentclass{article}
\usepackage{arxiv}
\usepackage[utf8]{inputenc}
\usepackage[T1]{fontenc}
\usepackage{url}
\usepackage[draft]{fixme}
\fxsetup{inline=true,margin=false}
\fxsetup{theme=color}
\definecolor{fxtarget}{rgb}{0.75,0.25,0.0}
\usepackage{graphicx}
\usepackage{xcolor}
\usepackage{algorithm}% http://ctan.org/pkg/algorithms
\usepackage{algpseudocode}
\usepackage{amsmath}
\usepackage{amsfonts}
\usepackage{multirow}
\usepackage{tabularx}
\usepackage{tcolorbox}
\usepackage{colortbl}
\usepackage{longtable}

\usepackage{dsfont}
\def\RSet{\mathds{R}}

\definecolor{myred}{RGB}{201,55,30}
\definecolor{mygreen}{RGB}{9,188,138}
\definecolor{myorange}{RGB}{255,147,79}

\definecolor{dimcolor}{HTML}{909090}
\def\dimmed{\color{dimcolor}}

\usepackage{bm}

\def\ONE#1{\mathds{1}\!\left\{#1\right\}}
\def\PAR#1{#1}

\def\TopBag{x}
\def\TopBagLabel{y}
\def\SubBag#1{x_{#1}}
\def\SubBagLabel#1{y_{#1}}
\def\Instance#1#2{x_{{#1},{#2}}}
\def\InstanceLabel#1#2{y_{{#1},{#2}}}
\def\InstanceSpace{\mathcal{X}}
\def\SetOfTopBagLabels{\mathcal{Y}}
\def\SetOfInstanceLabels{\mathcal{Y}^{\mathrm{inst}}}
\def\SetOfSubBagLabels{\mathcal{Y}^{\mathrm{sub}}}
\def\ApproxSetOfInstanceLabels{\mathcal{C}^{\mathrm{inst}}}
\def\ApproxSetOfSubBagLabels{\mathcal{C}^{\mathrm{sub}}}
\def\CardinalityApproxInstanceLabels{k^{\mathrm{inst}}}
\def\CardinalityApproxSubBagLabels{k^{\mathrm{sub}}}
\def\MMILNetwork{F}
\def\CardinalitySubBag#1{n_{#1}}
\def\CardinalityTopBag{n}
\def\SubBagWeights{w^{\mathrm{sub}}}
\def\SubBagBias{b^{\mathrm{sub}}}
\def\InstanceWeights{w^{\mathrm{inst}}}
\def\InstanceBias{b^{\mathrm{inst}}}
\def\InstanceReprSet{I}
\def\SubBagReprSet{S}
\def\SurrogateIntanceLabelPredictor{r}
\def\DecisionTreeInstToSub{s}
\def\DecisionTreeSubToTop{t}
\usepackage[mathscr]{euscript}
\let\mathscr\relax% just so we can load this and rsfs
\usepackage[scr]{rsfso}

\def\powermultiset#1{\mathscr{M}(#1)}

\usepackage{mathtools}
\DeclareMathOperator*{\Agg}{\Xi}

\usepackage{xcolor}
\usepackage{booktabs}

\usepackage{amsthm}

\theoremstyle{definition}
\newtheorem{example}{Example}[section]
\theoremstyle{definition}

\begin{document}
\title{Learning and Interpreting Multi-Multi-Instance Learning Networks}
\author{
  Alessandro Tibo\\
  Aalborg University, Institut for Datalogi\\
  \texttt{alessandro@cs.aau.dk}\\
  \And
  Manfred Jaeger \\
  Aalborg University, Institut for Datalogi\\
  \texttt{jaeger@cs.aau.dk}\\
  \And
  Paolo Frasconi \\
  DINFO, Universit\`{a} di Firenze\\
  \texttt{paolo.frasconi@unifi.it}
}
\maketitle

\begin{abstract}
We introduce an extension of the multi-instance learning problem where
examples are organized as nested bags of instances (e.g., a document
could be represented as a bag of sentences, which in turn are bags of
words).  This framework can be useful in various scenarios, such as
text and image classification, but also supervised learning over
graphs.  As a further advantage, multi-multi instance learning enables
a particular way of interpreting predictions and the decision
function.  Our approach is based on a special neural network layer,
called bag-layer, whose units aggregate bags of inputs of arbitrary
size.  We prove theoretically that the associated class of functions
contains all Boolean functions over sets of sets of instances and we
provide empirical evidence that functions of this kind can be actually
learned on semi-synthetic datasets.  We finally present experiments on
text classification, on citation graphs, and social graph data, which
show that our model obtains competitive results with respect to
accuracy when compared to other approaches such as convolutional networks on graphs,
while at the same time it supports a general approach to interpret the
learnt model, as well as explain individual predictions.
\end{abstract}


%%%%%%%%%%%%%%%%%%%%%%%%%%%%%%%%%%%%%%%%%%%%%%%%%%%%%%%%%%%%%%%%%%%%%%
% Introduction - newly Written from scratch
%%%%%%%%%%%%%%%%%%%%%%%%%%%%%%%%%%%%%%%%%%%%%%%%%%%%%%%%%%%%%%%%%%%%%

\section{Introduction}

Relational learning takes several different forms ranging from purely
symbolic (logical) representations, to a wide collection of
statistical
approaches~\citep{De-Raedt:2008:Towards-digesting-the-alphabet-soup}
based on tools such as probabilistic graphical
models~\citep{jaeger_relational_1997,%
2008:Probabilistic-inductive-logic,%
  Richardson:2006:Markov-logic-networks,%
  Getoor:2007:Introduction-to-statistical-relational}, kernel
machines~\citep{Landwehr:2010:Fast-learning-of-relational}, and neural
networks~\citep{Frasconi98,%
  scarselli_graph_2009,%
  niepert_learning_2016}.

Multi-instance learning (MIL) is perhaps the simplest form of
relational learning where data consists of labeled bags of instances.
Introduced in~\citep{dietterich_solving_1997}, MIL has attracted the
attention of several researchers during the last two decades and has
been successfully applied to problems such as image and scene
classification~\citep{maron1998multiple,%
  zha2008joint,%
  zhou_multi-instance_2012}, image annotation~\citep{yang2006region},
image retrieval~\citep{yang2000image,%
  rahmani2005localized}, Web mining~\citep{zhou2005multi}, text
categorization~\citep{zhou_multi-instance_2012} and diagnostic medical
imaging~\citep{hou2015efficient,yan2016multi}.  In classic MIL, labels
are binary and bags are positive iff they contain at least one
positive instance (existential semantics). For example, a visual scene
with animals could be labeled as positive iff it contains at least one
tiger. Various families of algorithms have been proposed for MIL,
including axis parallel rectangles~\citep{dietterich_solving_1997},
diverse density~\citep{maron_framework_1998}, nearest
neighbors~\citep{wang_solving_2000}, neural
networks~\citep{ramon_multi_2000}, and variants of support vector
machines~\citep{andrews_support_2002}.
Several other formulations of MIL are possible, see
e.g.~\citep{foulds_review_2010} and under the mildest assumptions MIL and
supervised learning on sets, i.e. the problem also formulated in previous
works such as~\citep{DBLP:conf/icml/KondorJ03,%
  DBLP:journals/corr/VinyalsBK15,zaheer2017deep}, essentially come together.

In this paper, we extend the MIL setting by considering examples
consisting of labeled nested bags of instances. Labels are observed
for top-level bags, while instances and lower level bags have
associated latent labels.
For example, a potential offside situation in a soccer match
can be represented by a bag of images showing the scene from different
camera perspectives. Each image, in turn, can be interpreted as a bag
of players with latent labels for their team membership and/or
position on the field.
  We call this setting multi-multi-instance
learning (MMIL), referring specifically to the case of
bags-of-bags\footnote{ the generalization to deeper levels of nesting
  is straightforward but not explicitly formalized in the paper for
  the sake of simplicity.}.  In our framework, we also relax the
classic MIL assumption of binary instance labels, allowing categorical
labels lying in a generic alphabet. This is important since MMIL with
binary labels under the existential semantics would reduce to classic
MIL after flattening the bag-of-bags.

Our solution to the MMIL problem is based on neural networks with a special
layer called \textit{bag-layer}~\citep{tibo2017network}, which fundamentally
relies on weight sharing like other neural network architectures such as
convolutional networks~\citep{lecun_backpropagation_1989}, graph convolutional
networks~\citep{kipfsemi-supervised2016,DBLP:conf/icml/GilmerSRVD17} and
essentially coincides with the invariant model used in
DeepSets~\citep{zaheer2017deep}. Unlike previous
neural network approaches to MIL learning~\citep{ramon_multi_2000},
where predicted instance labels are aggregated by (a soft version of)
the maximum operator, bag-layers aggregate internal representations of
instances (or bags of instances) and can be naturally intermixed with
other layers commonly used in deep learning. Bag-layers can be in fact
interpreted as a generalization of convolutional layers followed
by pooling, as commonly used in deep learning.

The MMIL framework can be immediately applied to solve problems where
examples are naturally described as bags-of-bags.  For example, a text
document can be described as a bag of sentences, where in turn each
sentence is a bag of words.  The range of possible applications of the
framework is however larger.  In fact, every structured data object
can be recursively decomposed into parts, a strategy that has been
widely applied in the context of graph kernels (see
e.g.,~\citep{haussler_convolution_1999,%
  gartner2004kernels,%
  passerini2006kernels,%
  shervashidze_efficient_2009,%
  costa_fast_2010,%
  orsini2015graph}). Hence, MMIL is also applicable to supervised
graph classification.
Experiments on bibliographical and social network
datasets confirm the practical viability of MMIL for these forms of
relational learning.

As a further advantage, multi-multi instance learning enables a
particular way of interpreting the models by reconstructing instance
and sub-bag latent variables. This allows to explain the prediction
for a particular data point, and to describe the structure of the
decision function in terms of symbolic rules. Suppose we could recover the
latent labels associated with instances or inner bags. These labels
would provide useful additional information about the data since we
could group instances (or inner bags) that share the same latent label
and attach some semantics to these groups by inspection. For example,
in the case of textual data, grouping words or sentences with the same
latent label effectively discovers \textit{topics} and the decision of
a MMIL text document classifier can be interpreted in terms of the
discovered topics. In practice, even if we cannot recover the true
latent labels, we may still cluster the patterns of hidden units activations
in the bag-layers and use the cluster indices as surrogates of the latent
labels.

This paper is an extended version of \citep{tibo2017network}, where the MMIL
problem was first introduced and solved with networks of bag layers. The main
extensions contained in this paper are a general strategy for interpreting
MMIL networks via clustering and logical rules, and a much extended range of
experiments on real-world data.  The paper is organized as follows.  In
Section~\ref{sec:framework} we formally introduce the MMIL setting.  In
Section~\ref{sec:model} we introduce bag layers and the resulting neural
network architecture for MMIL, and derive a theoretical expressivity
result. Section~\ref{sec:graph} relates MMIL to standard graph learning
problems.  Section~\ref{sec:interpreting} describes our approach to
interpreting MMIL networks by extracting logical rules from trained networks
of bag-layers.  In Section~\ref{sec:related} we discuss some related works. in
Section~\ref{sec:experiments} we report experimental results on five different
types of real-world datasets. Finally we draw some conclusions in
Section~\ref{sec:conclusions}.

%%%%%%%%%%%%%%%%%%%%%%%%%%%%%%%%%%%%%%%%%%%%%%%%%%%%%%%%%%%%%%%%%%%%%%
% Setting - newly Written from scratch
%%%%%%%%%%%%%%%%%%%%%%%%%%%%%%%%%%%%%%%%%%%%%%%%%%%%%%%%%%%%%%%%%%%%%

\section{Framework}%
\label{sec:framework}

\subsection{Traditional Multi-Instance Learning}%
\label{sec:MIL}
In the standard multi-instance learning (MIL) setting, data consists
of labeled bags of instances. In the following, $\InstanceSpace$ denotes
the \textit{instance space} (it can be any set), $\SetOfTopBagLabels$ the
\textit{bag label space} for the observed labels of example bags,
and $\SetOfInstanceLabels$ the \textit{instance label space} for the unobserved
(latent) instance labels.
For any set $A$, $\powermultiset{A}$ denotes
the set of all multisets of $A$. An example in MIL is a pair
$(\TopBag,\TopBagLabel)\in \powermultiset{\InstanceSpace}\times\SetOfTopBagLabels$, which
we interpret as the observed part of an instance-labeled
example $(\TopBag^{\mathrm{labeled}},\TopBagLabel)\in \powermultiset{\InstanceSpace\times\SetOfInstanceLabels}\times\SetOfTopBagLabels$.
$\TopBag =\{\SubBag{1},\dots,\SubBag{n}\}$ is thus a
multiset of instances, and $\TopBag^{\mathrm{labeled}}=\{( \SubBag{1},\SubBagLabel{1}),\dots,(\SubBag{n},\SubBagLabel{n})\}$ a multiset
of labeled instances.

Examples are drawn from a fixed and unknown distribution
$p(\TopBag^{\mathrm{labeled}},\TopBagLabel)$. Furthermore, it is typically assumed that the label of an
example is conditionally independent of the individual instances given
their labels, i.e.\
$p(\TopBagLabel |( \SubBag{1},\SubBagLabel{1}),\dots,(\SubBag{n},\SubBagLabel{n})) = p(\TopBagLabel|\SubBagLabel{1},\dots,\SubBagLabel{n})$
In
the classic setting, introduced in~\citep{dietterich_ensemble_2000} and
used in several subsequent works~\citep{maron_framework_1998,%
  wang_solving_2000,%
  andrews_support_2002}, the focus is on binary classification
($\SetOfInstanceLabels=\SetOfTopBagLabels=\{0,1\}$) and it is postulated that
$\TopBagLabel=\ONE{0<\sum_j \SubBagLabel{j}}$, (i.e., an example is positive iff at least one
of its instances is positive). More complex assumptions are possible
and thoroughly reviewed in~\citep{foulds_review_2010}.  Supervised
learning in this setting can be formulated in two ways: (1) learn a
function $\MMILNetwork:\powermultiset{\InstanceSpace}\mapsto\SetOfTopBagLabels$ that
classifies whole examples, or (2) learn a function
$f:\InstanceSpace\mapsto\SetOfInstanceLabels$ that classifies instances and then
use some aggregation function defined on the multiset of predicted
instance labels to obtain the example label.

\subsection{Multi-Multi-Instance Learning}%
\label{sec:MMIL}
In multi-multi-instance learning (MMIL), data consists of labeled
\textit{nested bags}
of instances. When the level of nesting is two, an example
is a labeled bag-of-bags
$(\TopBag,\TopBagLabel)\in\powermultiset{\powermultiset{\InstanceSpace}}\times\SetOfTopBagLabels$
drawn from a distribution
$p(\TopBag,\TopBagLabel)$. %, where $\powermultiset{A}$ denotes
% the set of all multisets of $A$.
Deeper levels of nesting, leading to
multi$^K$-instance learning are conceptually easy to introduce but we
avoid them in the paper to keep our notation simple.
We will also informally use the expression ``bag-of-bags'' to describe
structures with two or more levels of nesting.
In the MMIL setting, we call the
elements of $\powermultiset{\powermultiset{\InstanceSpace}}$ and
$\powermultiset{\InstanceSpace}$ \textit{top-bags} and \textit{sub-bags},
respectively.

Now postulating unobserved labels for both the instances and the sub-bags,
we interpret examples $(\TopBag,\TopBagLabel)$ as the observed part of fully labeled
data points $(\TopBag^{\mathrm{labeled}},\TopBagLabel)\in\powermultiset{\powermultiset{\InstanceSpace\times\SetOfInstanceLabels}\times\SetOfSubBagLabels}\times\SetOfTopBagLabels$,
where $\SetOfSubBagLabels$ is the space of sub-bag labels.
Fully labeled data points are drawn from a distribution $p(\TopBag^{\mathrm{labeled}},\TopBagLabel)$.

As in MIL, we make some conditional
independence assumptions. Specifically, we assume that instance and sub-bag labels
only depend on properties of the respective instances or sub-bags, and not on
other elements in the nested multiset structure $\TopBag^{\mathrm{labeled}}$
(thus excluding models for contagion or homophily, where, e.g., a specific label for an
instance could become more likely, if many other instances contained in the same
sub-bag also have that label). Furthermore, we assume that labels of sub-bags and
top-bags only depend on the labels of their constituent elements. Thus, for
$\SubBagLabel{j} \in \SetOfSubBagLabels$, and a bag of labeled instances
$S^{\mathrm{labeled}}=\{ (\Instance{j}{1},  \InstanceLabel{j}{1} ), \ldots,(\Instance{j}{\CardinalitySubBag{j}},  \InstanceLabel{j}{\CardinalitySubBag{j}} ) \}$
we have:
\begin{equation}
  \label{eq:cond-indep}
  p(\SubBagLabel{j}|S^{\mathrm{labeled}})=p(\SubBagLabel{j} |\InstanceLabel{j}{1},\dots\InstanceLabel{j}{\CardinalitySubBag{j}}).
\end{equation}
Similarly for the probability distribution of top-bag labels given the constituent
labeled sub-bags.

\begin{figure}[ht]
  \centering
  \includegraphics[width=.7\textwidth]{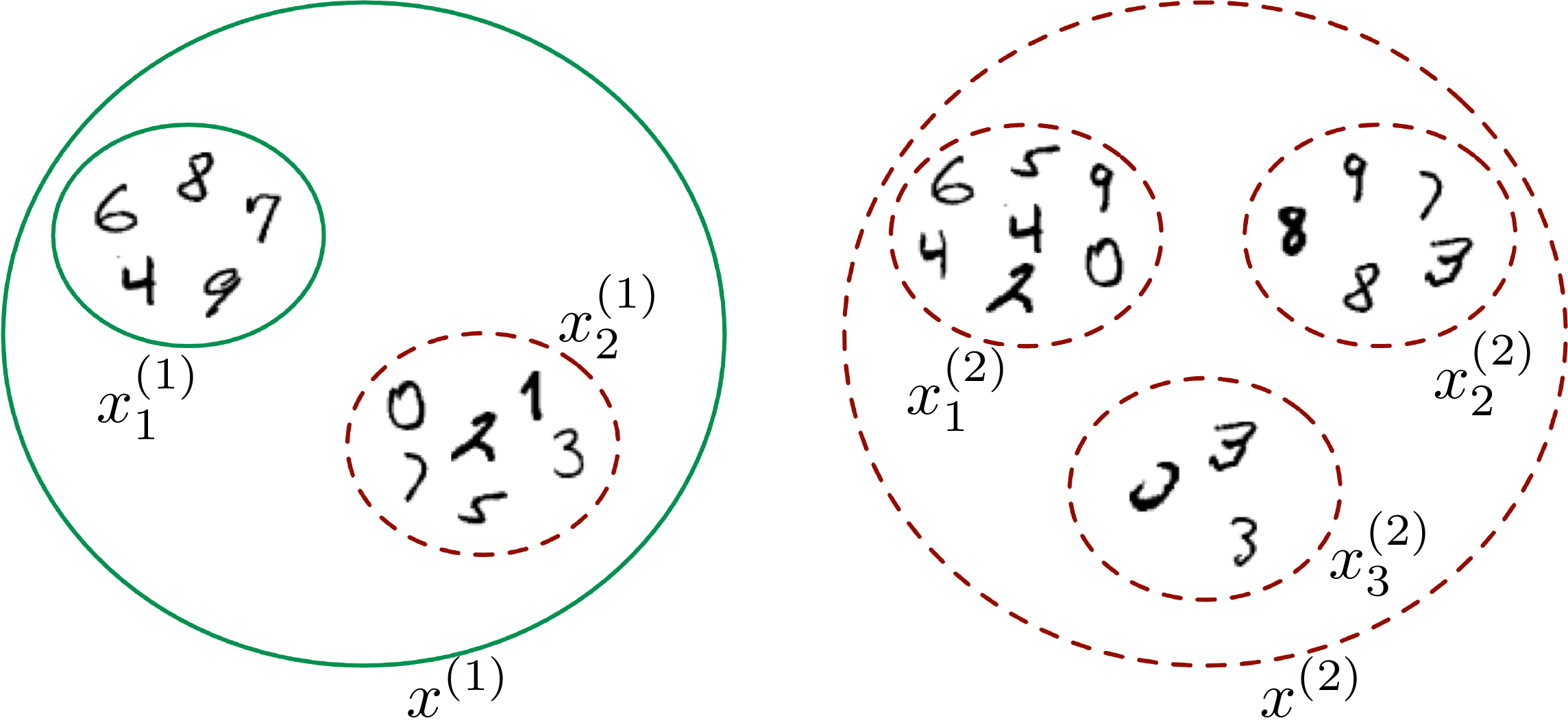}
  \caption{A positive (left) and a negative (right) top-bag for
    Example~\ref{ex:MNIST}. Solid green lines represent positive (sub-) bags
    while dashed red lines represent negative  (sub-) bags.}%
  \label{fig:example:mnist:topbags}
\end{figure}

\begin{example}%
  \label{ex:MNIST}
  In this example we consider bags-of-bags of handwritten digits (as
  in the MNIST dataset). Each instance (a digit) has attached its own
  latent class label in $\{0,\dots,9\}$ whereas sub-bag (latent) and
  top-bag labels (observed) are binary. In particular, a sub-bag is
  positive iff it contains an instance of class 7 and does not contain
  an instance of class 3. A top-bag is positive iff it contains at
  least one positive sub-bag. Figure~\ref{fig:example:mnist:topbags}
  shows a positive and a negative example.
\end{example}

\begin{example}%
  \label{ex:offside}
  A top-bag can consist of a set of images showing a potential offside situation in soccer
from different camera perspectives. The label of the bag corresponds to the referee decision
$\SetOfTopBagLabels\in\{\mbox{offside},\mbox{not offside}\}$.
Each individual image can either settle the offside question one way or another, or be
inconclusive. Thus, there are (latent) image labels
$\SetOfSubBagLabels\in\{\mbox{offside},\mbox{not offside},\mbox{inconclusive}\}$. Since no offside should
be called when in doubt, the top-bag is labeled as `not offside' if and only if it either contains
at least one image labeled `not offside', or all the images are labeled `inconclusive'.
Images, in turn, can be seen as bags of player instances that have a label
$\SetOfInstanceLabels\in\{\mbox{behind},\mbox{in front},\mbox{inconclusive}\}$ according to their
relative position with respect to the potentially offside player of the other team.
An image then is labeled `offside' if all the players in the image are labeled `behind'; it is
labeled `not offside' if it contains at least one player labeled `in front', and is labeled
`inconclusive' if it only contains players labeled `inconclusive' or `behind'.
\end{example}

\begin{example}%
  \label{ex:text}
  In text categorization, the bag-of-word representation is often used
  to feed documents to classifiers. Each instance in this case
  consists of the indicator vector of words in the document (or a
  weighted variant such as TF-IDF). The MIL approach has been applied
  in some cases~\citep{andrews_support_2002} where instances consist of chunks
  of consecutive words and each instance is an indicator vector. A
  bag-of-bags representation could instead describe a document as a
  bag of sentences, and each sentence as a bag of word vectors
  (constructed for example using Word2vec or GloVe).
\end{example}

%%%%%%%%%%%%%%%%%%%%%%%%%%%%%%%%%%%%%%%%%%%%%%%%%%%%%%%%%%%%%%%%%%%%%%
% Model - this part taken from Alessandro's last version of the
% journal paper
%%%%%%%%%%%%%%%%%%%%%%%%%%%%%%%%%%%%%%%%%%%%%%%%%%%%%%%%%%%%%%%%%%%%%
\subsection{A Network Architecture for MMIL}%
\begin{figure}[ht]
  \centering
  \includegraphics[width=.6\textwidth]{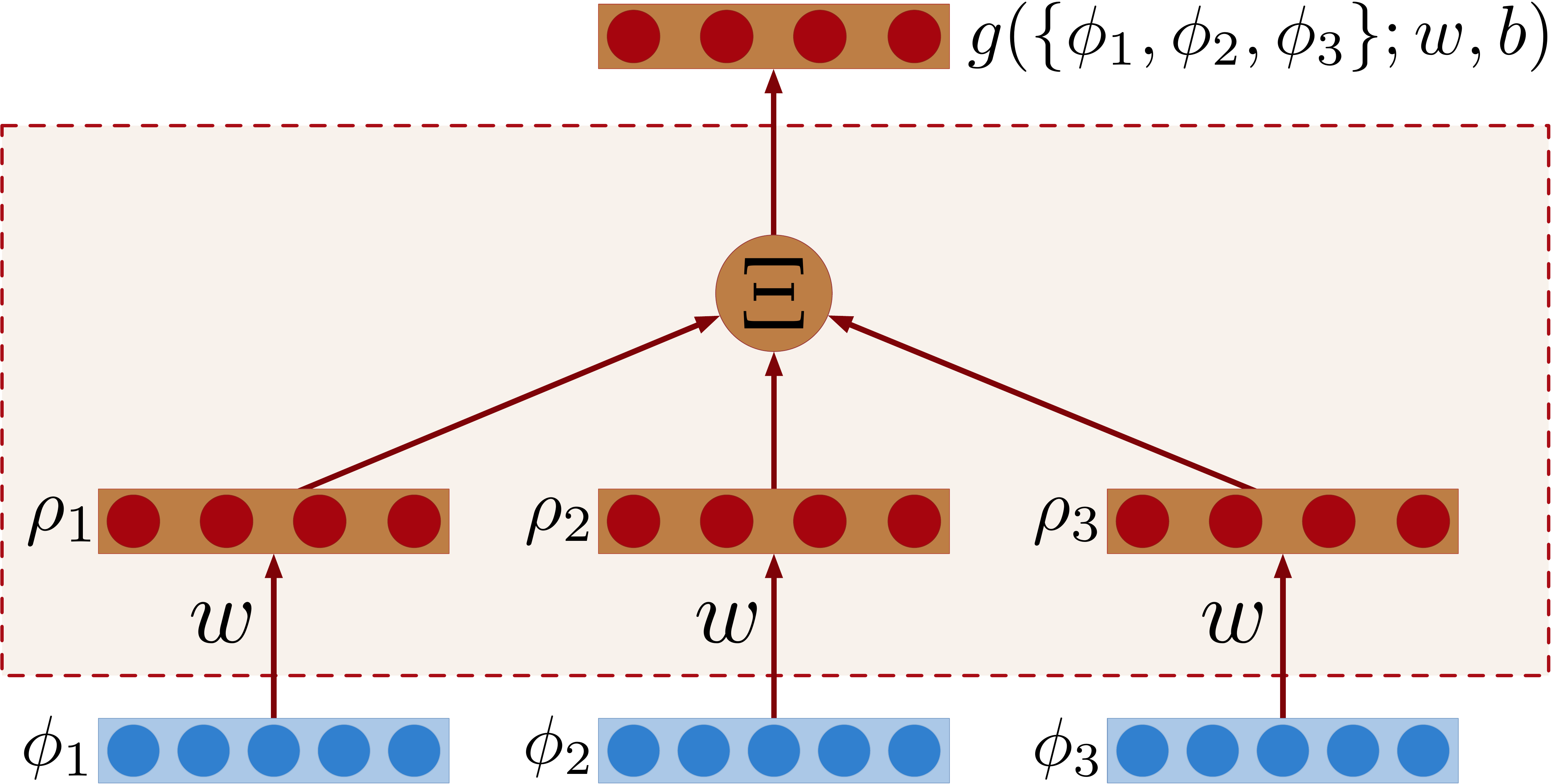}
  \caption{A bag-layer receiving a bag of cardinality $n=3$. In this example
  $k=4$ and $m=5$.}%
  \label{fig:bag-layer}
\end{figure}

\label{sec:model}
We model the conditional distribution $p(\TopBagLabel | \TopBag)$ with a neural network
architecture that handles bags-of-bags of variable sizes by
aggregating intermediate internal representations.  For this purpose,
we define a \textit{bag-layer} as follows:
\begin{itemize}
\item the input is a bag of $m$-dimensional vectors. $\{\phi_1,\dots,\phi_n\}$
\item First, $k$-dimensional representations are computed as
  \begin{equation}
    \label{eq:rhos}
    \rho_i= \alpha\left( w \phi_i + b \right)
  \end{equation}
  using a weight matrix $w\in\RSet^{k \times m}$, a bias vector $b\in\RSet^k$
  (both tunable parameters), and an activation function $\alpha$ (such as
  ReLU, tanh, or linear).
\item The output is
  \begin{equation}
    \label{eq:max_units}
    g(\{\phi_1,\dots,\phi_n\};w,b) = \Agg_{i=1}^{n} \rho_i
  \end{equation}
  where $\Agg$ is element-wise aggregation operator (such as max or
  average). Both $w$ and $b$
  are tunable parameters.
\end{itemize}
Note that Equation~\ref{eq:max_units} works
with bags of arbitrary cardinality. A bag-layer is illustrated in
Figure~\ref{fig:bag-layer}.

\begin{figure}[ht]
  \centering
  \includegraphics[width=1.\textwidth]{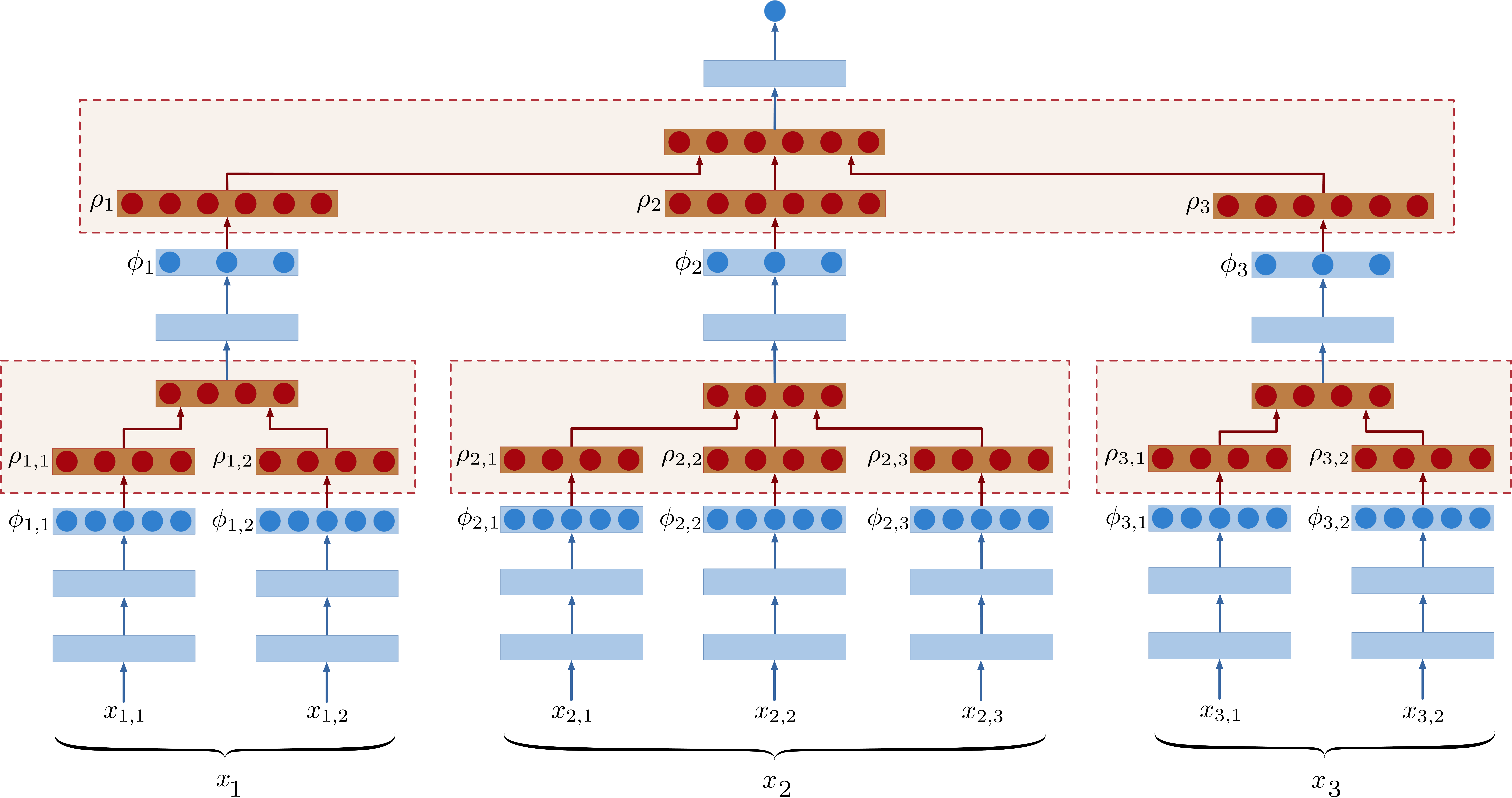}
  \caption{Network for multi-multi instance learning applied to the
    bag-of-bags
    $\{ \{x_{1,1},x_{1,2}\}, \{x_{2,1},x_{2,2},x_{2,2}\},
    \{x_{3,1},x_{3,2}\} \}$. Bag-layers are depicted in red with
    dashed borders. Blue boxes are standard (e.g., dense) neural
    network layers. Parameters in each of the
    seven bottom vertical columns are shared, and so are the parameters
    in the middle three columns.}
  \label{fig:Architecture}
\end{figure}

Networks with a single bag-layer can process bags of instances (as in
the standard MIL setting). To solve the MMIL problem, two bag-layers
are required. The bottom bag-layer aggregates over internal
representations of instances; the top bag-layer aggregates over
internal representations of sub-bags, yielding a representation for
the entire top-bag. In this case, the representation of each sub-bag
$\SubBag{j}=\{\Instance{j}{1},\dots,\Instance{j}{\CardinalitySubBag{j}}\}$ would be obtained as
\begin{equation}
  \label{eq:sub-bags-representation}
  \phi_j = g(\{\Instance{j}{1},\dots,\Instance{j}{\CardinalitySubBag{j}}\};\InstanceWeights, \InstanceBias) \ \ j=1,\dots,\CardinalityTopBag
\end{equation}
and the representation of a top-bag $\TopBag=\{\SubBag{1},\dots,\SubBag{\CardinalityTopBag}\}$ would be
obtained as
\begin{equation}
  \label{eq:top-bags-representation}
  \phi = g(\{\phi_{1},\dots,\phi_{\CardinalityTopBag}\};\SubBagWeights, \SubBagBias)
\end{equation}
where $(\InstanceWeights, \InstanceBias)$ and $(\SubBagWeights, \SubBagBias)$ denote the parameters used to
construct sub-bag and top-bag representations.
Multiple bag-layers with different aggregation functions can be also be used in parallel,
and  bag-layers can be intermixed with
standard neural network layers, thereby forming networks of arbitrary
depth.
An illustration of a possible  overall architecture involving two bag-layers is shown in
Figure~\ref{fig:Architecture}.

It is shown in~\citep{tibo2017network} that networks with two bag layers with max aggregation
can solve all MMIL problems that satisfy the restrictions if being \emph{deterministic} (essentially
saying that the conditional probability distributions (\ref{eq:cond-indep}) become deterministic
functions) and \emph{non-counting} (the multiplicities of elements in the bags do not matter).

\section{MMIL for Graph Learning}
\label{sec:graph}
The MMIL perspective can also be used to derive algorithms suitable
for supervised learning over graphs, i.e., tasks such as graph
classification, node classification, and edge prediction. In all these
cases, one first needs to construct a representation for the object of
interest (a whole graph, a node, a pair of nodes) and then apply a
classifier. A suitable representation can be obtained in our framework
by first forming a bag-of-bags associated with the object of interest
(a graph, a node, or an edge) and then feeding it to a network with
bag-layers. In order to construct bags-of-bags, we follow the classic
$R$-decomposition strategy introduced
by~\cite{haussler_convolution_1999}. In the present context, it simply
requires us to introduce a relation $R(A,a)$ which holds true if $a$
is a ``part'' of $A$ and to form $R^{-1}(A) = \{a:R(A,a)\}$, the bag
of all parts of $A$. Parts can in turn be decomposed in a similar
fashion, yielding bags-of-bags. In the following, we focus on
undirected graphs $G=(V,E)$ where $V$ is the set of nodes and
$E = \{\{u,v\}:u,v\in V\}$ is the set of edges. We also assume that a
labeling function $\xi:V\mapsto\mathcal{X}$ attaches attributes to
vertices. Variants with directed graphs or labeled edges are
straightforward and omitted here in the interest of brevity.

\paragraph{Graph classification.}
A simple solution is to define the part-of relation $R(G,g)$ between
graphs to hold true iff $g$ is a subgraph of $G$ and to introduce a
second part-of relation $S(g,v)$ that holds true iff $v$ is a node in
$g$. The bag-of-bags associated with $G$ is then constructed as
$x = \{  \{\xi(v): v\in S^{-1}(g)   \} : g\in R^{-1}(G)\}$.
In general, considering all
subgraphs is not practical but suitable feasible choices for $R$ can
be derived borrowing approaches already introduced in the graph kernel
literature, for example decomposing $G$ into cycles and
trees~\citep{horvath_cyclic_2004}, or into neighbors or neighbor
pairs~\citep{costa_fast_2010} (some of these choices may require three
levels of bag nesting, e.g., for grouping cycles and trees
separately).

\paragraph{Node classification.}
In some domains, the node labeling function itself is bag-valued. For
example in a citation network, $\xi(v)$ could be the bag of words in the
abstract of the paper associated with node $v$. A bag-of-bags in this
case may be formed by considering a paper $v$ together all papers in
its neighborhood $N(v)$ (i.e., its cites and citations):
$x^{(v)}=\{\xi(u), u\in\{v\}\cup N(v)\}$. A slightly more rich description
with three layers of nesting could be used to set apart a node and its
neighborhood: $x^{(v)} = \{\{\xi(v)\}, \{\xi(u), u\in N(v)\} \}$.

\section{Interpreting Networks of Bag-Layers}
\label{sec:interpreting}
Interpreting the predictions in the supervised learning setting
amounts to provide a human understandable explanation of the
prediction. Transparent techniques such as rules or trees retain much
of the symbolic structure of the data and are well suited in this
respect. On the contrary, predictions produced by methods based on
numerical representations are often opaque, i.e., difficult to explain
to humans. In particular, representations in neural networks are
highly distributed, making it hard to disentangle a clear semantic
interpretation of any specific hidden unit. Although many works exist
that attempt to interpret neural networks, they mostly focus on
specific application domains such as
vision~\citep{lapuschkin2016lrp,samek2016interpreting}.

The MMIL settings offers
some advantages in this respect. Indeed, if instance or sub-bag labels
were observed, they would provide more information about bag-of-bags
than mere predictions.
To clarify our vision, MIL
approaches like mi-SVM and MI-SVM in~\citep{andrews_support_2002} are
not equally interpretable: the former is \textit{more} interpretable
than the latter since it also provides individual instance labels
rather than simply providing a prediction about the whole bag. These
standard MIL approaches make two assumptions: first all labels are
binary, second the relationship between the instance labels and the
bag label is predefined to be the existential quantifier. The MMIL model
relaxes these assumptions by allowing labels in an a-priori unknown categorical
alphabet, and by allowing more complex mappings between bags of
instance labels and sub-bag labels.
We follow the standard MIL approaches in that our interpretation
approach is also based on the assumption of
a deterministic mapping from component to bag labels, i.e.,
0,1-valued probabilities in (\ref{eq:cond-indep}).

The idea we propose in the following consists of two major components. First,
given MMIL data and a MMIL network, we infer label sets
$\ApproxSetOfInstanceLabels, \ApproxSetOfSubBagLabels$,
labeling functions for instances and sub-bags, and  sets of rules for the
mapping from instance to sub-bag labels, and sub-bag to top-bag labels.
This component is purely algorithmic and described in Section~\ref{sec:learnrules}.
Second, in order to support interpretation, semantic explanations of the
constructed labels and inferred rules are provided. This component is hightly
domain and data-depend. Several general solution strategies are described in
Section~\ref{sec:explainrules}.

\subsection{Learning Symbolic Rules}
\label{sec:learnrules}

For ease of exposition, we first describe the construction of synthetic labels and
the learning of classification rules as two separate procedures. In the final algorithm
these two procedures are interleaved (cf. Algorithms~\ref{alg:interpretability},~\ref{alg:fidelity}~\ref{alg:best:interpretability}).

\paragraph{Synthetic Label Construction.}
We construct sets $\ApproxSetOfInstanceLabels, \ApproxSetOfSubBagLabels$ as clusters of
internal instance and sub-bag representations. Let $\MMILNetwork$ be a
MMIL network  trained on labeled top-bag data $\{(\TopBag^{(i)},\TopBagLabel^{(i)}), i=1,\dots,m\}$.
Let $\CardinalityApproxInstanceLabels, \CardinalityApproxSubBagLabels$ be target cardinalities for $\ApproxSetOfInstanceLabels$ and $\ApproxSetOfSubBagLabels$,
respectively.

The inputs $\{\TopBag^{(i)}, i=1,\dots,m\}$ generate  multi-sets of sub-bag and instance
representations computed by the bag layers of $F$:
\begin{equation}\label{eq:sub:repr}
\SubBagReprSet = \{\rho^{(i)}_j\mid i=1,\dots,m, j=1,\dots, \CardinalityTopBag^{(i)}\}.
\end{equation}

\begin{equation}\label{eq:inst:repr}
\InstanceReprSet  = \{\rho^{(i)}_{j,\ell}\mid i=1,\dots,m, j=1,\dots,\CardinalityTopBag^{(i)}, \ell=1,\dots,\CardinalitySubBag{j}^{(i)}\}
\end{equation}
where the $\rho^{(i)}_j$ and $\rho^{(i)}_{j,\ell}$ are the representations
according to (\ref{eq:rhos}) (cf. Figure~\ref{fig:Architecture}).  We  cluster (separately)
the sets $\InstanceReprSet,\SubBagReprSet$, setting the target number of clusters to $\CardinalityApproxInstanceLabels$ and $\CardinalityApproxSubBagLabels$, respectively.
Each resulting cluster is associated with a synthetic cluster identifier $u_i$, respectively
$v_i$, so that $\ApproxSetOfInstanceLabels:=\{u_1,\ldots,u_{\CardinalityApproxInstanceLabels}\}$ and $\ApproxSetOfSubBagLabels:=\{v_1,\ldots,v_{\CardinalityApproxSubBagLabels}\}$.
Any instance $\Instance{j}{\ell}$ and sub-bag $\SubBag{j}$ in an example $\TopBag$ (either one of the training examples
$\TopBag^{(i)}$, or a new test example) is then associated with the identifier of the cluster whose centroid
is closest to the representation $\rho_{j,\ell}$, respectively $\rho_j$ computed by $\MMILNetwork$ on $\TopBag$.
We denote the resulting labeling with
cluster identifiers by $\InstanceLabel{j}{\ell}^{(i)}\in \ApproxSetOfInstanceLabels$ and
$\SubBagLabel{j}^{(i)}\in \ApproxSetOfSubBagLabels $.

We use K-means clustering as the underlying clustering method. While other clustering methods could be considered,
it is important that the clustering method also provides a function that maps
new examples to one of the constructed clusters.

\paragraph{Learning  rules.}
We next describe how we construct symbolic rules  that
approximate the actual (potentially noisy) relationships between cluster identifiers
in the MMIL network.

Let us denote a bag of cluster identifiers  as
$\{\SubBagLabel{\ell}:c_{\ell} \mid \ell=1,\dots, |\mathcal{Y}|\}$, where $c_{\ell}$ is the
multiplicity of $\SubBagLabel{\ell}$. An attribute-value representation of the
bag can be immediately obtained in the form of a frequency vector
$(f_{c_1},\ldots,f_{c_{|\mathcal{Y}|}} )$, where
$f_{c_{\ell}}=c_{\ell}/\sum_{p=1}^{|\mathcal{Y}|} c_p$ is the frequency of
identifier $\ell$ in the bag.
Alternatively, we can also use a $0/1$-valued occurrence
vector $(o_{c_1},\ldots,o_{c_{|\mathcal{Y}|}} )$
with $o_{c_l}=\ONE{c_l>0}$.
Jointly with the example label $y$, this attribute-value
representation provides a supervised example that is now described at
a compact symbolic level. Examples of this kind are well suited for
transparent and interpretable classifiers that can naturally operate
at the symbolic level. Any rule-based learner could be applied here
and in the following we will use decision trees because of their
simplicity and low bias.

In the two level MMIL case, we learn in this way functions $\DecisionTreeInstToSub,\DecisionTreeSubToTop$
mapping multisets of instance cluster identifiers to sub-bag cluster identifiers, and
multisets of sub-bag cluster identifiers to top-bag labels, respectively.
In the second case, our target labels are the predicted labels of the
original MMIL network, not the actual labels of the training examples.
Thus, we aim to construct rules that best approximate the MMIL model, not rules that
provide the highest accuracy themselves.

Let $\SurrogateIntanceLabelPredictor(\Instance{j}{\ell}):=\InstanceLabel{j}{\ell}$ be the instance labeling function defined for all
$\Instance{j}{\ell}\in \InstanceSpace$. Together with the learned functions $\DecisionTreeInstToSub,\DecisionTreeSubToTop$ we obtain
a complete classification model for
a top-bag based on the input features of its instances: $\hat{F}(\TopBag)\doteq   \DecisionTreeSubToTop(\DecisionTreeInstToSub(\SurrogateIntanceLabelPredictor(x) ))$. We refer to the accuracy of
this model with regard to the predictions of the original MMIL model as its
\emph{fidelity}, defined as
$$
\mathrm{Fidelity} = \frac{1}{|\mathcal{D}|}\sum_{(\TopBag,\TopBagLabel)\in \mathcal{D}} \ONE{\MMILNetwork(\TopBag)=\hat{F}(\TopBag)}.
$$
We use fidelity on a validation set as the criterion to select the
cardinalities for $\ApproxSetOfSubBagLabels$ and $\ApproxSetOfInstanceLabels$ by performing a grid
search over $\CardinalityApproxSubBagLabels,\CardinalityApproxInstanceLabels$ value combinations. In Algorithms~\ref{alg:interpretability},~\ref{alg:fidelity}, ~\ref{alg:best:interpretability} we reported the pseudo-codes for learning the best symbolic rules for a MMIL network $\MMILNetwork$. In particular Algorithm~\ref{alg:interpretability} computes an \emph{explainer}, an object consisting of cluster centroids and a decision tree, for interpreting a single level of $\MMILNetwork$. Algorithm~\ref{alg:fidelity} calculates the fidelity score, and Algorithm~\ref{alg:best:interpretability} searches the best explainer for $\MMILNetwork$. For the sake of simplicity we condensed the pseudo-codes by exploiting  the following subroutines:
\begin{itemize}
  \item \textsc{Flatten($S$)} is a function which takes as input a set of multi-sets and return a set containing all the elements of each multi-set of $S$;
  \item \textsc{KMeans($T$, $k$)} is the KMeans algorithm which takes as input a set of vectors $T$ and $k$ clusters and returns the $k$ $centroids$;
  \item \textsc{Assign-Labels($S$, $centroids$)} is a function that takes as input a set of multi-sets $S$ and returns a set of multi-sets $labels$.  $labels$ has the same structure of $S$ and each instance is replaced by its cluster index with respect to the $centroids$;
  \item \textsc{Frequencies($labels$)} is a function which takes as input a set of multi-sets $S$ of cluster index and returns for each multi-set a vectors containing the frequencies of each cluster index within the muti-set;
  \item \textsc{Intermediate-Representations}($F$, $X$) takes as input a MMIL network $F$ and a set of top-bags $X$ and return the multi-sets of intermediate representations for  sub-bags and instances as described in Equations  \ref{eq:sub:repr} and \ref{eq:inst:repr}, respectively.
\end{itemize}

\subsection{Explaining  Rules and Predictions}
\label{sec:explainrules}

The functions $\SurrogateIntanceLabelPredictor$, $\DecisionTreeInstToSub$, $\DecisionTreeSubToTop$ provide a complete symbolic approximation $\hat{F}$ of the
given MMIL model $\MMILNetwork$. Being expressed in terms of a (small number of) categorical symbols and
simple classification rules, this approximation is more amenable to human interpretation than
the original $\MMILNetwork$. However, the interpretability of $\hat{F}$ sill hinges on the interpretability
of its constituents, notably the semantic interpretability of the cluster identifiers. There is no
general algorithmic solution to provide such semantic interpretations, but a range of possible
(standard) strategies that we briefly mention here, and whose use in our particular context is
illustrated in our experiments:
\begin{itemize}
\item Direct visualization: in the case
where the instances forming a cluster are given by points in low-dimensional
Euclidean space, whole clusters can be plotted directly. Examples of this case will be found
in Sections~\ref{ex:point:clouds} and \ref{sec:plants}.
\item Cluster representatives: often clusters are described in terms of a small number of
  most representative elements.  For example, in the case of textual data,
  this was suggested in the area of topic
  modelling~\citep{blei_latent_2003,griffiths_2004_finding}. We follow a similar approach in
  Section~\ref{sec:exp:imdb}.
\item Ground truth labels:   in some cases the cluster elements may be equipped with some true, latent
label. In such cases we can alternatively characterize clusters in terms
of their association with these actual labels. An example of this can
be found in Section~\ref{sec:semi-mnist}.
\end{itemize}

$\hat{F}$ in conjunction with an interpretation of the cluster identifiers constitutes a global
(approximate) explanation of the model $\MMILNetwork$. This global explanation leads to example-specific
explanations of individual predictions by tracing for an example $\TopBag$ the rules in $\DecisionTreeInstToSub$, $\DecisionTreeSubToTop$ that
were were used to determine $\hat{F}(\TopBag)$, and by identifying  the critical substructures of $\TopBag$
(instances, sub-bags) that activated these rules (cf.
the classic multi-instance setting, where a positive classification will be triggered by a
single positive instance).

%%%%%%%%%%%%%%%%%%%%%%%%%%%%%%%%%%%%%%%%%%%%%%%%%%%%%%%%%%%%%%%
% PSEUDOCODE
%%%%%%%%%%%%%%%%%%%%%%%%%%%%%%%%%%%%%%%%%%%%%%%%%%%%%%%%%%%%%%%

\begin{algorithm}[ht]
  \caption{Explain a bag-layer for a MMIL network}\label{alg:interpretability}
%    \hspace*{\algorithmicindent} \textbf{Input}:
%  \hspace*{\algorithmicindent} \textbf{Output}: Rules
  \begin{algorithmic}[1]
    \Require{$S$ set of multi-sets of representations computed by the bag layer, with
  corresponding labels $Y$;  $k$ number of
  desired clusters.}
  \Ensure{an object explainer $e$ which consists of two attributes: cluster $centroids$ and decision tree $f$.}
    \Procedure{Build-Explainer}{$S$, $Y$, $k$}
    \State $e.centroids=$ \textsc{KMeans}(\textsc{Flatten}($S$), $k$)
    \State $labels=$ \textsc{Assign-Labels}($S$, $e.centroids$)
    \State $F =$ \textsc{Frequencies}($labels$)
    \State $e.f=$ \textsc{Decision-Tree}($F$, $Y$)
    \State \textbf{return} $e$
    \EndProcedure
  \end{algorithmic}
\end{algorithm}

\begin{algorithm}[ht]
  \caption{Compute the fidelity between an explainer and a MMIL network}\label{alg:fidelity}
  \begin{algorithmic}[1]
    \Require{$e^{\mathrm{inst}}$, $e^{\mathrm{sub}}$ explainers for instances and sub-bags; $F$ MMIL network; set of top-bags $X$. }
  \Ensure{the fidelity $fid$.}
    \Procedure{Fidelity}{$e^{\mathrm{inst}}$, $e^{\mathrm{sub}}$, $F$, $X$}
	\State $\InstanceReprSet$, $\SubBagReprSet$ =  \textsc{Intermediate-Representations}($F$, $X$)    
	\State $r = $ \textsc{Assign-Labels}($I$, $e^{\mathrm{inst}}.centroids$)
	\State $s,\ t = e^{\mathrm{inst}}.f, \ e^{\mathrm{sub}}.f$
	\State $\hat{F} = t(\mbox{\textsc{Frequency}}(s( \mbox{\textsc{Frequency}}(r))))$
	\State $fid = \frac{1}{|X|} \sum_{i=1}^{|X|} \ONE{F(X_i) = \hat{F} _i}$
    \State \textbf{return} $fid$
    \EndProcedure
  \end{algorithmic}
\end{algorithm}

\begin{algorithm}[ht]
  \caption{Best Explainer for a MMIL network}\label{alg:best:interpretability}
  \begin{algorithmic}[1]
    \Require{$F$ MMIL network; $X_\mathrm{train}$, $X_\mathrm{valid}$ training and validation sets of top-bags ; $k_\mathrm{max}$ maximum number of clusters.}
  \Ensure{best explainer for $F$.}
    \Procedure{Find-Best-Explainer}{$\MMILNetwork$, $X_\mathrm{train}$, $X_\mathrm{valid}$, $k_{\mathrm{max}}$}
    \State $E = \emptyset $

    \State  $\InstanceReprSet_\mathrm{train}$, $\SubBagReprSet_\mathrm{train}=$ \textsc{Intermediate-Representations}($F, X_\mathrm{train}$)

    \For{$\CardinalityApproxSubBagLabels=2$ \textbf{to} $k_{\mathrm{max}}$}
          \State $e^\mathrm{sub}  = $ \textsc{Build-Explainer}($\SubBagReprSet_\mathrm{train}$,  $\MMILNetwork(X_\mathrm{train})$, $\CardinalityApproxSubBagLabels$)
          \For{$\CardinalityApproxInstanceLabels=2$ \textbf{to} $k_{\mathrm{max}}$}
          \State $c = \mbox{\textsc{Assign-Labels}}(\SubBagReprSet_\mathrm{train}, e^{\mathrm{sub}}.centroids)$
            \State $e^\mathrm{inst}  = $ \textsc{Build-Explainer}($\InstanceReprSet_\mathrm{train}$,  $c$, $\CardinalityApproxInstanceLabels$)
            \State $E = E \cup \left\{ (e^\mathrm{sub},e^\mathrm{inst}) \right\} $
      \EndFor
    \EndFor

    \State \textbf{return} $\displaystyle \arg \max_{(e^{\mathrm{inst}},e^{\mathrm{sub}})\in E}
    \textsc{Fidelity}(e^{\mathrm{inst}},e^{\mathrm{sub}},F,X_\mathrm{valid})$
    \EndProcedure
  \end{algorithmic}
\end{algorithm}

%%%%%%%%%%%%%%%%%%%%%%%%%%%%%%%%%%%%%%%%%%%%%%%%%%%%%%%%%%%%%%%%%%%%%%
% Related works - Taken from the ECML version for now
%%%%%%%%%%%%%%%%%%%%%%%%%%%%%%%%%%%%%%%%%%%%%%%%%%%%%%%%%%%%%%%%%%%%%
\section{Related Works}
\label{sec:related}

\subsection{Invariances, Symmetries, DeepSets}
\label{sec:deepsets}

Understanding invariances in neural networks is a foundational issue that has
attracted the attention of researchers since \citep{Minsky_1988} with results
for multilayered neural networks going back to
\citep{Shawe-Taylor-89}. Sum-aggregation of representations constructed via
weight sharing has been applied for example in~\citep{Lusci_2013} where
molecules are described as sets of breadth-first trees constructed from every
vertex. \cite{zaheer2017deep} proved that a function
operating on sets over a countable universe can always be expressed as a function
of the sum of a suitable representation of the
set elements. Based on this result they introduced the DeepSets learning architecture. 
The aggregation of representations exploited in the bag-layer defined
in Section~\ref{sec:model} has been used in the invariant model version of
DeepSets~\citep{zaheer2017deep} (in the case of examples described by bags,
i.e. in the MIL setting), and in the preliminary version of this
paper~\citep{tibo2017network} (in the case of examples described by bags of
bags).

\subsection{Multi-Instance Neural Networks}
\cite{ramon_multi_2000} proposed a neural network
solution to MIL where each instance $x_j$ in a bag
$\TopBag=\{\SubBag{1},\dots,\SubBag{\CardinalitySubBag{j}}\}$ is first processed by a replica of a neural
network $f$ with weights $w$. In this way, a bag of output values
$\{f(\SubBag{1};w),\dots,f(\SubBag{\CardinalitySubBag{j}};w)\}$ computed for each bag of
instances. These values are then aggregated by a smooth version of the
max function:
$$
F(\TopBag) = \frac{1}{M} \log\left( \sum_j e^{Mf(\SubBag{j};w)} \right)
$$
where $M$ is a constant controlling the sharpness of the aggregation (the
exact maximum is computed when $M\to\infty$).  A single bag-layer (or a
DeepSets model) can used to solve the MIL problem. Still, a major difference
compared to the work of~\citep{ramon_multi_2000} is the aggregation is
performed at the \textit{representation} level rather than at the output
level. In this way, more layers can be added on the top of the aggregated
representation, allowing for more expressiveness. In the classic MIL setting
(where a bag is positive iff at least one instance is positive) this
additional expressiveness is not required. However, it allows us to solve
slightly more complicated MIL problems. For example, suppose each instance has
a latent variable $y_j\in{0,1,2}$, and suppose that a bag is positive iff it
contains at least one instance with label $0$ and no instance with label
$2$. In this case, a bag-layer with two units can distinguish positive and
negative bags, provided that instance representations can separate instances
belonging to the classes $0,1$ and $2$. In this case, the network proposed
in~\citep{ramon_multi_2000} would not be able to separate positive from
negative bags.

\subsection{Convolutional Neural Networks}
\label{sec:conv}
Convolutional neural networks
(CNN)~\citep{fukushima_neocognitron:_1980,lecun_backpropagation_1989}
are the state-of-the-art method for image classification (see,
e.g.,~\citep{szegedy_inception-v4_2016}). It is easy to see that the
representation computed by one convolutional layer followed by
max-pooling can be emulated with one bag-layer by just creating bags
of adjacent image patches.  The representation size $k$ corresponds to
the number of convolutional filters. The major difference is that a
convolutional layer outputs spatially ordered vectors of size $k$,
whereas a bag-layer outputs a set of vectors (without any
ordering). This difference may become significant when two or more
layers are sequentially stacked.
\begin{figure}[ht]
  \centering \includegraphics[width=\textwidth]{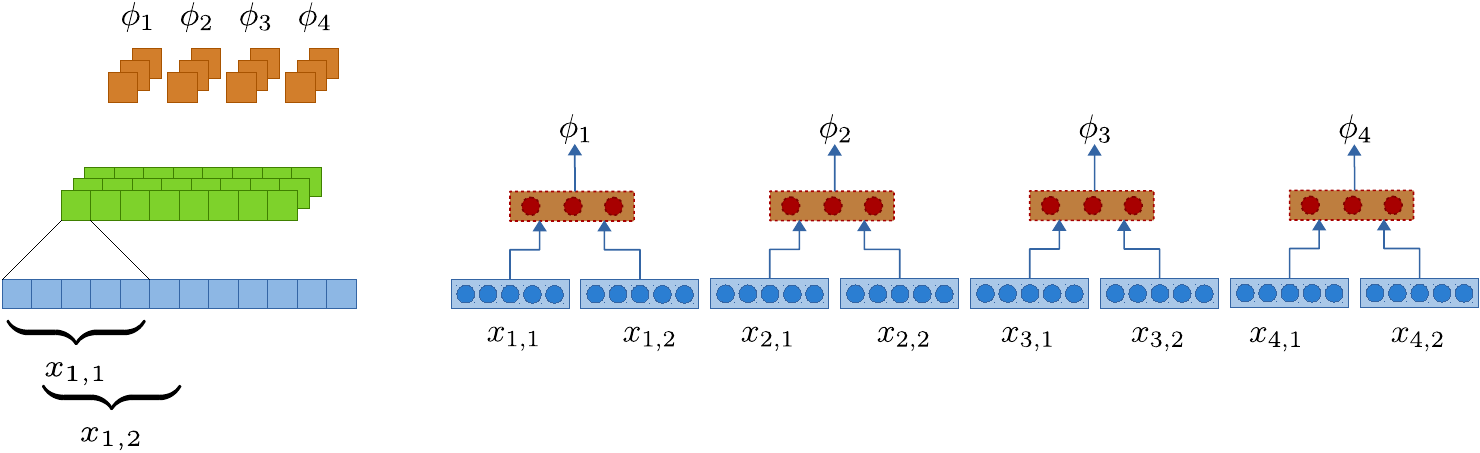}
  \caption{One convolutional layer with subsampling (left) and the
    corresponding bag-layer (right). Note that the convolutional layer
    outputs $[\phi_1,\phi_2, \phi_3,\phi_4]$ whereas the bag-layer
    outputs $\{\phi_1,\phi_2, \phi_3,\phi_4\}$.}
  \label{fig:conv1d-as-bag-layer}
\end{figure}
Figure~\ref{fig:conv1d-as-bag-layer} illustrates the relationship
between a convolutional layer and a bag-layer, for simplicity assuming
a one-dimensional signal (i.e., a sequence). When applied to signals,
a bag-layer essentially correspond to a disordered convolutional layer
and its output needs further aggregation before it can be fed into a
classifier. The simplest option would be to stack one additional
bag-layer before the classification layer. Interestingly, a network of
this kind would be able to detect the presence of a short subsequence
regardless of its position within the whole sequence, achieving
invariance to arbitrarily large translations

We finally note that it is possible to emulate a CNN with two layers
by properly defining the structure of bags-of-bags. For example, a
second layer with filter size 3 on the top of the CNN shown in
Figure~\ref{fig:conv1d-as-bag-layer} could be emulated with two
bag-layers fed by the bag-of-bags
$$
\{
\{ \{x_{1,1},x_{1,2}\},\{x_{2,1},x_{2,2}\},\{x_{3,1},x_{3,2}\} \},
\{ \{x_{2,1},x_{2,2}\},\{x_{3,1},x_{3,2}\},\{x_{4,1},x_{4,2}\} \}
\}.
$$
A bag-layer, however, is not limited to pooling adjacent elements in a
feature map. One could for example segment the image first (e.g.,
using a hierarchical strategy~\citep{arbelaez_contour_2011}) and then
create bags-of-bags by following the segmented regions.

\subsection{Graph Convolutional Networks}
\label{sec:gcn}

The convolutional approach has been also recently employed for learning with
graph data. The idea is to reinterpret the convolution operator as a message
passing algorithm on a graph where each node is a signal sample (e.g., a
pixel) and edges connect a sample to all samples covered by the filter when
centered around its position (including a self-loop). In a general graph
neighborhoods are arbitrary and several rounds of propagation can be carried
out, each refining representations similarly to layer composition in CNNs.
This message passing strategy over graphs was originally proposed
in~\citep{gori2005new,scarselli_graph_2009} and has been reused with variants
in several later works. A general perspective of several such algorithms is
presented in~\citep{DBLP:conf/icml/GilmerSRVD17}.  In this respect, when our
MMIL setting is applied to graph learning (see Section~\ref{sec:graph}),
message passing is very constrained and only occurs from instances to subbags
and from subbags to the topbag.

When extending convolutions from signals to graphs, a major difference is that
no obvious ordering can be defined on neighbors.
\cite{kipfsemi-supervised2016} for example, propose to address the ordering
issue by sharing the same weights for each neighbor (keeping them distinct
from the self-loop weight), which is the same form of sharing exploited in a
bag-layer (or in a DeepSet layer). They show that their message-passing is
closely related to the 1-dimensional Weisfeiler-Lehman (WL) method for
isomorphism testing (one convolutional layer corresponding to one iteration of
the WL-test) and can be also motivated in terms of spectral convolutions on
graphs. On a side note, similar message-passing strategies were also used
before in the context of graph
kernels~\citep{shervashidze_weisfeiler-lehman_2011,neumann_efficient_2012}.

Several other variants exist.  \cite{niepert_learning_2016} proposed ordering
via a ``normalization'' procedure that extends the classic canonicalization
problem in graph isomorphism. \cite{hamilton2017inductive} propose an
extension of the approach in~\citep{kipfsemi-supervised2016} with generic
aggregators and a neighbor sampling strategy, which is useful for large
networks of nodes with highly variable degree. Additional related works
include~\citep{duvenaud_convolutional_2015}, where CNNs are applied to
molecular fingerprint vectors, and~\citep{atwood_diffusion-convolutional_2016}
where a diffusion process across general graph structures generalizes the CNN
strategy of scanning a regular grid of pixels.

A separate aspect of this family of architectures for graph data concerns the
function used to aggregate messages arriving from
neighbors. GCN~\citep{kipfsemi-supervised2016} rely on a simple
sum. GraphSAGE~\citep{hamilton2017inductive}, besides performing a
neighborhood sampling, aggregates messages using a general differentiable
function that can be as simple as the sum or average, the maximum, or as
complex as a recurrent neural network, which however requires messages to be
linearly ordered. An even more sophisticated strategy is employed in graph
attention networks (GAT)~\citep{DBLP:conf/iclr/VelickovicCCRLB18} where each
message receives a weight computed as a tunable function of the other
messages. In this respect, the aggregator in our formulation in
Eq.~(\ref{eq:max_units}) is typically instantiated as the maximum (as in one
version of GraphSAGE) or the sum (as in GCNs) and could be modified to
incorporate attention. \cite{tibo2017network} showed that the maximum
aggregator is sufficient if labels do not depend on instance counts.

\begin{figure}[ht]
  \centering \includegraphics[width=\textwidth]{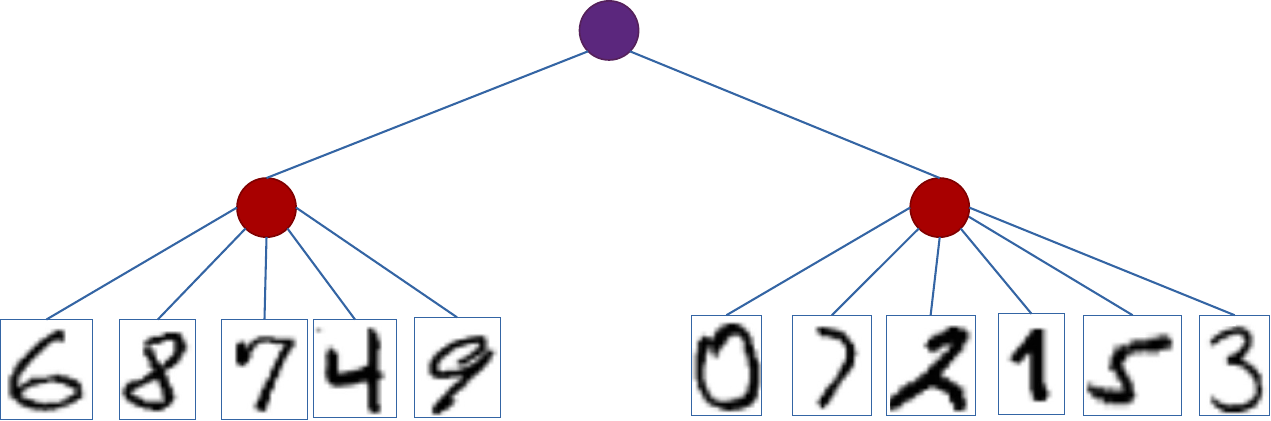}
  \caption{Mapping a bag-of-bags into an MMI tree.}
  \label{fig:MMI-tree}
\end{figure}

To gain more intuition about the similarities and differences between GCNs and
our approach, observe that a MMIL problem could be mapped to a graph
classification problem by representing each bag-of-bags as an MMI tree whose
leaves are instances, internal (empty) nodes are subbags, and whose root is
associated with the topbag. This is illustrated in
Figure~\ref{fig:MMI-tree}. The resulting MMI trees could be given as input to
any graph learning algorithm, including GCNs. For example when using
the~\citep{kipfsemi-supervised2016} GCN, in order to ensure an equivalent
computation, the self-loop weights should be set to zero and the message
passing protocol should be modified to prevent propagating information
``downwards'' in the tree (otherwise information from one subbag would leak
into the representation of other subbags).
Note, however, that in the scenario of Section~\ref{sec:graph} (where the MMIL
problem is derived from a graph learning problem) the above reduction would
produce a rather different graph learning problem instead of recovering the
original one. Interestingly, we shown in Section~\ref{sec:social:data} that
using a MMIL formulation can outperform many types of neural networks for
graphs on the original node classification problem.

\subsection{Nested SRL Models}
\label{sub:sec:rw:srl}

In Statistical Relational Learning (SRL) a great number of approaches
have been proposed for constructing probabilistic models for
relational data.  Relational data has an inherent bag-of-bag
structure: each object $o$ in a relational domain can be interpreted
as a bag whose elements are all the other objects linked to $o$ via a
specific relation. These linked objects, in turn, also are bags
containing the objects linked via some relation.  A key component of
SRL models are the tools employed for aggregating (or combining)
information from the bag of linked objects.  In many types of SRL
models, such an aggregation only is defined for a single
level. However, a few proposals have included models for nested
combination~\citep{jaeger_relational_1997,natarajan_learning_2008}. Like
most SRL approaches, these models employ concepts from first-order
predicate logic for syntax and semantics,
and~\citep{jaeger_relational_1997} contains an expressivity result
similar in spirit to the one reported in~\cite{tibo2017network} for MMIL.

A key difference between SRL models with nested combination constructs
and our MMIL network models is that the former build models based on
rules for conditional dependencies which are expressed in first-order
logic and typically only contain a very small number of numerical
parameters (such as a single parameter quantifying a noisy-or
combination function for modelling multiple causal influences).  MMI
network models, in contrast, make use of the high-dimensional
parameter spaces of (deep) neural network architectures.  Roughly
speaking, MMIL network models combine the flexibility of SRL models to
recursively aggregate over sets of arbitrary cardinalities with the
power derived from high-dimensional parameterisations of neural
networks.

\subsection{Interpretable Models}
\label{sec:related:interpret}
Recently, the question of interpretability
has become particularly prominent in  image processing and the
neural network context in general
\citep{uijlings2012visual,hentschel2015image,bach2015pixel,lapuschkin2016lrp,samek2016interpreting}.
In all of these works, the  predictions of a classifier $f$
are explained for each instance
$x\in\RSet^n$, by attributing
scores to each entry of $x$. A positive
$R_i > 0$ or negative $R_i< 0$ score is then assigned
to $x_i$, depending whether $x_i$ contributes
for predicting the target or not. In the case where input instances $x$ are
images,  the relevance scores are usually illustrated in the form of heatmaps over the
images.

\cite{ribeiro2016should} also provided explanations for individual
predictions as a solution to the ``trusting a prediction'' problem
by approximating a machine learning model
with an interpretable model. The authors assumed that
instances are given in a representation which
is understandable to humans,
regardless of the actual features used by the
model. For example for text classification an interpretable
representation may be the binary vector indicating the presence
or absence of a word. An ``interpretable'' model is defined as a model that
can be readily presented to the user with visual
or textual artefacts (linear models,
decision trees, or falling rule lists), which
locally approximates the original machine learning model.

A number of interpretation approaches have been described for classification models that use
a transformation of the raw input
data (e.g. images) to a bag of (visual) word representation by some form of vector
quantization~\citep{uijlings2012visual,hentschel2015image,bach2015pixel}.
Our construction of synthetic labels via clustering of internal representations also is
a form of vector quantization, and  we also learn classification models using
bags of cluster identifiers as features. However, our approach described in Section~\ref{sec:interpreting}
differs from previous
work in  fundamental aspects: first, in previous work, bag of words representations
were used in the actual classifier, whereas in our approach only the interpretable
approximation $\hat{F}$ uses the bag of identifiers representation. Second, the
cluster identifiers and their interpretability are a core component of our explanations, both
at the model level, and the level of individual predictions. In previous work, the
categorical (visual) words were not used for the purpose of explanations, which at the end
always are given as
a relevance map over the original input features.

The most fundamental differences between all those previous methods  and our
interpretation framework, however,
is that with the latter we are able to provide a global explanation
for the whole MMIL network, and not only to explain predictions for individual examples.

%%%%%%%%%%%%%%%%%%%%%%%%%%%%%%%%%%%%%%%%%%%%%%%%%%%%%%%%%%%%%%%%%%%%%%
% Experiments - Taken from Alessandro's version of journal paper
% for now. Needs to be thoroughly checked
%%%%%%%%%%%%%%%%%%%%%%%%%%%%%%%%%%%%%%%%%%%%%%%%%%%%%%%%%%%%%%%%%%%%%

\section{Experimental Results}
\label{sec:experiments}

We performed experiments in the MMIL setting in several different problems,
summarized below:
\begin{description}
\item[Pseudo-synthetic data] derived from MNIST as in Example~\ref{ex:MNIST},
  with the goal of illustrating the interpretation of models trained in the
  MMIL setting in a straightforward domain.
\item[Sentiment analysis] The goal is to compare models trained in the MIL and
  in the MMIL settings in terms of accuracy and interpretability on textual data.
\item[Graphs data] We report experiments on standard citation datasets (node
  classification) and social networks (graph classification), with the goal of
  comparing our approach against several neural networks for graphs.
\item[Point clouds] A problem where data is originally described in terms of
  bags and where the MMIL setting can be applied by describing objects as bags
  of point clouds with random rotations, with the goal of comparing MIL
  (DeepSets) against MMIL.
\item[Plant Species] A novel dataset of geo-localized plant species in
  Germany, with the goal of comparing our MMIL approach against more
  traditional techniques like Gaussian processes and matrix factorization.
\end{description}

\subsection{A Semi-Synthetic Dataset}\label{sec:semi-mnist}

The problem is described in Example~\ref{ex:MNIST}. We formed
a balanced training set of 5,000 top-bags using MNIST digits. Both
sub-bag and top-bag cardinalities were uniformly sampled in
$[2,6]$. Instances were sampled with replacement from the MNIST
training set (60,000 digits). A test set of 5,000 top-bags was
similarly constructed but instances were sampled from the MNIST test
set (10,000 digits).  Details on the network architecture and the
training procedure are reported in Appendix~\ref{apx:exp:mnist} in Table~\ref{tab:mnist:structure}.
We stress the fact that instance and
sub-bag labels were not used for training. The learned network achieved
an accuracy on the test
set of $98.42\%$, confirming that the network is able to recover the
latent logic function that was used in the data generation process
with a high accuracy.

We show next how the general approach of
Section~\ref{sec:interpreting} for constructing interpretable rules
recovers the latent labels and logical rules used in the data
generating process.  Interpretable rules are learnt with the
procedure described in Section~\ref{sec:interpreting}. Clustering was
performed with K-Means using the Euclidean distance%in Eq.~(\ref{eq:score})
. Decision trees were used as propositional learners.  As
described in Section~\ref{sec:interpreting}, we
determined the number of clusters at the instance and at the sub-bag level by maximizing the fidelity
of the interpretable model on the validation data via grid search,
and in this way found $\CardinalityApproxInstanceLabels=6$, and  $\CardinalityApproxSubBagLabels=2$, respectively. Full
results of the grid search are depicted as a heat-map in Appendix~\ref{apx:exp:mnist}
(Figure~\ref{fig:mnist:faith}).

\begin{figure}[ht]
  \centering
  \includegraphics[width=0.99\textwidth]{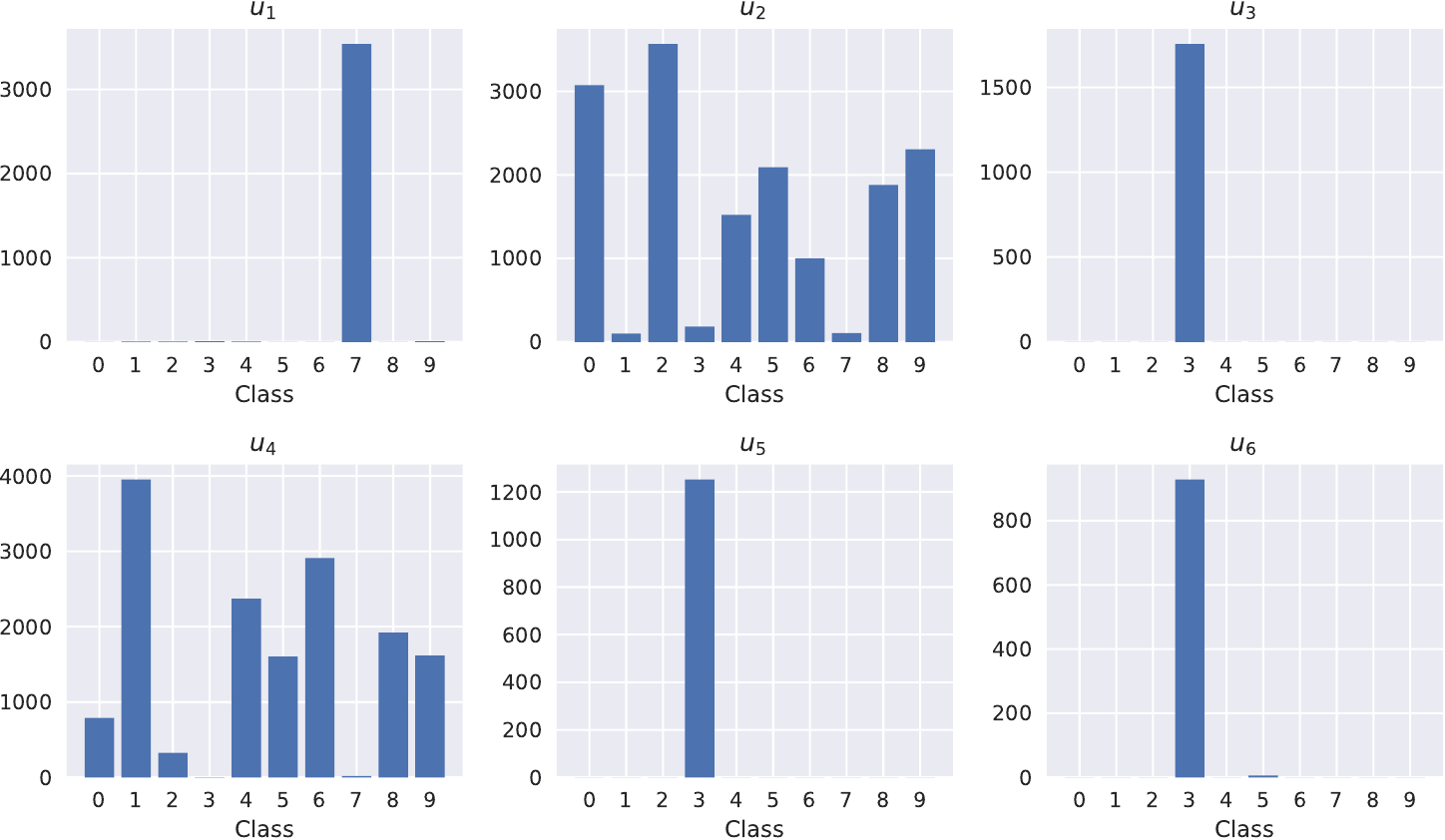}
  \caption{Correspondence between cluster identifiers $u_i$ and actual digit class labels}%
  \label{fig:mnist:acts}
\end{figure}

We can interpret the instance clusters by analysing their correspondence with
the actual digit labels. It is then immediate to recognize that
cluster $u_1$ corresponds to the digit 7, $u_3$, $u_5$, and $u_6$
all correspond to digit 3, and $u_2$ and $u_4$ correspond to digits other than 7 and 3.
All correspondences are shown by histograms in Figure~\ref{fig:mnist:acts}.
From a decision tree trained to predict cluster identifiers of sub-bags $\SubBag{j}$ from instance-level occurrence vectors  $(o_{u_1},\ldots,o_{u_6})$
we then extract the following rules defining the function $\DecisionTreeInstToSub$:

\begin{equation}
  \label{eq:mnist1}
  \begin{array}{ll}
    1&\DecisionTreeInstToSub=v_1 \leftarrow \PAR{o_{u_1}\mathord{=}1}, \PAR{o_{u_3}\mathord{=}0}, \PAR{o_{u_5}\mathord{=}0}, \PAR{o_{u_6}\mathord{=}0}.\\
    2&\DecisionTreeInstToSub=v_2 \leftarrow \PAR{o_{u_1}\mathord{=}0}.\\
    3&\DecisionTreeInstToSub=v_2 \leftarrow \PAR{o_{u_3}\mathord{=}1}.\\
    4&\DecisionTreeInstToSub=v_2 \leftarrow \PAR{o_{u_5}\mathord{=}1}.\\
    5&\DecisionTreeInstToSub=v_2 \leftarrow \PAR{o_{u_6}\mathord{=}1}.\\
  \end{array}
\end{equation}
Based on the already established interpretation of the instance clusters
$u_1,u_3,u_5,u_6$ we thus find that the sub-bag cluster $v_1$ gets attached
to the sub-bags that contain a seven and not a three, i.e., it corresponds to the
latent 'positive' label for sub-bags.

Similarly, we extracted the following rule that predict the
class label of a top-bag $\TopBag$ based on the sub-bag occurrence vector
$(o_{v_1},o_{v_2})$.
\begin{equation}
  \label{eq:mnist2}
  \begin{array}{ll}
    1 & \DecisionTreeSubToTop=\mbox{positive} \leftarrow \PAR{o_{v_1}\mathord{=}1}\\
    2 & \DecisionTreeSubToTop=\mbox{negative} \leftarrow \PAR{o_{v_1}\mathord{=}0}\\
  \end{array}
\end{equation}
Hence, in this example, the true rules behind the data generation
process were perfectly recovered. Note that perfect recovery does not
necessarily imply perfect accuracy of the resulting rule-based
classification model $\SurrogateIntanceLabelPredictor,\DecisionTreeInstToSub, \DecisionTreeSubToTop$, since the initial
instance clusters $\SurrogateIntanceLabelPredictor(\Instance{j}{\ell})$ do not correspond to digit
labels with 100\% accuracy.  Nonetheless, in this experiment the classification accuracy
of the interpretable rule model on the test set was $98.18\%$, only
$0.24\%$ less than the accuracy of the original model, which it approximated
with a fidelity of $99.16\%$.

\subsection{Sentiment Analysis}
\label{sec:exp:imdb}

In this section, we apply our approach to a real-world dataset for sentiment analysis.
The main objective of this experiment is to demonstrate the feasibility of our model
interpretation framework on real-world data, and to explore the trade-offs
between an MMIL and MIL approach.
We use the
IMDB~\citep{maas-EtAl:2011:ACL-HLT2011} dataset, which is a
standard benchmark movie review dataset for binary sentiment
classification. We remark that this IMDB dataset differs from
the IMDB graph datasets described in Section~\ref{sec:social:data}.
IMDB  consists of 25,000 training reviews, 25,000 test
reviews and 50,000 unlabeled reviews. Positive and negative labels
are balanced within the training and test sets.
Text data exhibits a natural bags-of-bags structure by viewing
a text as a bag of sentences, and each sentence as a bag of words.
Moreover, for the IMDB data it is reasonable to associate with each sentence a
(latent) sentiment label (positive/negative, or maybe something
more nuanced), and to assume that the overall sentiment of the review
is a (noisy) function of the sentiments of its sentences. Similarly, sentence
sentiments can be explained by latent sentiment labels of the words
it contains.

\begin{figure}[ht]
  \centering
  \includegraphics[width=0.99\textwidth]{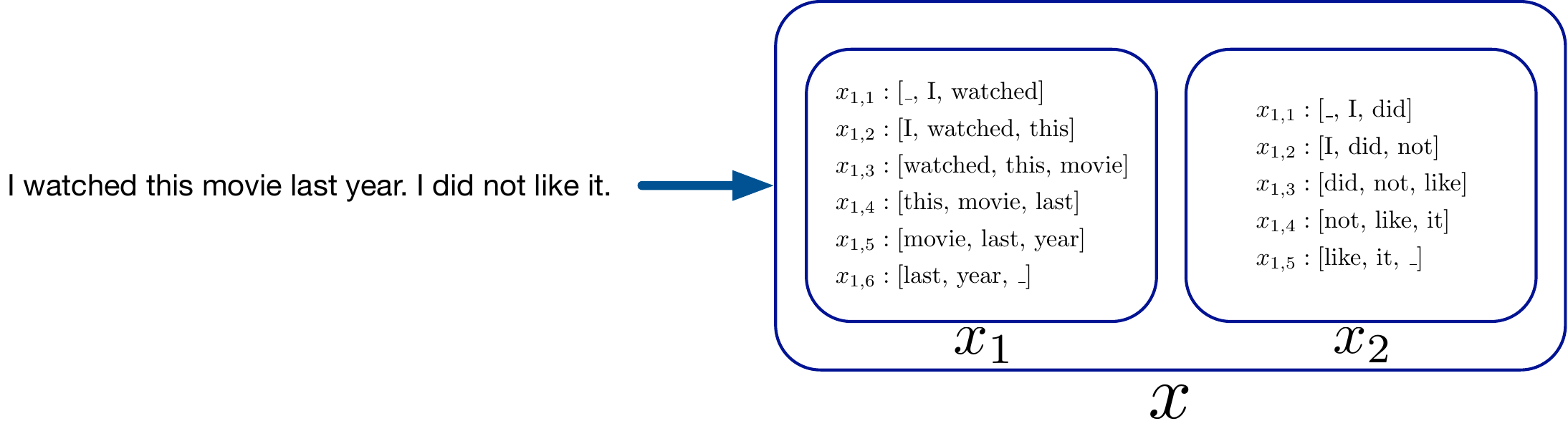}
  \caption{A review transformed into MMIL data. The word ``\_''
    represents the padding.}
  \label{fig:imdb:ex}
\end{figure}

A MMIL dataset was constructed from the reviews, where then each
review (top-bag) is a bag of sentences. However, instead of modeling
each sentence (sub-bag) as a bag of words, we represented sentences
as bags of trigrams in order to take into account possible negations, e.g.
``not very good'', ``not so bad''.
Figure \ref{fig:imdb:ex} depicts an example of the
decomposition of a two sentence review $\TopBag$ into MMIL data.
Each word is represented with Glove word
vectors~\citep{pennington2014glove} of size 100, trained on the
dataset. The concatenation of its three Glove word vectors then is the
feature vector we use to represent a trigram.
We here use Glove word vectors for a more pertinent  comparison of our model
with the state-of-the-art~\citep{miyato2016virtual}. Nothing prevents us
from using a one-hot representation even for this scenario.
In order to compare MMIL against multi-instance (MIL) we also
constructed a multi-instance dataset in which a review is simply
represented as a bag of trigrams.

We trained two neural networks for MMIL and MIL data respectively, which
have the following structure:
\begin{itemize}
\item \textbf{MMIL network}: a Conv1D layer with 300 filters (each trigram is treated separately), ReLU
  activations and kernel size of 100 (with stride 100),
  two stacked bag-layers (with ReLU activations) with
  500 units each (250 max-aggregation, 250 mean-aggregation)
  and an output layer with sigmoid activation;
\item \textbf{MIL network}: a Conv1D layer with 300 filters (each trigram is treated separately), ReLU
  activations and kernel size of 100 (with stride 100),
  one bag-layers (with ReLU activations) with
  500 units (250 max-aggregation, 250 mean-aggregation)
  and an output layer with sigmoid activation;
\end{itemize}

The models were trained by minimizing the binary cross-entropy loss.
We ran 20 epochs of the Adam optimizer with learning rate 0.001,
on mini-batches of size 128. We used also
virtual adversarial training~\citep{miyato2016virtual} for
regularizing the network and exploiting the unlabeled reviews during
the training phase.
Although our model does not outperform the state-of-the-art ($94.04\%$,
~\cite{miyato2016virtual}), we obtained a final accuracy of $92.18 \pm 0.04$ for the MMIL network and $91.41 \pm 0.08$ for the MIL network, by running the experiments 5 times for both the networks.
Those results
show that the MMIL representation here leads to a slightly higher accuracy
than the MIL representation.

When accuracy is not the only concern, our models have the advantage
that we can distill them into interpretable sets of rules following
our general strategy.  As in Section~\ref{sec:semi-mnist}, we constructed interpretable rules both in the MMIL and in the MIL setting.  Using
2,500 reviews as a validation set, we obtained  in the MMIL case 4 and 5 clusters
for sub-bags and instances, respectively, and in the MIL case  6
clusters for instances.  Full grid search results on the
validation set are reported in Appendix~\ref{apx:imdb}
(Figure~\ref{fig:imdb:faith}).

In this case we interpret clusters by representative elements. Using centroids or
points close to centroids as representatives here produced points (triplets, respectively
sentences) with relatively little interpretative value. We therefore focused on
inter-cluster separation rather than intra-cluster cohesion, and used the minimum distance
to any other cluster centroid as a cluster
representativeness score.

Tables~\ref{tab:imdb:top20sents} and~\ref{tab:imdb:top20words}
report the top-scoring sentences and trigrams, respectively, sorted by
decreasing score. It can be seen that sentences labeled by $v_1$ or $v_4$
express negative judgments, sentences labeled by $v_2$ are either
descriptive, neutral or ambiguous, while sentences labeled by $v_3$
express a positive judgment. Similarly, we see that trigrams labeled
by $u_1$ express positive judgments while trigrams labeled by $u_2$ or
$u_4$ express negative judgments. Columns printed in grey
correspond to clusters that do not actually appear
in the extracted rules (see below), and they do not generally
correspond to a clearly identifiable sentiment. Percentages in
parenthesis in the headers of these tables refer to fraction of
sentences or trigrams associated with each cluster (the total
number of sentences in the dataset is approximately 250 thousand
while the total number of trigrams is approximately 4.5 million).  A
similar analysis was performed in the MIL setting (results in
Table~\ref{tab:imdb:mitop20words}).

\begin{table}[htp]
\caption{Interpreting sentence (sub-bag) clusters in the MMIL
    setting.}
\label{tab:imdb:top20sents}
\vskip 0.15in
\begin{center}
\begin{small}
%\begin{sc}
\begin{tabularx}{\textwidth}{XXXX}
\toprule
\multicolumn{1}{c}{$v_1$ $(11.37\%)$} & \multicolumn{1}{c}{$v_2$ $(41.32\%)$} & \multicolumn{1}{c}{$v_3$ $(15.80\%)$} & \multicolumn{1}{c}{$v_4$ $(31.51\%)$}  \\
\midrule
    overrated poorly written badly acted & {\leavevmode\dimmed I highly recommend you to NOT waste your time on this movie as I have} &  I loved this movie and I give it an 8/ 10 & It's not a total waste \\
    \midrule
    It is badly written badly directed badly scored badly filmed & {\leavevmode\dimmed This movie is poorly done but that is what makes it great} & Overall I give this movie an 8/ 10 & horrible god awful \\
    \midrule
    This movie was poorly acted poorly filmed poorly written and overall horribly executed & {\leavevmode\dimmed Although most reviews say that it isn't that bad i think that if you are a true disney fan you shouldn't waste your time with...} & final rating for These Girls is an 8/ 10 & Awful awful awful \\
    \midrule
    Poorly acted poorly written and poorly directed & {\leavevmode\dimmed  I've always liked Madsen and his character was a bit predictable but this movie was definitely a waste of time both to watch and make...} &  overall because of all these factors this film deserves an 8/ 10 and stands as my favourite of all the batman films & junk forget it don't waste your time etc etc \\
    \midrule
    This was poorly written poorly acted and just overall boring & {\leavevmode\dimmed  If you want me to be sincere The Slumber Party Massacre Part 1 is the best one and all the others are a waste of...} & for me Cold Mountain is an 8/ 10 & Just plain god awful \\
\bottomrule
\end{tabularx}
%\end{sc}
\end{small}
\end{center}
\vskip -0.1in
\end{table}

\begin{table}[htp]
\caption{Interpreting trigram (instance) clusters in the MMIL
    setting.}
\label{tab:imdb:top20words}
\vskip 0.15in
\begin{center}
\begin{small}
%\begin{sc}
\begin{tabular}{lllll}
\toprule
\multicolumn{1}{c}{$u_1$ $(5.73\%)$} & \multicolumn{1}{c}{$u_2$ $(8.68\%)$} & \multicolumn{1}{c}{$u_3$ $(28.86\%)$} & \multicolumn{1}{c}{$u_4$ $(2.82\%)$} & \multicolumn{1}{c}{$u_5$ $(53.91\%)$} \\
\midrule
\_ 8/ 10 & trash 2 out & {\leavevmode\dimmed   had read online} & it's pretty poorly & {\leavevmode\dimmed  give this a} \\
an 8/ 10 & to 2 out & {\leavevmode\dimmed   had read user} & save this poorly & {\leavevmode\dimmed  like this a} \\
for 8/ 10 & \_ 2 out & {\leavevmode\dimmed   on IMDb reading} & for this poorly & {\leavevmode\dimmed  film is 7} \\
HBK 8/ 10 & a 2 out & {\leavevmode\dimmed   I've read innumerable} & just so poorly & {\leavevmode\dimmed  it an 11} \\
Score 8/ 10 & 3/5 2 out & {\leavevmode\dimmed   who read IMDb} & is so poorly & {\leavevmode\dimmed  the movie an} \\
to 8/ 10 & 2002 2 out & {\leavevmode\dimmed   to read IMDb} & were so poorly & {\leavevmode\dimmed  this movie an} \\
verdict 8/ 10 & garbage 2 out & {\leavevmode\dimmed   had read the} & was so poorly & {\leavevmode\dimmed  40 somethings an} \\
Obscura 8/ 10 & Cavern 2 out & {\leavevmode\dimmed   I've read the} & movie amazingly poorly & {\leavevmode\dimmed  of 5 8} \\
Rating 8/ 10 & Overall 2 out & {\leavevmode\dimmed   movie read the} & written poorly directed & {\leavevmode\dimmed  gave it a} \\
it 8/ 10 & rating 2 out & {\leavevmode\dimmed   Having read the} & was poorly directed & {\leavevmode\dimmed  give it a} \\
fans 8/ 10 & film 2 out & {\leavevmode\dimmed   to read the} & is very poorly & {\leavevmode\dimmed  rating it a} \\
Hero 8/ 10 & it 2 out & {\leavevmode\dimmed   I read the} & It's very poorly & {\leavevmode\dimmed  rated it a} \\
except 8/ 10 & score 2 out & {\leavevmode\dimmed   film reviews and} & was very poorly & {\leavevmode\dimmed  scored it a} \\
Tracks 8/ 10 & Grade 2 out & {\leavevmode\dimmed   will read scathing} & a very poorly & {\leavevmode\dimmed  giving it a} \\
vote 8/ 10 & Just 2 out & {\leavevmode\dimmed   \_ After reading} & very very poorly & {\leavevmode\dimmed  voting it a} \\
as 8/ 10 & as 2 out & {\leavevmode\dimmed   about 3 months} & Poorly acted poorly & {\leavevmode\dimmed  are reasons 1} \\
strong 8/ 10 & and 2 out & {\leavevmode\dimmed   didn't read the} & are just poorly & {\leavevmode\dimmed  it a 8} \\
rating 8/ 10 & rated 2 out & {\leavevmode\dimmed   even read the} & shown how poorly & {\leavevmode\dimmed  vote a 8} \\
example 8/ 10 & Rating 2 out & {\leavevmode\dimmed   have read the} & of how poorly & {\leavevmode\dimmed  a Vol 1} \\
... 8/ 10 & conclusion 2 out & {\leavevmode\dimmed   the other posted} & watching this awful & {\leavevmode\dimmed  this story an} \\
\bottomrule
\end{tabular}
%\end{sc}
\end{small}
\end{center}
\vskip -0.1in
\end{table}

\begin{table}[htp]
\caption{Interpreting trigram (instance) clusters in the MIL setting.}
\label{tab:imdb:mitop20words}
\vskip 0.15in
\begin{center}
\begin{small}
%\begin{sc}
\begin{tabularx}{\textwidth}{XXXXXX}
\toprule
 \multicolumn{1}{c}{$u_1$ $(13.53\%)$} & \multicolumn{1}{c}{$u_2$ $(41.53\%)$} & \multicolumn{1}{c}{$u_3$ $(3.03\%)$} & \multicolumn{1}{c}{$u_4$ $(5.47\%)$} & \multicolumn{1}{c}{$u_5$ $(31.58\%)$} & \multicolumn{1}{c}{$u_6$ $(4.85\%)$} \\
\midrule
{\leavevmode\dimmed  production costs \_} & {\leavevmode\dimmed  give it a} & only 4/10 \_ & {\leavevmode\dimmed  is time well-spent} & {\leavevmode\dimmed  ... 4/10 ...} & \_ Recommended \_ \\
{\leavevmode\dimmed  all costs \_} & {\leavevmode\dimmed  gave it a} & score 4/10 \_ & {\leavevmode\dimmed  two weeks hairdressing} & {\leavevmode\dimmed  .. 1/10 for} & Highly Recommended \_ \\
{\leavevmode\dimmed  its costs \_} & {\leavevmode\dimmed  rated it a} & a 4/10 \_ & {\leavevmode\dimmed  2 hours \_} & {\leavevmode\dimmed  rate this a} & Well Recommended \_ \\
{\leavevmode\dimmed  ALL costs \_} & {\leavevmode\dimmed  rating it a} & \_ 4/10 \_ & {\leavevmode\dimmed  two hours \_} & {\leavevmode\dimmed  gave this a} & \_ 7/10 \_ \\
{\leavevmode\dimmed  possible costs \_} & {\leavevmode\dimmed  scored it a} & average 4/10 \_ & {\leavevmode\dimmed  finest hours \_} & {\leavevmode\dimmed  give this a} & 13 7/10 \_ \\
{\leavevmode\dimmed  some costs \_} & {\leavevmode\dimmed  giving it a} & vote 4/10 \_ & {\leavevmode\dimmed  off hours \_} & {\leavevmode\dimmed  rated this a} & rate 7/10 \_ \\
{\leavevmode\dimmed  cut costs \_} & {\leavevmode\dimmed  voting it a} & Rating 4/10 \_ & {\leavevmode\dimmed  few hours \_} & {\leavevmode\dimmed  \_ Not really} & .. 7/10 \_ \\
{\leavevmode\dimmed  rate this a} & {\leavevmode\dimmed  gave this a} & .. 4/10 \_ & {\leavevmode\dimmed  slow hours \_} & {\leavevmode\dimmed  4/10 Not really} & this 7/10 \_ \\
{\leavevmode\dimmed  gave this a} & {\leavevmode\dimmed  give this a} & is 4/10 \_ & {\leavevmode\dimmed  three hours \_} & {\leavevmode\dimmed  a 4/10 or} & Score 7/10 \_ \\
{\leavevmode\dimmed  rating this a} & {\leavevmode\dimmed  rate this a} & this 4/10 \_ & {\leavevmode\dimmed  final hours \_} & {\leavevmode\dimmed  of 4/10 saying} & solid 7/10 \_ \\
{\leavevmode\dimmed  give this a} & {\leavevmode\dimmed  giving this a} & of 4/10 \_ & {\leavevmode\dimmed  early hours \_} & {\leavevmode\dimmed  rate it a} & a 7/10 \_ \\
{\leavevmode\dimmed  and this an} & {\leavevmode\dimmed  gives this a} & movie 4/10 \_ & {\leavevmode\dimmed  six hours \_} & {\leavevmode\dimmed  give it a} & rating 7/10 \_ \\
{\leavevmode\dimmed  give this an} & {\leavevmode\dimmed  like this a} & verdict 4/10 \_ & {\leavevmode\dimmed  48 hours \_} & {\leavevmode\dimmed  gave it a} & to 7/10 \_ \\
{\leavevmode\dimmed  given this an} & {\leavevmode\dimmed  film merits a} & gave 4/10 \_ & {\leavevmode\dimmed  4 hours \_} & {\leavevmode\dimmed  given it a} & viewing 7/10 \_ \\
{\leavevmode\dimmed  gave this an} & {\leavevmode\dimmed  Stupid Stupid Stupid} & 13 4/10 \_ & {\leavevmode\dimmed  6 hours \_} & {\leavevmode\dimmed  giving it a} & it 7/10 \_ \\
{\leavevmode\dimmed  rating this an} & {\leavevmode\dimmed  \_ Stupid Stupid} & disappointment 4/10 \_ & {\leavevmode\dimmed  five hours \_} & {\leavevmode\dimmed  scored it a} & score 7/10 \_ \\
{\leavevmode\dimmed  rate this an} & {\leavevmode\dimmed  award it a} & at 4/10 \_ & {\leavevmode\dimmed  nocturnal hours \_} & {\leavevmode\dimmed  award it a} & movie 7/10 \_ \\
{\leavevmode\dimmed  all costs ...} & {\leavevmode\dimmed  given it a} & rating 4/10 \_ & {\leavevmode\dimmed  17 hours \_} & {\leavevmode\dimmed  Cheesiness 0/10 Crappiness} & is 7/10 \_ \\
{\leavevmode\dimmed  all costs ..} & {\leavevmode\dimmed  makes it a} & ... 4/10 \_ & {\leavevmode\dimmed  for hours \_} & {\leavevmode\dimmed  without it a} & drama 7/10 \_ \\
{\leavevmode\dimmed  \_ Avoid \_} & {\leavevmode\dimmed  Give it a} & rate 4/10 \_ & {\leavevmode\dimmed  wasted hours \_} & {\leavevmode\dimmed  deserves 4/10 from} & Recommended 7/10 \_ \\
\bottomrule
\end{tabularx}
%\end{sc}
\end{small}
\end{center}
\vskip -0.1in
\end{table}

\paragraph{MMIL rules} Using a decision tree learner taking
 frequency vectors $(f_{u_1},\ldots,f_{u_5})$ as inputs,
we obtained the rules reported in Table~\ref{tab:imdb1mmi}.
\begin{table}[tp]
  \caption{MMIL rules mapping instance cluster frequencies ($f_{u_i}$) into
    sub-bag cluster identifiers. Numbers express percentages and rules are written
    as definite clauses in a Prolog-like syntax where $\leftarrow$ is the
    implication and conjuncted literals are joined by a comma.}
  \label{tab:imdb1mmi}
  $$
  \begin{array}{ll}
  \toprule
    1&v_1 \leftarrow  \PAR{f_{u_1}\mathord{\le} 6.03}, \PAR{f_{u_2}\mathord{>}10.04},   \PAR{f_{u_4}\in (2.77,12.59]}.\\
    2&v_1 \leftarrow  \PAR{f_{u_1}\mathord{\le} 16.90}, \PAR{f_{u_4}\mathord{>}12.59}.\\
    %3&\hat{g}=v_1 \leftarrow  \PAR{f_{u_1}\in (8.96, 16.90]},  \PAR{f_{u_4}\mathord{>}12.59}.\\
    {\dimmed 3}& {\dimmed v_2 \leftarrow  \PAR{f_{u_1}\mathord{\le} 8.43}, \PAR{f_{u_2}\mathord{\le}8.88},\PAR{f_{u_4}\mathord{\le} 2.77}}.\\
    %5&\hat{g}=v_2 \leftarrow  \PAR{f_{u_1}\mathord{\le} 8.43}, \PAR{f_{u_2}\in (4.32,8.88]}.\\
    {\dimmed 4}&{\dimmed v_2 \leftarrow  \PAR{f_{u_1}\mathord{>} 3.20},   \PAR{f_{u_2}\in (8.88,20.39]},\PAR{f_{u_4}\mathord{\le} 2.77}}.\\
    {\dimmed 5}&{\dimmed v_2 \leftarrow  \PAR{f_{u_1}\mathord{>} 6.03},   \PAR{f_{u_2}\mathord{\le}6.03},  \PAR{f_{u_4}\in (2.77,12.59]}}.\\
    6&v_3 \leftarrow  \PAR{f_{u_1}\mathord{>} 8.43},   \PAR{f_{u_2}\mathord{\le}8.88},\PAR{f_{u_4}\mathord{\le}2.77}.\\
    7&v_4 \leftarrow  \PAR{f_{u_1}\mathord{\le}3.20},  \PAR{f_{u_2}\mathord{>}8.88},\PAR{f_{u_4}\mathord{\le}2.77}.\\
    8&v_4 \leftarrow  \PAR{f_{u_1}\mathord{>}3.20},    \PAR{f_{u_2}\mathord{>}20.39},\PAR{f_{u_4}\mathord{\le}2.77}.\\
    9&v_4 \leftarrow  \PAR{f_{u_1}\mathord{\le}6.03},  \PAR{f_{u_2}\mathord{\le}10.04}, \PAR{f_{u_4}\in (2.77,12.59]}.\\
    10&v_4 \leftarrow  \PAR{f_{u_1}\mathord{>}6.03},    \PAR{f_{u_2}\mathord{>}6.03},    \PAR{f_{u_4}\in(2.77,12.59]}.\\
    11&v_4 \leftarrow  \PAR{f_{u_1}\mathord{>}16.90},   \PAR{f_{u_4}\mathord{>}12.59}.\\
  \bottomrule
  \end{array}
  $$
\end{table}
Even though these rules are somewhat more difficult to parse than the
ones we obtained in Section~\ref{sec:semi-mnist}, they still express
relatively simple relationships between the triplet and sentence
clusters.  Especially the single sentence cluster $v_3$ that
corresponds to a clearly positive sentiment has a very succinct
explanation given by the rule of line 6.
Rules related to sentence cluster $v_2$ are printed in grey.
Since $v_2$ is not used by any of the rules shown in Table~\ref{tab:imdb2mmi}
that map sub-bag (sentence) cluster identifiers to the top-bag (review) class labels,
the rules for $v_2$ will never be required to explain a particular classification.

\begin{table}[tp]
  \caption{MMIL rules mapping sentence cluster frequencies
    into review sentiment labels. See the caption of
    Table~\ref{tab:imdb1mmi} for details on the syntax.}
    \label{tab:imdb2mmi}
  $$
  \begin{array}{ll}
    \toprule
    1&\mbox{positive} \leftarrow \PAR{f_{v_1}\mathord{\le}4.04},\PAR{f_{v_3}\mathord{\le}12.63},  \PAR{f_{v_4}\mathord{\le}39.17}.\\
    2&\mbox{positive} \leftarrow \PAR{f_{v_1}\mathord{\le}12.97}, \PAR{f_{v_3}\mathord{>}12.63}.\\
    3&\mbox{positive} \leftarrow \PAR{f_{v_1}\mathord{>}12.97},   \PAR{f_{v_3}\mathord{>}25.66}.\\
    4&\mbox{negative} \leftarrow \PAR{f_{v_1}\mathord{\le}4.04},  \PAR{f_{v_3}\mathord{\le}12.63},\PAR{f_{v_4}\mathord{>}39.17}.\\
    5&\mbox{negative} \leftarrow \PAR{f_{v_1}\mathord{>}4.04},\PAR{f_{v_3}\mathord{\le}12.63}.\\
    6&\mbox{negative} \leftarrow \PAR{f_{v_1}\mathord{>}12.97},    \PAR{f_{v_3}\in (12.63, 25.66]}.\\
    \bottomrule
  \end{array}
  $$

\end{table}

\paragraph{MIL rules.} For the MIL model, rules map a bag $\TopBag$,
described by its instance frequency vector
$(f_{u_1},\ldots,f_{u_5})$, to the bag class label. They are reported
in Table~\ref{tab:imdb2mi}. Note that only two out of the six instance
clusters are actually used in these rules.
\begin{table}[tp]
  \caption{MIL classification rules. See the caption of Table~\ref{tab:imdb1mmi} for details on
    the syntax.}
    \label{tab:imdb2mi}
  $$
  \begin{array}{ll}
    \toprule
    1&\mbox{positive} \leftarrow \PAR{f_{u_3}\mathord{\le}1.11}, \PAR{f_{u_6}\mathord{\le}3.42}.\\
    2&\mbox{positive} \leftarrow \PAR{f_{u_3}\mathord{\le}2.21}, \PAR{f_{u_6}\mathord{>}3.42}.\\
    3&\mbox{positive} \leftarrow \PAR{f_{u_3}\in(2.21, 5.81]}, \PAR{f_{u_6}\mathord{>}6.30}.\\
    4&\mbox{negative} \leftarrow \PAR{f_{u_3}\in(1.11, 2.21]}, \PAR{f_{u_6}\mathord{\le}3.42}.\\
    5&\mbox{negative} \leftarrow \PAR{f_{u_3}\mathord{>}2.21}, \PAR{f_{u_6}\mathord{\le}6.30}.\\
    6&\mbox{negative} \leftarrow \PAR{f_{u_3}\mathord{>}5.81}, \PAR{f_{u_6}\mathord{>}6.30}.\\
    \bottomrule
  \end{array}
  $$

\end{table}

By classifying IMDB using the rules and cluster identifiers, we achieved
an accuracy of  $87.49\%$ on the test set  for the MMIL case and
$86.37\%$ for the MIL case. Fidelities in the MMIL and in the MIL settings were
$90.40\%$ and $88.10\%$, respectively. We thus see that the somewhat higher
complexity of the rule-based explanation of the model learned in the MMIL setting also corresponds
to a somewhat higher preservation of accuracy. As we demonstrate
by  the following example, the multi-level
explanations derived from model learned in the MMIL setting can also lead to more transparent
explanations for individual predictions.

\paragraph{An example of prediction explanation.}
As an example we consider a positive test-set review for the movie
Bloody Birthday, which was classified correctly by the MMIL rules and
incorrectly by the MIL rules. Its full text is reported in
Table~\ref{tab:sample}. Classification in the MMIL setting was positive
due to applicability of rule 2 in Table~\ref{tab:imdb2mmi}. This rule only
is based on sentences in clusters $v_1$ and $v_3$, and therefore
sentences assigned in any other cluster do not actively contribute to this classification.
These irrelevant sentences are dimmed in the printed text (Table~\ref{tab:sample}, top part).
A first, high-level
explanation of the prediction is thus obtained by simply using the sentences
that are active for the classification as a short summary of the most
pertinent parts of the review.

This sentence-level explanation can be refined by also explaining the
clusters for the individual sentences. For example,
sentence ``Bloody Birthday a $\dots$'' was assigned to identifier $v_1$
using rule 2 of Table~\ref{tab:imdb1mmi}.  This rule is based on frequencies
of the trigram cluster identifiers $u_1$ and $u_4$. Occurrences of trigrams with
these identifiers are highlighted in boldface and superscripted with the trigram
cluster identifier  in the text, thus exhibiting the sub-structures in the
sentence that are pertinent for the classification.
Similarly, the other three relevant
sentences were all assigned to identifier $v_3$ because of rule 6 in
Table~\ref{tab:imdb1mmi}, which is based on identifiers $u_1,u_2,u_4$.
These formal, logical explanations
for the classifications are complemented by the semantic insight into the
cluster identifiers provided by Tables~\ref{tab:imdb:top20words} and
\ref{tab:imdb:top20sents}.

The review was classified as negative in the
MIL setting. The applicable rule here was rule 5 in Table~\ref{tab:imdb2mi} which
involved triplet identifiers $u_3$, and $u_6$. The relevant triplets are
highlighted in boldface in the lower part of Table~\ref{tab:sample}.

A second example of prediction explanation is reported in
Appendix~\ref{apx:imdb}.

\begin{table}
  \caption{A sample positive review. Top: MMIL labeling. Bottom: MIL labeling.}
  \label{tab:sample}
  \begin{tcolorbox}[colback=white,sharp corners,boxrule=0.1mm]
    {\small
      \sf
      {\dimmed Story about three eclipse (maybe even Indigo, ha) children
        beginning their love for murder. Oh, and the people who are ``hot''
        on their trail.}

      [$v_1$] Bloody \textbf{Birthday, a pretty mediocre title}$^4$ for
      the \textbf{film, was a nice lil}$^{1}$ surprise. {\dimmed I was
        in no way expecting a film that dealt with blood-thirsty
        psychopath kids.}

      [$v_3$] And I may say it's also \textbf{one of the best flicks}$^1$ I've seen
      with kids as the villains.
      {\dimmed By the end of the movie I
        seriously wanted these kids to die in horrible fashion.}

      [$v_3$] \textbf{It's a really solid 80s}$^1$ horror flick, but how these kids are
      getting away with all this mayhem and murder is just something
      that \textbf{you can't not}$^2$ think about.
      {\dimmed Even the slightest bit of investigation would easily
        uncover these lil sh!ts as the murderers. But there seems to be
        only a couple police in town, well by the end, only one, and he
        seemed like a dimwit, so I suppose they could have gotten away
        with it. Haha, yeah, and I'm a Chinese jet-pilot.}

      {\dimmed Nevertheless, this movie delivered some evilass kids who
        were more than entertaining, a lot of premarital sex and a
        decent amount of boobage. No kiddin! If you're put off by the
        less than stellar title, dash it from your mind and give this
        flick a shot.}
      [$v_3$] It's \textbf{a very recommendable and underrated 80s}$^1$ horror
      flick.}
  \end{tcolorbox}
  \begin{tcolorbox}[colback=white,sharp corners,boxrule=0.1mm]
    {\small
      \sf

      Story about three eclipse (maybe even Indigo, ha) children
      beginning their love for murder. Oh, and the people who are
      ``hot'' on their trail.

      Bloody \textbf{Birthday, a pretty mediocre title}$^3$ for the
      film, was a nice lil surprise. I was in no way expecting a film
      that dealt with blood-thirsty psychopath kids.  And I may say it's
      also \textbf{one of the best flicks}$^6$ I've seen with kids as the villains.
      By the end of the movie I seriously wanted these kids \textbf{to die in
        horrible fashion}$^3$.

      \textbf{It's a really solid}$^6$ 80s horror flick, but how these
      kids are getting away with all this mayhem and murder is just
      something that you can't not think about.  Even the slightest bit
      of investigation would easily uncover these lil sh!ts as the
      murderers. But there seems to be only a couple police in town,
      well by the end, only one, and he seemed like a dimwit, so I
      suppose they could have gotten away with it. Haha, yeah, and I'm a
      Chinese jet-pilot.

      Nevertheless, this movie delivered some evilass kids who were more
      than entertaining, a lot of premarital sex and a decent amount of
      boobage. No kiddin! If you're put off by \textbf{the less than}$^6$ stellar
      title, dash it from your mind and give this flick a shot.  It's \textbf{a
        very recommendable and underrated 80s}$^6$ horror flick.
    }
  \end{tcolorbox}
\end{table}

\subsection{Citation Datasets}\label{sec:cit-datasets}
In the following experiments, we apply MMIL to graph learning
according to the general strategy described in Section
\ref{sec:graph}. We also present a (generalized) MIL approach to graph
learning in which latent instance labels need not be binary, and need not be
related to the bag label according to the conventional MIL rule.  We considered
three citation datasets from~\citep{sen_collective_2008}: Citeseer,
Cora, and PubMed. Finally, the MMIL network trained on PubMed will be
mapped into an interpretable model using the procedure described in
Section~\ref{sec:interpreting}.

We view the datasets as graphs where
nodes represent papers described by titles and abstracts, and
edges are citation links.
We treat the citation links as undirected edges, in order to have a setup
as close as possible to earlier works,
\citep{kipfsemi-supervised2016,hamilton2017inductive}.
The goal is to classify papers  according to their subject area.

We collected the years of publication for all the papers of each dataset, and
for each dataset determined two thresholds $yr_1 < yr_2$, so that
papers with publication year $yr\leq yr_1$ amount to approximately $40\%$ of
the data and are used as the training set, papers with publication year $yr_1<yr\leq yr_2$
formed a validation set of about $20\%$, and
papers with publication year $yr>yr_2$ are the test set of $40\%$ of the data.
Table \ref{tab:citation-datasets}
reports the statistics for each dataset. More details on the
temporal distributions in the three datasets  are given
in Appendix~\ref{apx:citation} (Figure~\ref{fig:citation:dist}).
This particular split is justified by the fact that we would like to evaluate those datasets in a more realistic and challenging scenario, where we used past publications to predict the class of the new ones. Furthermore, CiteSeer, Cora, and PubMed are typically evaluated in a transductive setting, where the feature vectors, associated with the test set nodes, are allowed to be used during the training. Here, instead, we consider an inductive setup, where the test set nodes remain unobserved during the training. However, for an exhaustive comprehension, we evaluated the three datasets considering also 10 random splits (with the same proportions reported in Table \ref{tab:citation-datasets}) while maintaining the inductive setting.

\begin{figure}[ht]
  \centering
  \includegraphics[width=\textwidth]{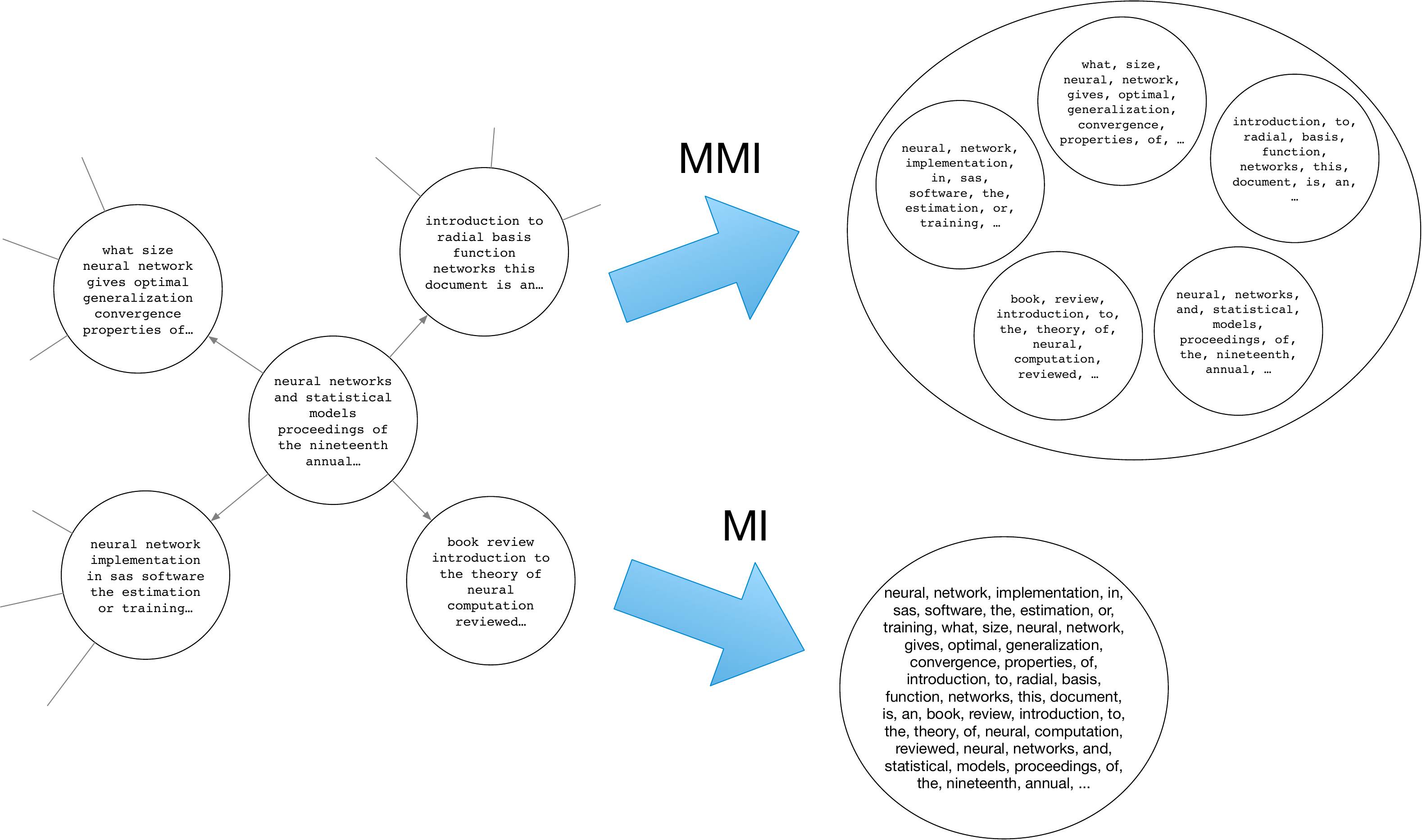}
  \caption{Given node an its neighborhood of a citation graph (left picture) we decomposed it as MMIL data (upper right picture) and MIL data (bottom right picture).}%
  \label{fig:citation:decomposition}
\end{figure}

\begin{table}[ht]
\caption{Structure of the citation graphs. With $yr$ we denote the year of publication. Citeseer classes are 6 among Agents,
    Artificial Intelligence (AI), Database (DB), Human-computer Interaction (HCI), Information Retrieval (IR),
    Machine Learning (ML). Cora classes are 7 among
    Case Based, Genetic Algorithms, Neural Networks, Probabilist Methods,
    Reinforcement Learning, Rule Learning, Theory. PubMed classes are 3 among
    Diabetes Mellitus Experimental (DME), Diabetes Mellitus Type 1 (DMT1),
    Diabetes Mellitus Type 2(DMT2).}
\label{tab:citation-datasets}
\vskip 0.15in
\begin{center}
\begin{small}
\begin{sc}
\begin{tabular}{ccccccc}
\toprule
Dataset & \# Classes & \# Nodes & \# Edges & \# Training & \# Validation & \# Test \\
\midrule
\multirow{2}[1]{*}{CiteSeer} & \multirow{2}[1]{*}{6} & \multirow{2}[1]{*}{3,327} & \multirow{2}[1]{*}{4,732} & 1,560 & 779 & 988 \\
&  &  &  & $(yr \le `99)$ & $(`99<yr \le `00)$ & $(yr>`00)$ \\
\cmidrule{1-7}
\multirow{2}[1]{*}{Cora} & \multirow{2}[1]{*}{7} & \multirow{2}[1]{*}{2,708} & \multirow{2}[1]{*}{5,429} & 1,040 & 447 & 1,221  \\
&  &  &  & $(yr \le `94)$ & $(`94<yr \le `95)$ & $(yr>`95)$ \\
\cmidrule{1-7}
\multirow{2}[1]{*}{PubMed} & \multirow{2}[1]{*}{3} & \multirow{2}[1]{*}{19,717} & \multirow{2}[1]{*}{44,338} & 8,289 & 3,087 & 8,341 \\
&  &  &  & $(yr \le `97)$ & $(`97<yr \le `01)$ & $(yr>`01)$ \\
\bottomrule
\end{tabular}
\end{sc}
\end{small}
\end{center}
\vskip -0.1in
\end{table}
MMIL data was constructed from citation networks in which a top-bag $\TopBag$
corresponds to a paper represented as the
bag of nodes $\SubBag{j}$ containing the paper itself and all its  neighbors.
The nodes $\SubBag{j} \in \TopBag$ are further decomposed as (sub-) bags of the words
contained in the  text (i.e. title and abstract) attached to the node.
An instance $\Instance{j}{\ell} \in \SubBag{j}$ is a word.
Similarly, MIL data was constructed  in which each paper is simply
represented as the bag of all words appearing in the text of the paper
or its neighbors.
Figure \ref{fig:citation:decomposition} shows an example of
MMIL and MIL decompositions starting from a node and its neighborhood
of a citation graph.
Words are encoded as one-hot vectors, in order to evaluate
the capability of
our model to learn relevant intermediate representations of bags from scratch.

We used an MMIL network model with two stacked bag-layers
with ReLU activations with 250 units. The MIL model has
one bag-layer with ReLU activations with 250 units. For both MMIL and
MIL we proposed two versions which differ only for the aggregation
functions for the bag-layers: one version uses max, the other uses
mean aggregation.
All models were trained by minimizing the softmax cross-entropy loss.
We ran 100 epochs of the Adam optimizer
with learning rate 0.001 and we early stopped
the training according to the loss on the
validation set.

As baselines, we considered N\"aive Bayes and logistic regression.
For these two models we reduced the task to a standard classification
problem in which  papers are represented by bag of words
feature vectors (only for the words associated with the papers themselves, not
considering citation neighbors).
We  also compared our models against GCN~\citep{kipfsemi-supervised2016}, GraphSAGE~\citep{hamilton2017inductive} and GAT~\citep{DBLP:conf/iclr/VelickovicCCRLB18}, which are briefly described
in Section \ref{sec:related}. GCN and GAT represent nodes as bags of words, while
GraphSAGE exploits the sentence embedding approach
described by~\citep{arora2016simple}. For comparison reasons and given that
bag of words represent the most challenging and standalone approach which
does not rely in any embedding representation of words,
we encoded the nodes as bag of words for GCN, GraphSAGE, and GAT.
As GraphSAGE allows to use both max and mean as aggregation functions, we
compared our models against both versions, with a subsampling neighbor size of 25.

Results in Table \ref{tab:citation:res} report the accuracy and the standard deviation for all the models. The columns \textsc{Temp} refer to our temporal splits, while the columns \textsc{Rnd} refer to random splits. The random splits are kept fixed for all the experiments. The standard deviations are evaluated on 10 restarts for the \textsc{Temp}  and on 10 randoms splits for the \textsc{Rnd}.
The MMIL networks outperform the other methods.
MIL networks show a similar performance to GCN, GraphSage and GAT on
Cora and Citeseer, and are close to MMIL on PubMed.
It is noteworthy that the quite generic MMIL framework which here only
is instantiated for graph data as a special case outperforms the
methods that are specifically designed for graphs. urthermore, the experimental results suggest that the temporal splits provide a more challenging task than random splits. Finally, as typically Cora, CiteSeer, and PubMed are used in the transductive setup we also report in Table \ref{tab:citation:transductive:res} the results for this setting. Here we used the same splits described in \cite{kipfsemi-supervised2016} and \cite{DBLP:conf/iclr/VelickovicCCRLB18}. \textsc{MMIL-Mean}
achieved comparable results with GCN and slightly worse than GAT.
Hyperparameters for models appearing in both Table~\ref{tab:citation:res}, 
and~\ref{tab:citation:transductive:res} are the same and they come from
the respective papers.

\begin{table}[ht]
\caption{\textbf{Inductive setting}. Accuracies with standard deviations on the test sets. Best results are highlighted in bold.}
\label{tab:citation:res}
\vskip 0.15in
\begin{center}
\begin{small}
\begin{sc}
\begin{tabular}{lcccccc}
\toprule
\multirow{2}[1]{*}{Model} &  \multicolumn{2}{c}{Cora} & \multicolumn{2}{c}{CiteSeer} & \multicolumn{2}{c}{PubMed} \\
\cmidrule(lr){2-3} \cmidrule(lr){4-5} \cmidrule(lr){6-7}
 & Temp & Rnd & Temp & Rnd & Temp & Rnd \\
\midrule
N. Bayes & 71.3 & $73.8 \pm 1.4$ & 63.8 & $71.6 \pm 1.0$ & 75.5 & $79.4 \pm 0.5$ \\
\midrule
Log Reg & 74.9 & $76.5 \pm 1.0$ & 64.4 & $71.1 \pm 1.7$ & 73.7 &$86.0 \pm 0.3$ \\
\midrule
GCN & $81.2 \pm 0.7$ & $83.7 \pm 0.8$ & $66.4 \pm 0.6$ & $73.6 \pm 1.5$ & $78.0 \pm 1.0$ &$83.7 \pm 0.3$ \\
\midrule
GS-Mean & $80.1 \pm 0.8$ & $83.4 \pm 1.1$ & $65.3 \pm 1.6$ & $73.6 \pm 0.6$ & $75.4 \pm 0.9$ & $83.8 \pm 0.4$\\
\midrule
GS-Max  & $80.0 \pm 0.7$ & $83.4 \pm 1.1$ & $65.3 \pm 1.2$ & $73.3 \pm 1.1$ & $74.3 \pm 0.7$ & $83.9 \pm 0.6$\\
\midrule
GAT  & $80.2 \pm 2.4$ & $82.0 \pm 1.7$ & $68.2 \pm 1.8$ & $68.7 \pm 1.7$ & $80.8 \pm 1.4$ & $84.9 \pm 0.8$\\
\midrule
Mil-Mean  & $81.0 \pm 1.0$ & $84.0 \pm 1.4$ & $70.2 \pm 0.6$ & $\bm{75.5 \pm 1.5}$ & $80.1 \pm 0.9$ & $\bm{86.2 \pm 0.2}$\\
\midrule
Mil-Max  & $81.3 \pm 0.8$ & $83.8 \pm 1.2$ & $68.7 \pm 0.8$ & $75.0 \pm 1.2$ & $79.2 \pm 0.9$ & $84.7 \pm 0.6$\\
\midrule
Mmil-Mean  & $83.0 \pm 0.9$ & $83.3 \pm 1.2$ & $\bm{71.3 \pm 1.1}$ & $74.5 \pm 1.6$ & $\bm{82.4 \pm 0.9}$ & $86.0 \pm 0.3$\\
\midrule
Mmil-Max  & $\bm{83.6 \pm 0.5}$ & $\bm{84.1 \pm 1.5}$ & $70.1 \pm 0.6$ & $75.0 \pm 1.4$ & $79.0 \pm 1.7$ & $85.0 \pm 0.7$\\
\bottomrule
\end{tabular}
\end{sc}
\end{small}
\end{center}
\vskip -0.1in
\end{table}

\begin{table}[ht]
\caption{\textbf{Transductive setting}. Accuracies with standard deviations on the test sets. Best results are highlighted in bold.}
\label{tab:citation:transductive:res}
\vskip 0.15in
\begin{center}
\begin{small}
\begin{sc}
\begin{tabular}{lccc}
\toprule
Model &  Cora & CiteSeer & PubMed \\
\midrule
GCN & $81.5$ & $70.3$ &  $79.0$ \\
\midrule
GAT  &  $83.0 \pm 0.7$ & $72.5 \pm 0.7$ & $79.0 \pm 0.3$ \\
\midrule
Mil-Mean  & $78.5 \pm 0.7$ & $66.8 \pm 0.9$ & $77.1 \pm 0.5$ \\
\midrule
Mil-Max  & $75.7 \pm 0.9$ & $66.8 \pm 1.1$ & $73.1 \pm 0.6$ \\
\midrule
Mmil-Mean  & $82.1 \pm 0.5$ & $69.7 \pm 0.4$ & $78.1 \pm 0.3$ \\
\midrule
Mmil-Max  & $77.9 \pm 0.7$ & $66.6 \pm 1.2$ & $73.4 \pm 0.6$ \\
\bottomrule
\end{tabular}
\end{sc}
\end{small}
\end{center}
\vskip -0.1in
\end{table}

Our general approach to interpretability can also be applied to models learned
in the MIL and in the MMIL settings for the citation graphs. The associated
interpretability study on PubMed data is reported in
Appendix~\ref{apx:citation}.

\subsection{Social Network Datasets}\label{sec:social:data}

We finally test our model on a slightly different type of prediction problems for
graph data, where the task is
graph classification, rather than node classification as in the previous section.
For this we use the following six publicly available datasets
first proposed by
\cite{yanardag2015deep}.

\begin{itemize}
        \item COLLAB is a dataset where each graph represent the ego-network of a researcher, and the task is to determine the field of study of the researcher, which is one of \emph{High Energy Physics}, \emph{Condensed Matter Physics}, or \emph{Astro Physics}.
        \item IMDB-BINARY, IMDB-MULTI are datasets derived from IMDB. First, genre-specific collaboration networks are constructed where nodes represent actors/actresses who are connected by an edge if they have appeared together in a  movie of a given genre. Collaboration networks are generated for the genres \emph{Action} and \emph{Romance} for IMDB-BINARY and \emph{Comedy}, \emph{Romance}, and \emph{Sci-Fi} for IMDB-MULTI. The data then consists of the ego-graphs for all actors/actresses in all genre networks, and the task is to identify the genre from which an ego-graph has been extracted.
        \item REDDIT-BINARY, REDDIT-MULTI5K, REDDIT-MULTI12K are datasets where each graph is derived from a discussion thread from Reddit. In those graphs each vertex represent a distinct user and two users are connected by an edge if one of them has responded to a post of the other in that discussion. The task in REDDIT-BINARY is to discriminate between threads originating from a discussion-based subreddit (TrollXChromosomes, atheism) or from a question/answers-based subreddit (IAmA, AskReddit).

        The task in REDDIT-MULTI5K and REDDIT-MULTI12K is a multiclass classification problem where each graph is labeled with the subreddit where it has originated (\emph{worldnews, videos, AdviceAnimals, aww, mildlyinteresting} for REDDIT-MULTI5K and \emph{AskReddit, AdviceAnimals, atheism, aww, IAmA, mildlyinteresting, Showerthoughts, videos, todayilearned, worldnews, TrollXChromosomes} for REDDIT-MULTI12K).
\end{itemize}

We transformed each dataset into MMIL data by treating
each graph as a top-bag $\TopBag$.
Each node of the graph with its neighborhood,
is a sub-bag $\SubBag{j} \in \TopBag$ while each node  $\Instance{j}{\ell} \in \SubBag{j}$ is an instance.

In these six datasets no features are
attached to the nodes. We therefore defined a node feature vector based
on the degrees $deg(\Instance{j}{\ell} )$ of the nodes as follows: let  $deg^{*}$ be the maximum degree
of any node. For $i=1,\ldots,deg^{*}$ we then define
\begin{equation}
        \Instance{j}{\ell} ^ i =
        \begin{cases}
                \frac{1}{\sqrt{deg(\Instance{j}{\ell})}} & \quad if \ i < deg(\Instance{j}{\ell}) \\
                0 & \quad otherwise,
        \end{cases}
\end{equation}
By using this representation
the scalar product of two node feature vectors
will be high if the nodes
have similar degrees, and it will be low for nodes with very different degrees.

The MMIL networks have the same structure for all the datasets:
a dense layer with 500 nodes and ReLU activation, two stacked bag-layers
with 500 units (250 max units and 250 mean units), and a dense
layer with $dim_{out}$ nodes and linear activation, where $dim_{out}$
is $3$ for COLLAB, $2$ for IMDB-Binary,
and $3$ for IMDB-MULTI, $2$ for REDDIT-BINARY, $5$
for REDDIT-MULTI5K, and $11$ for REDDIT-MULTI12K.
We performed a 10 times 10 fold cross-validation, training
the MMIL networks by minimizing the binary cross-entropy
loss (for REDDIT-BINARY and IMDB-BINARY)
and the softmax cross-entropy loss (for COLLAB, IMDB-MULTI, REDDIT-5K, REDDIT-12K).
We ran 100 epochs of the Adam optimizer with learning rate 0.001 on mini-batches of size 20.

We compared our method against DGK~\citep{yanardag2015deep},
Patchy-SAN~\citep{niepert_learning_2016}, and SAEN~\citep{orsini2018shift}.

\begin{table}[ht]
\caption{Accuracies with standard deviations in graph classification. Best results are highlighted in bold.}
\label{tab:result-social-datasets}
\vskip 0.15in
\begin{center}
\begin{small}
\begin{sc}
\begin{tabular}{lcccc}
\toprule
Dataset & DGK & Patchy-SAN & SAEN & MMIL \\
\midrule
Collab & $73.09 \pm 0.25$ & $72.60 \pm 2.15$ & $78.50 \pm 0.69$ & $\bm{79.46 \pm 0.31}$ \\
Imdb-Binary & $66.96 \pm 0.56$ & $71.00 \pm 2.29$ & $71.59 \pm 1.20$ & $\bm{72.62 \pm 1.04}$ \\
Imdb-Multi & $44.55 \pm 0.52$ & $45.23 \pm 2.84$ & $48.53 \pm 0.76$ & $\bm{49.42 \pm 0.68}$ \\
Reddit-Binary & $78.04 \pm 0.39$ & $86.30 \pm 1.58$ & $\bm{87.22 \pm 0.80}$ & $86.54 \pm 0.64$ \\
Reddit-Multi5k & $41.27 \pm 0.18$ & $49.10 \pm 0.70$ & $\bm{53.63 \pm 0.51}$ & $53.42 \pm 0.67$ \\
Reddit-Multi12k & $32.22 \pm 0.10$ & $41.32 \pm 0.42$ & $\bm{47.27 \pm 0.42}$ & $45.25 \pm 0.48$ \\
\bottomrule
\end{tabular}
\end{sc}
\end{small}
\end{center}
\vskip -0.1in
\end{table}

Results in Table \ref{tab:result-social-datasets} show that MMIL networks
and SAEN perform comparably, with some advantages of these two methods over Patch-SAN, and
more pronounced advantages over DGK.

\subsection{Point Clouds}\label{ex:point:clouds}
Following the experiments reported in~\citep{zaheer2017deep}, we aim at
illustrating the benefits of the MMIL setting for point clouds datasets and at
demonstrating results of our interpretation framework. We start from the
ModelNet40 dataset~\citep{wu20153d} which consists of $9,843$ training and
$2,468$ test point clouds of objects distributed over $40$ classes. The
dataset is preprocessed with the same procedure described
by~\cite{zaheer2017deep}. To investigate the effect of spatial resolution, we
produce point clouds with 100 and 5,000 three-dimensional points (instances)
from the mesh representation of objects using the point-cloud library's
sampling routine~\citep{rusu20113d}.  Each set is normalized to have zero mean
(along individual axes) and unit (global) variance.  We call P100 and P5000
the two datasets, which are readily usable for the MIL setting (or DeepSets).

\begin{figure}[ht]
  \centering
  \includegraphics[width=\textwidth]{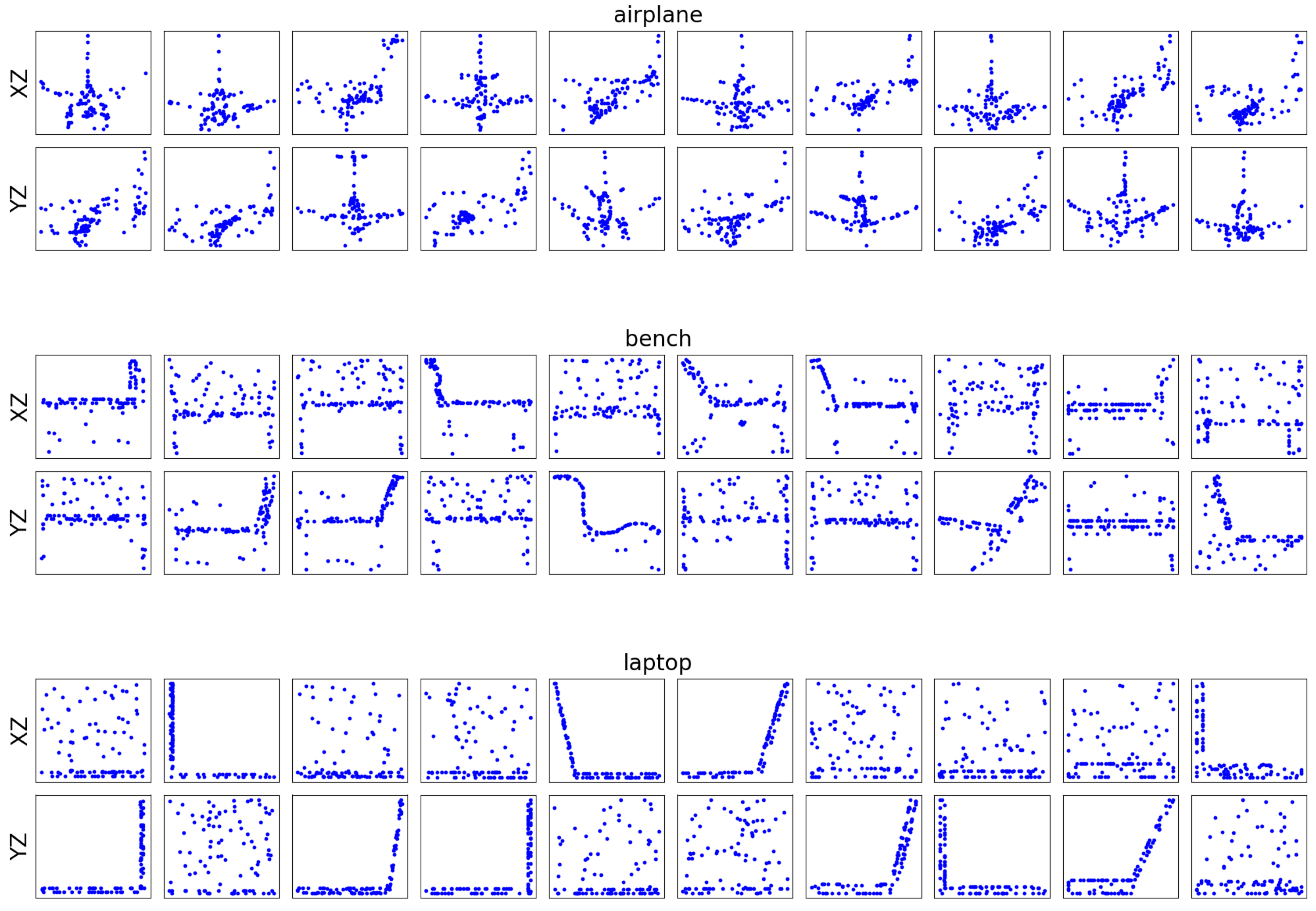}
  \caption{Examples of point clouds projected into the XZ and YZ planes.}%
  \label{fig:point:cloud:proj}
\end{figure}

The orientations of the point clouds in the dataset are different as clearly shown in Figure \ref{fig:point:cloud:proj}, where we reported the projections on the XZ and YZ planes of some point clouds drawn from \emph{airplane}, \emph{bench}, and \emph{laptop} classes. Each column of each class represents the same point cloud projected into the XZ and YZ planes, respectively. Methods that do not take into account this fact might encounter difficulties. On other hand, the bags-of-bags representation is a very natural way to effectively handle different orientations for this dataset. Therefore, we subsequently created bags of bags by considering $R$ equally distributed
rotations, i.e. $\{ \frac{2\pi i}{R}\}_{i=1}^N$, around the $z$-axis.  We used
$R=5$ in P100 and $R=16$ in P5000.  A top-bag $\TopBag$ is thus a set of rotated
versions of the same point cloud, i.e. a set of sub-bags
$\SubBag{j} \in \TopBag, \ j=1,\dots, R$.  %Representing the point clouds as MMIL data is a natural way to effectively handle different orientations, as confirmed by t.

In the MMIL setting, networks have exactly the same
structure (and the same hyper-parameters) of the DeepSets permutation
invariant model described by~\cite{zaheer2017deep}, except for adding a
further Bag-Layer of $40$ sum units before the last layer. We compare our MMIL
results against the DeepStets results reported in~\citep{zaheer2017deep} but
also against DeepSets trained on the ``augmented'' datasets obtained by
flattening the MMIL datasets at the level of subbags.  Results in
Table~\ref{tab:point:cloud} show that in both datasets there is an advantage
in using the MMIL setting and the difference is more pronounced at low spatial
resolution (i.e. on the P100 dataset).

We applied our interpretability approach also to the P100 dataset,
whose study is reported in Appendix~\ref{apx:point:cloud}.

\begin{table}[t]
\caption{Accuracies with standard deviations for the ModelNet40 dataset. Best results are highlighted in bold.}
\label{tab:point:cloud}
\vskip 0.15in
\begin{center}
\begin{small}
\begin{sc}
\begin{tabular}{lll}
\toprule
Model           & Dataset           & Accuracy \\
\midrule
MIL (DeepSets)  & P100              & $82.00 + 2.00\%$ \\
MIL (DeepSets)  & P100 w/ rotations & $85.35 \pm 0.49\%$ \\
MMIL            & P100 bags-of-bags &  $\bm{88.10 \pm 0.43\%}$ \\
\cmidrule{1-3}
MIL (DeepSets)  & P5000             & $90.00 + 0.30\%$ \\
MIL (DeepSets)  & P5000 w/ rotations& $89.28 \pm 0.39\%$ \\
MMIL            & P5000 bags-of-bags&  $\bm{91.17 \pm 0.47\%}$ \\
\bottomrule
\end{tabular}
\end{sc}
\end{small}
\end{center}
\vskip -0.1in
\end{table}

\clearpage

\subsection{Plant Distribution Data}
\label{sec:plants}
In this section we apply our MMIL framework to a botanical dataset containing
plant species distribution data for Germany~\citep{geoflora}.
In this data, Germany is divided on a regular grid into 12,948 geographic regions.
For each region $i$ and each of 4,842 different  plant species $j$  the data contains a binary variable  $a_{ij}$ indicating whether species $j$ has been observed to occur in region $i$. In addition, for each regions $i$ the data provides
the latitude and longitude of $i$'s center. For our experiments, we reduced the data by selecting
only the 842 most frequent species, which includes all plants that occur in at least 10 \% of the regions. Deleting all regions that then do not contain any of the 842 selected species, also leads to a slight
reduction in the number of regions to 12,665.
The data is based on observations in the field by human experts, and is very unlikely to be completely correct.
In particular, some occurrences of a species will be overlooked, leading to
false negatives $a_{ij}=0$ in the data (there can also be false positives, but these can be assumed to be much
more rare).
The real-world task is to identify the most likely false negatives in the data, which could guide efforts to
improve data completeness by additional field studies.
We observe that our task is very similar to recommendation problems from binary, positive-only
data~\citep{verstrepen2017collaborative}. The main difference to a generic collaborative filtering
setting lies in the fact that in addition to the pure occurrence matrix $A$ we also can use
the known spatial relationships between regions. In the following, we consider methods that do or do not
try to exploit the spatial information. Methods of the latter type can be applied to a multitude of similarly
structured recommendation problems.

Since we lack the ground truth complete data, we proceed
as follows to evaluate the potential of different prediction techniques with regard to the real-world task:
let $A = (a_{ij})\in \{0,1\}^{12,665 \times 842}$ denote the original data. We take $A$ as a surrogate for the
ground truth, and construct incomplete versions  $A^P$ of $A$ by randomly setting for each plant $j$
$P\%$ of the occurrences $a_{ij}=1$ to  $a^P_{ij}=0$. We constructed such matrices $A^P\in \{0,1\}^{12,665 \times 842}$
for  $P \in \{5,10,15,20,25,30\}$.

We now consider methods that using $A^P$ as training data compute for each pair $(i,j)$ a score
$\hat{a}_{ij}$ for the actual occurrence of $j$ at $i$. We evaluate the methods based on how highly
the false 0's of $A^P$ are ranked by $\hat{a}_{ij}$. Concretely, for a fixed region $i$ let
$Z_i:= \{j | a^P_{ij} = 0\}$ and $Q_i:=\{j \in Z_i|  a_{ij}=1\}$. For $j\in Z_i$ let $\rho(j)$ be
the rank of $j$ when $Z_i$ is sorted according to the scores $\hat{a}_{ij}$. We then define the mean
average precision for the scores at region $i$ as

\begin{equation}\label{eq:map}
mAP_i = \frac{1}{|Q_i|}\sum\limits_{i=1}^{|Q_i|} \frac{1}{i}
\sum\limits_{r=1}^i  a_{i\rho^{-1}(r)}  .
\end{equation}
$mAP_i$ attains a maximal value of 1 if the plants $j\in Q_i$ are exactly
the highest scoring species in $Z_i$. Note that $ mAP_i$ does not depend on the score values
$\hat{a}_{ij}$ for species $j$ with $a^P_{ij} =1$. For an overall evaluation, we take the
average of the $mAP_i$ over all regions $i$.

Several methods will depend on proximity measures between regions $i_1$ and $i_2$.
In the following, $a_{i*}$ denotes the $i$th row in the matrix $A$.
We consider the following two metrics:
\begin{itemize}
\item \textbf{Hamming distance}: the Hamming distance between the vectors $a^P_{i_1 *}$ and  $a^P_{i_2*}$.
\item \textbf{Euclidean distance}: latitude and longitude of $i_1,i_2$ are converted into Cartesian
  coordinates, and then Euclidean metric is applied.
\end{itemize}
Note that only Euclidean distance exploits the available spatial information.

\begin{figure}[htp]
  \centering
  \includegraphics[width=0.6\textwidth]{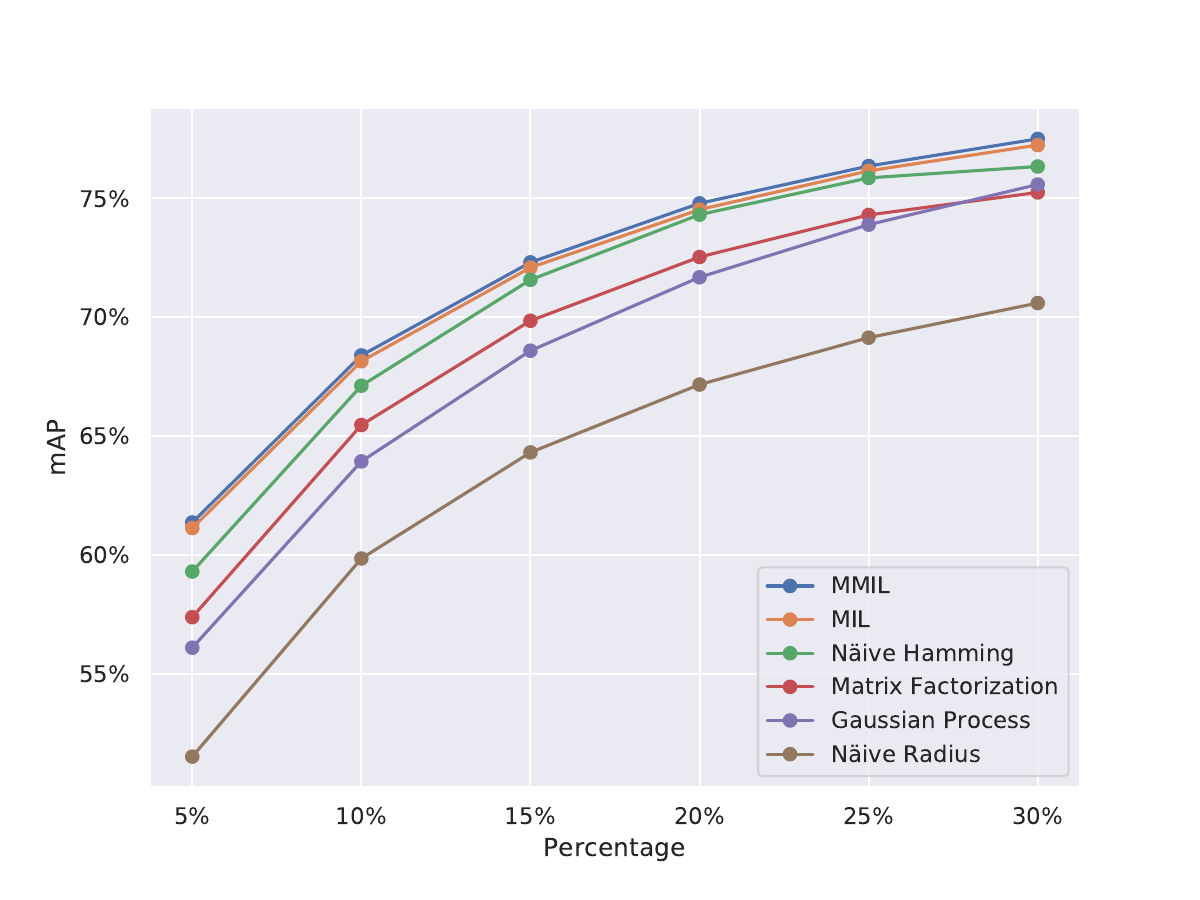}
  \caption{mAP for the all the methods and for all the datasets.}%
  \label{fig:map:botanic}
\end{figure}

From $A^P$ we created a MMIL dataset as follows: each region $i$ is a top-bag.
Each plant $j$ with $a^P_{ij}=1$ is a sub-bag of $i$. The instances contained in a sub-bag $j$ are
again regions: among all regions $i'$ with  $a^P_{i'j}=1$ we include in $j$ the 25 regions with minimal
Hamming distance to $i$.
We also created MIL dataset where we simply merge all the sub-bags of the MMIL dataset.
We compare both MIL and MMIL models against two N\"{a}ive models, Gaussian processes for binary classification~\citep{williams2006gaussian}, and matrix factorization~\citep{zhou2008large}.
\begin{itemize}
\item \textbf{N\"{a}ive 1}. For each region $i$, we first select the 25 closest regions ${k_1}, \ldots, {k_{25}}$ according to the Hamming distance, and then define $\hat{a}_{i *} = \frac{1}{25}\sum_{t=1}^{25} a_{k_t *}$.
\item \textbf{N\"{a}ive 2.}
  As   \textbf{N\"{a}ive 1}, but closest regions are selected according to Euclidean distance: all neighboring regions
  with a Euclidean distance below a certain threshold are selected, where the threshold is set such that most regions
  have approximately 25 neighbors.
  \item \textbf{Gaussian processes}, (also known as kriging when applied to geostatistical problems, see e.g.~\cite{oliver1990kriging}). For each different plant we trained a separate Gaussian process model using the approach described by~\cite{hensman2015scalable} and implemented by~\cite{gardner2018gpytorch} within an efficient PyTorch framework. The models take as input the coordinates (representing a region) and outputs a real value between 0 and 1, which indicates the probability of the existence of the plants for the given inputs. We used the Mercer kernel as results of an hyper-parameter search.
  \item \textbf{Matrix factorization}.
    Motivated by the collaborative filtering analogy, we also considered a matrix factorization approach
    computing
    \begin{displaymath}
      A^P \simeq U^T V =: \hat{A}
    \end{displaymath}
    with $U \in \{0,1\}^{12,665 \times k},V \in \{0,1\}^{k\times 842}$ for some $k$ learned by  minimizing
    a regularized sum of squared errors loss function. Hyperparameters $k$ and  $\lambda_1,\lambda_2$
    for the matrix norms of $U$ and $V$ in the regularization term were determined through grid search
    as  $k=50$, $\lambda_1=0$, $\lambda_2=0$.
  \item \textbf{MMIL and MIL networks}. We used a MMIL network model with two stacked bag-layers (only for the MIL network) with ReLU activations with 128 units and max as aggregation functions. On top of the bag-layers we used a dense layer with 128 units and ReLU activations, followed by the output dense layer with 842 units (corresponding to the number of plants) and sigmoid activations. Both the models are trained by minimizing the binary cross entropy loss for 100 epochs with Adam optimizer (learning rate $0.001$ until the 80th epoch and then $0.0001$) on batches of size 64.
\end{itemize}
Results with respect the mAP measure (see Equation \ref{eq:map}) are
depicted in Figure~\ref{fig:map:botanic}. We can notice that MMIL and MIL outperforms the other models.
Even though the difference between MMIL and MIL is small, it is consistent: we ran 6 repetitions of the experiment both for MMIL and MIL with different random initializations, and the best result of MIL was still worse than the worst result of MMIL.
Among the alternative methods the naive Hamming approach showed the strongest performance by far. For most percentage values $P$ of deleted species there is still a marked gap between naive Hamming and the MI approaches. It is notable that methods that use the spatial information (Gaussian process, naive radius) perform generally worse than methods only based on the occurrence matrix $A$.

\begin{figure}[htp]
  \centering
  \includegraphics[width=0.98\textwidth]{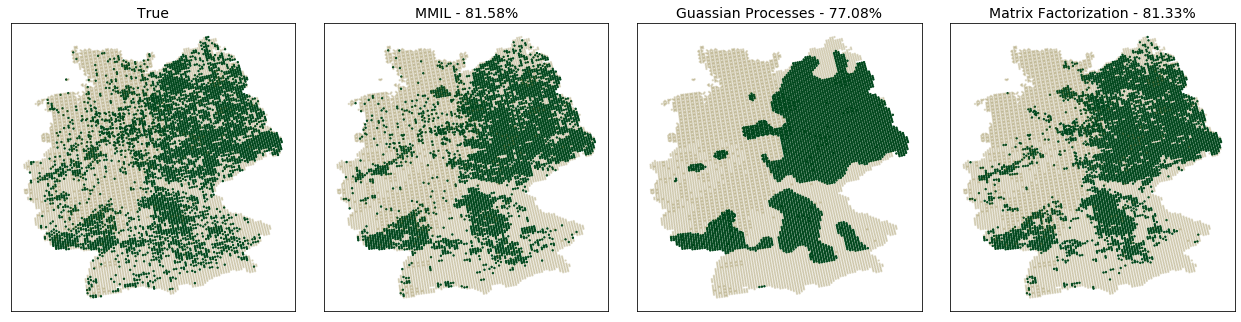}
  \caption{True and predicted distributions of Vicia Villosa.}%
  \label{fig:botanic:comparison}
\end{figure}

Even though our primary objective is ranking, we can also use all models as classifiers by setting a threshold for the scores $\hat{a}_{ij}$. For the following we selected one plant $j$=\emph{Vicia Villosa},
used the scores learned from $A^{5}$,
 and determined for each method an optimal threshold for  $\hat{a}_{ij}$ for predicting $a_{ij}=1$.
Figure \ref{fig:botanic:comparison} shows the true distribution and the distributions predicted by
MMIL, Gaussian processes, and matrix factorization.

We also use the classification problem for Vicia Villosa to illustrate the interpretability
of the MMIL model.
Using 1,000 regions as validation set we obtained an optimal number of 6 and 8 clusters for
sub-bags and instances, respectively. The decision tree had fidelity of $81.57\%$.

\begin{figure}[htp]
  \centering
  \includegraphics[width=0.7\textwidth]{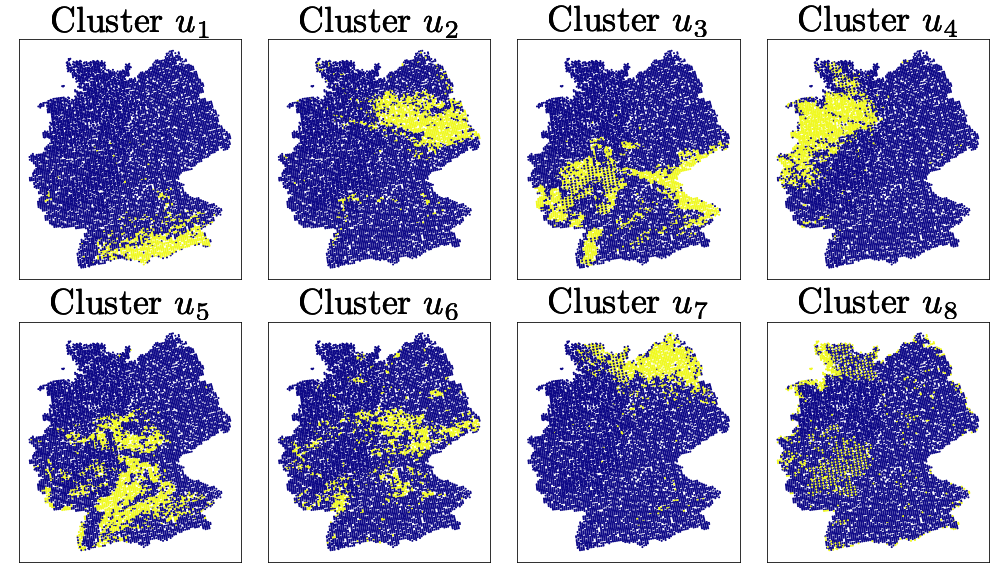}
  \parbox[b]{0.25\textwidth}{\begin{center}
    \includegraphics[width=0.24\textwidth]{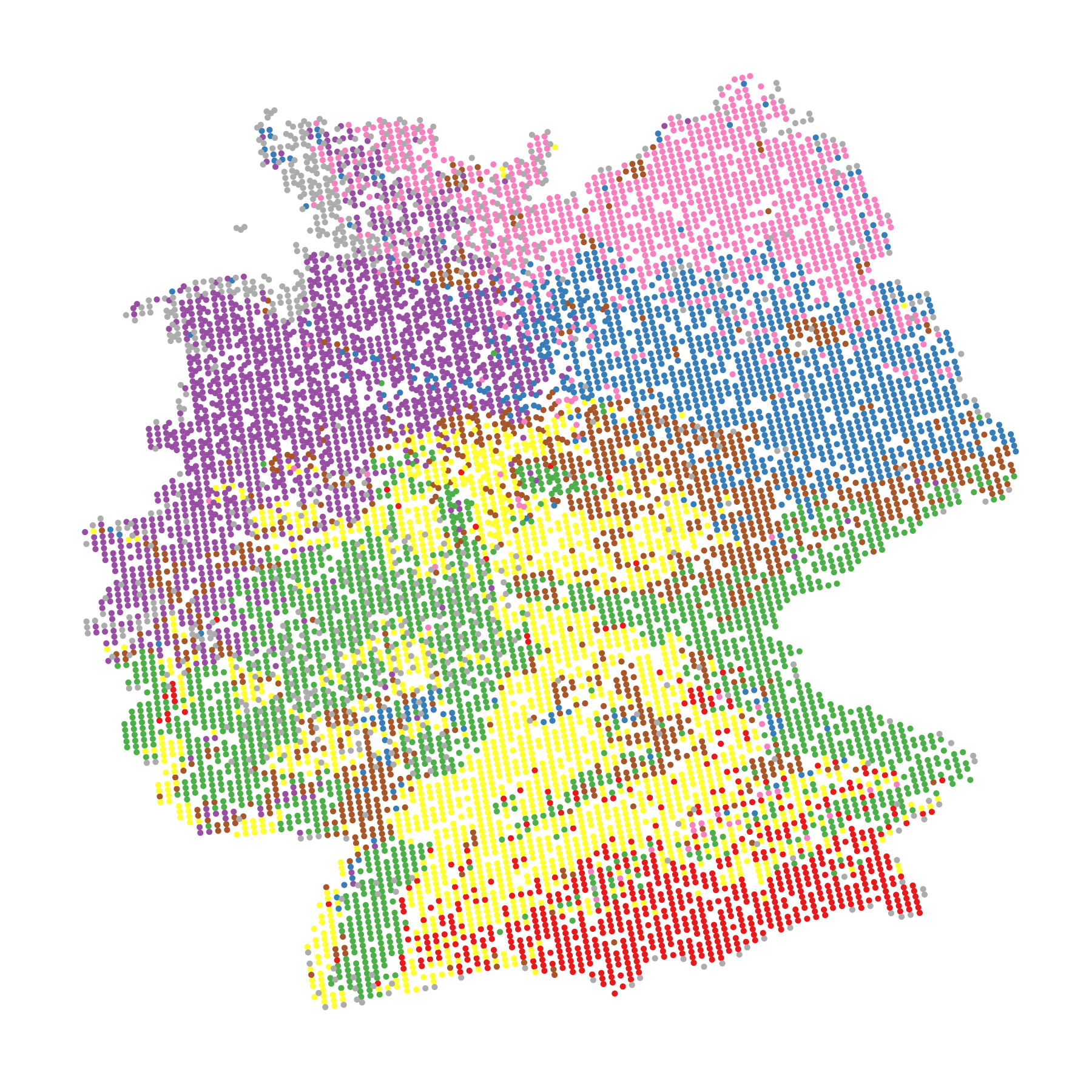}
    \includegraphics[width=0.24\textwidth]{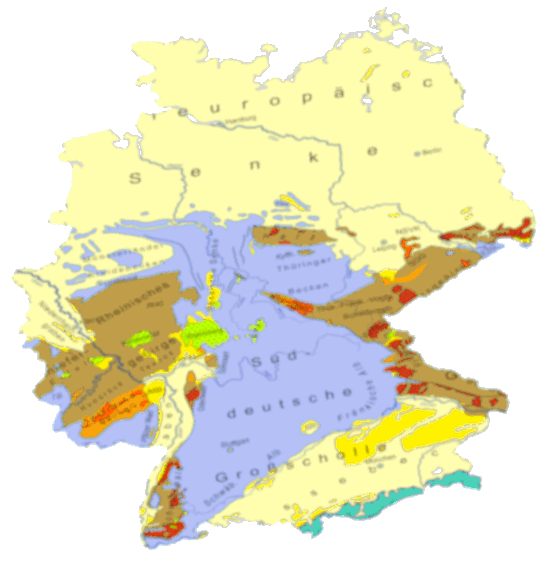}
    \vspace*{-5mm}
    \end{center}
  }
  \caption{Visualization of the instance clusters: left: regions (colored yellow) belonging to
    clusters $u_1$-$u_8$;  top right: combined visualization of partition defined by different clusters;
    bottom right: geological map of Germany (image adapted from \protect\url{https://commons.wikimedia.org/wiki/File:Geomap_Germany.png} used under creative commons license CC BY 4.0). }
  \label{fig:botanic:instance:pl}
\end{figure}

\begin{figure}[htp]
  \centering
  \includegraphics[width=0.7\textwidth]{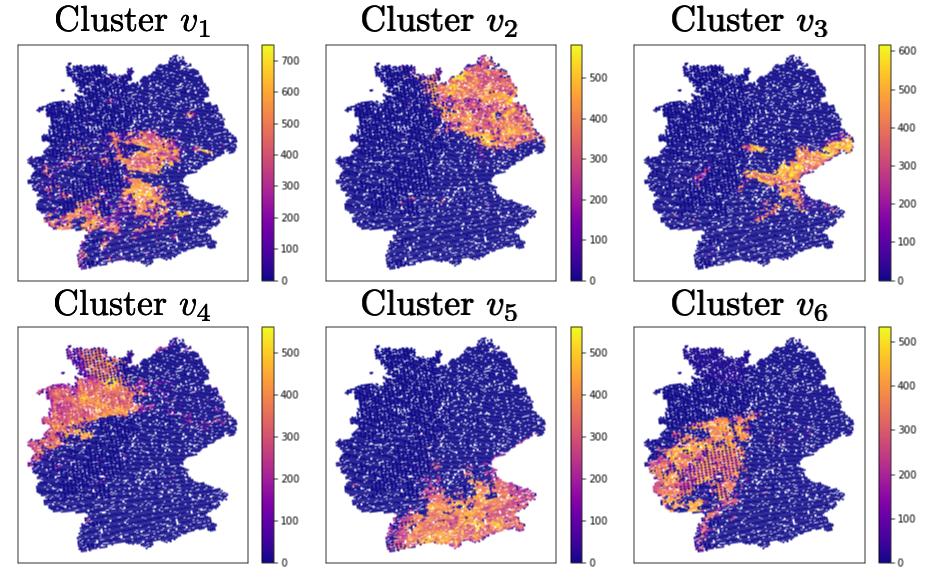}
  \caption{Visualization of the sub-bag clusters. Color is proportional to the number of species in each cluster.}%
  \label{fig:botanic:subbag:pl}
\end{figure}

Figure~\ref{fig:botanic:instance:pl} shows instance (region) clusters. A comparison with
a geological map of Germany reveals that clusters correspond quite closely to
distinct geological formations.
Figure~\ref{fig:botanic:subbag:pl} shows the geographical distribution of sub-bags clusters (i.e., plants).

\begin{figure}[htp]
  \centering

  \begin{small}
    % \begin{sc}
    \begin{tabular}{llll}
      \toprule
      $1$ & $v_{ 1 }$ & $\leftarrow$ & $f_{u_1}=0, f_{u_3}=0, f_{u_4}=0, f_{u_5}=1.$\\
      $2$ & $v_{ 2 }$ & $\leftarrow$ & $f_{u_1}=0, f_{u_3}=0, f_{u_4}=0, f_{u_5}=0.$\\
      $3$ & $v_{ 3 }$ & $\leftarrow$ & $f_{u_1}=0, f_{u_3}=1, f_{u_5}=0, f_{u_6}=1.$\\
      $4$ & $v_{ 3 }$ & $\leftarrow$ & $f_{u_1}=1,  f_{u_3}=1, f_{u_6}=1, f_{u_7}=0.$\\
      $5$ & $v_{ 4 }$ & $\leftarrow$ & $f_{u_1}=0,  f_{u_3}=0, f_{u_4}=1, f_{u_5}=0.$\\
      $6$ & $v_{ 5 }$ & $\leftarrow$ & $f_{u_1}=1,  f_{u_3}=0.$\\
      $7$ & $v_{ 5 }$ & $\leftarrow$ & $f_{u_1}=1,  f_{u_3}=1, f_{u_7}=1.$\\
      $8$ & $v_{ 5 }$ & $\leftarrow$ & $f_{u_1}=1,  f_{u_6}=0.$\\
      $9$ & $v_{ 5 }$ & $\leftarrow$ & $f_{u_1}=1,  f_{u_6}=1, f_{u_7}=1.$\\
      $10$ & $v_{ 6 }$ & $\leftarrow$ & $f_{u_1}=0,  f_{u_3}=0, f_{u_4}=1, f_{u_5}=1.$\\
      $11$ & $v_{ 6 }$ & $\leftarrow$ & $f_{u_1}=0,  f_{u_3}=1, f_{u_5}=1, f_{u_6}=1.$\\
      $12$ & $v_{ 6 }$ & $\leftarrow$ & $f_{u_1}=0,  f_{u_3}=1, f_{u_6}=0.$\\
      \bottomrule
    \end{tabular}
    % \end{sc}
  \end{small}

  \includegraphics[width=0.98\textwidth]{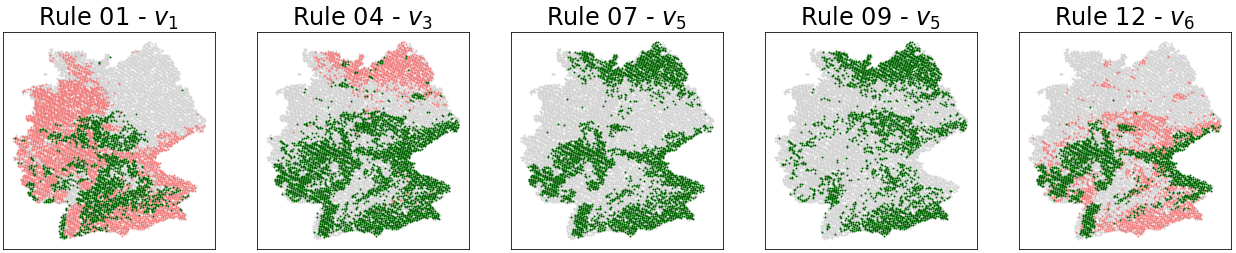}
  \caption{Top: Instance to sub-bag rules extracted from the MMIL network. Bottom: Visualization of instance to sub-bag rules $1$, $4$, $7$, $9$, and $11$.}%
  \label{fig:botanic:explain:sb}
\end{figure}

\begin{figure}[htp]
  \centering
  \begin{center}
    \begin{small}
      % \begin{sc}
      \begin{tabular}{llll}
        \toprule
        $1$ & $negative$ & $\leftarrow$ & $f_{v_1}=0, f_{v_2}=0, f_{v_3}=0.$\\
        $2$ & $negative$ & $\leftarrow$ & $f_{v_1}=0, f_{v_2}=0, f_{v_4}=0.$\\
        $3$ & $negative$ & $\leftarrow$ & $f_{v_1}=1, f_{v_2}=0, f_{v_5}=1, f_{v_6}=0.$\\
        $4$ & $negative$ & $\leftarrow$ & $f_{v_1}=1, f_{v_2}=0, f_{v_6}=1.$\\
        $5$ & $negative$ & $\leftarrow$ & $f_{v_2}=1, f_{v_3}=0, f_{v_4}=1, f_{v_6}=1.$\\
        $6$ & $positive$ & $\leftarrow$ & $f_{v_2}=1, f_{v_4}=0.$\\
        $7$ & $positive$ & $\leftarrow$ & $f_{v_1}=0, f_{v_2}=0, f_{v_3}=1, f_{v_4}=1.$\\
        $8$ & $positive$ & $\leftarrow$ & $f_{v_1}=1, f_{v_2}=0, f_{v_5}=0, f_{v_6}=0.$\\
        $9$ & $positive$ & $\leftarrow$ & $f_{v_2}=1, f_{v_3}=1, f_{v_4}=1.$\\
        $10$ & $positive$ & $\leftarrow$ & $f_{v_2}=1, f_{v_4}=1, f_{v_6}=0.$\\
        \bottomrule
      \end{tabular}
      % \end{sc}
    \end{small}
  \end{center}
  \includegraphics[width=0.98\textwidth]{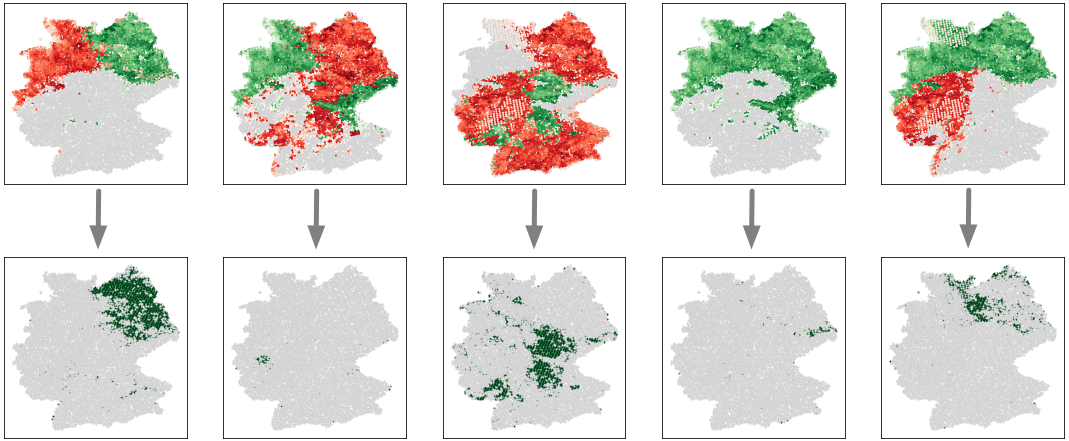}
  \caption{Instance to sub-bag rules extracted from the MMIL network.
    Visualization of sub-bag to label rules for the positive class.}%
  \label{fig:botanic:topbag:visualization}
\end{figure}

Instance to sub-bag and sub-bag to label rules are listed in
Figures~\ref{fig:botanic:explain:sb} and~\ref{fig:botanic:topbag:visualization}, respectively.
Figure~\ref{fig:botanic:explain:sb} illustrates some of the instance to
sub-bag rules.  Recall that each sub-bag (species) contains 25 instances
(regions). Rule 1, for example, says that a sub-bag has cluster identifier $v_1$,
if among these 25 regions there is at least one region in cluster $u_5$
(plotted in green), and none with in clusters $u_1,u_3,u_4$ (jointly plotted
in red).  Figure~\ref{fig:botanic:topbag:visualization} illustrates sub-bag to
label rules for the positive class. The top row corresponds to rules bodies
(color proportional to the number of positive literals minus the number of
negative literals); the bottom row shows regions that are classified as
positive by each rule. The overall prediction is the union of these regions.

\clearpage

\section{Conclusions}
\label{sec:conclusions}
We have introduced the MMIL framework for handling data organized
in nested bags.
The MMIL setting allows for a natural hierarchical
organization of data, where components at different levels of the hierarchy
are unconstrained in their cardinality. We have identified several learning problems that
can be naturally expressed as MMIL problems. For instance, image, text or graph
classification are promising application areas, because here the examples
can be objects of varying structure and size, for which a bag-of-bag data
representation is quite suitable, and can provide a natural
alternative to graph kernels or convolutional network for graphs.
Furthermore we proposed new way of thinking in terms of interpretability.
Although some MIL models can be easily interpreted by exploiting
the learnt instance labels and the assumed rule, MMIL
networks can be interpreted in a finer level:
by removing the common assumptions
of the standard MIL, we are more flexible and we can first associate
labels to instances and sub-bags and then combine them in order
to extract new rules.
Finally, we proposed a different perspective to see convolutions on graphs.
In most of the neural network for graphs approaches convolutions can be
interpreted as message passing schema, while in our approach we provided
a decomposition schema.

We proposed a neural network architecture involving the new construct of
bag-layers for learning in the MMIL setting. Theoretical results show the
expressivity of this type of model. In the empirical results we have shown that
learning MMIL models from data is feasible, and the theoretical capabilities of
MMIL networks can be exploited in practice, e.g., to learn accurate models for
noiseless data.
Furthermore MMIL networks can be applied in a wide spectrum of scenarios, such
as text, image, and graphs. For this latter we showed that MMIL is competitive
with the state-of-the-art models on node and graph classification tasks,
and, in many cases, MMIL models outperform the others.

In this paper, we have focused on the setting where
whole bags-of-bags are to be
classified. In conventional MIL learning, it is also
possible to define a task where
individual instances are to be classified. Such a task is however
less clearly defined in our setup since we do not assume to
know the label spaces at the instance and sub-bag level, nor the functional
relationship between the labels at the different levels.

\section*{Acknowledgement}
PF would like to acknowledge support for this project from the
  Italian Ministry of University and Research (MIUR grant 2017TWNMH2,
  RexLearn).  AT was with DINFO, Università di Firenze, when this work
  was initially submitted.

\clearpage
\appendix
\section{Details for the Experiments on Semi-Synthetic Data (Section 7.1)}
\label{apx:exp:mnist}

\begin{table}[htp]
\caption{Neural network structure for MMIL MNIST dataset. The model was trained by minimizing the binary cross entropy loss. We ran 200 epochs of the Adam optimizer \citep{kingma2014adam} with learning rate 0.001 and mini-batch size of 20. }
\label{tab:mnist:structure}
\vskip 0.15in
\begin{center}
\begin{small}
\begin{sc}
\begin{tabular}{ll}
\toprule
Layer & Parameters\\
\midrule
    Convolutional Layer  & kernel size $5\times 5$ with 32 channels \\
    Batch Normalization  &   \\
    ReLU  &    \\
    Max Pooling  & kernel size $2\times 2$ \\
    Dropout  & probability 0.5  \\
    Convolutional Layer  & kernel size $5\times 5$ with 64 channels \\
    Batch Normalization  &    \\
    ReLU  &    \\
    Max Pooling  & kernel size $2\times 2$  \\
    Dropout  & probability 0.5  \\
    Dense  & $1024$ units  \\
    ReLU  &    \\
    Dropout  & probability 0.5  \\
    BagLayer (ReLU activation) & $200$ units  \\
    ReLU  &    \\
    BagLayer (ReLU activation) & $200$ units  \\
    ReLU  &   \\
    Dense  & $1$ unit  \\
\bottomrule
\end{tabular}
\end{sc}
\end{small}
\end{center}
\vskip -0.1in
\end{table}

\begin{figure}[ht]
  \centering
  \includegraphics[width=0.99\textwidth]{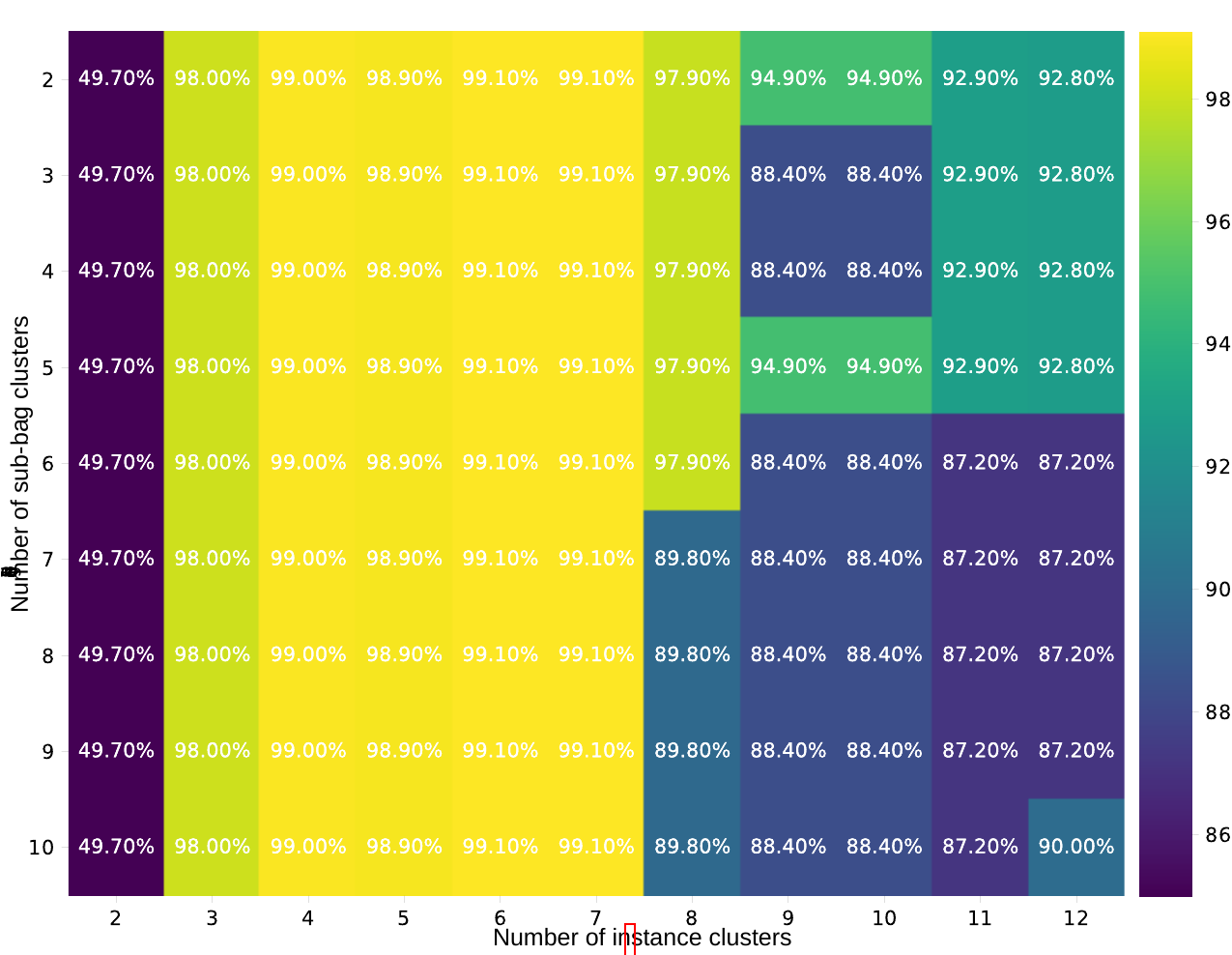}
  \caption{MNIST: Validation fidelity as a function of cluster sizes.}%
  \label{fig:mnist:faith}
\end{figure}
\clearpage

\section{Details for the Experiments on Sentiment Analysis (Section~\ref{sec:exp:imdb})}
\label{apx:imdb}
We report here a second example of classification explanation.
Here we are considering a positive review that was misclassified as
negative by the MMIL rules, and correctly classified by the MIL model.
Following the same typesetting conventions as used in Table~\ref{tab:sample},
the review and the labeling of the prediction-relevant parts are
shown in Table~\ref{tab:sample2}. In the MMIL case, classification
was  due to rule 6 in Table~\ref{tab:imdb2mmi}.
The sentence ``The storyline is $\dots$'' was assigned
cluster identifier $v_1$ by rule 1 in Table~\ref{tab:imdb1mmi},
whereas the sentence
``The mental patients$\dots$'' was assigned
cluster identifier $v_3$ by rule 6 in Table~\ref{tab:imdb1mmi}.
The positive classification in the
MIL case was due to rule 2 in Table~\ref{tab:imdb2mi}, which
is based on clusters $u_3$ and $u_6$.

\begin{table}[htp]
  \caption{A sample positive review. Top: MMIL labeling. Bottom: MIL labeling.}
  \label{tab:sample2}
  \begin{tcolorbox}[colback=white,sharp corners,boxrule=0.1mm]
    {\small
      \sf
          {\dimmed Young, ambitious nurse Ms. Charlotee (Rosie Holotik)
           is sent to work at a mental asylum out in the middle of nowhere.}
          {\dimmed During the course of 3 days, she encounters strange happenings,
            even a patient in her bedroom watching her, yet she still stays.}
          [$v_3$] The mental patients are all a little eye rolling
          (espically by the \textbf{Judge), but my favorite was}$^1$
          the old crazy biddy (Rhea MacAdams).

          [$v_1$] \textbf{The storyline is}$^4$\textbf{ okay at best}$^2$
          \textbf{, and}$^1$ the acting is \textbf{surprisingly
          alright, but}$^2$ after awhile it's gets to be
          \textbf{a little much}$^2$.
          {\dimmed But, still it's fun, quirky, strange, and original.
          xNote: The thing inside the basement is hardly horrifying,
          so the title is a little bananas.}
    }
  \end{tcolorbox}
  \begin{tcolorbox}[colback=white,sharp corners,boxrule=0.1mm]
    {\small
      \sf
           Young, ambitious nurse Ms. Charlotee (Rosie Holotik)
           is sent to work at a mental asylum out in the middle of nowhere.
           During the course of 3 days, she encounters strange happenings,
           even a patient in her bedroom watching her, yet she still stays.
          The mental patients are all a little eye rolling
          (espically by the \textbf{Judge), but my favorite was}$^6$
          the old crazy biddy (Rhea MacAdams).

          \textbf{The storyline is okay}$^3$\textbf{  at best, and}$^6$
          the \textbf{acting is surprisingly}$^6$
          alright, but after awhile it's gets to be
          a little much.
          \textbf{But, still it's fun, quirky, strange}$^6$, and original.
          xNote: The thing inside the basement is hardly horrifying,
          so the title is a little bananas.
    }
  \end{tcolorbox}
\end{table}

\begin{figure}[ht]
  \centering
  \includegraphics[width=0.99\textwidth]{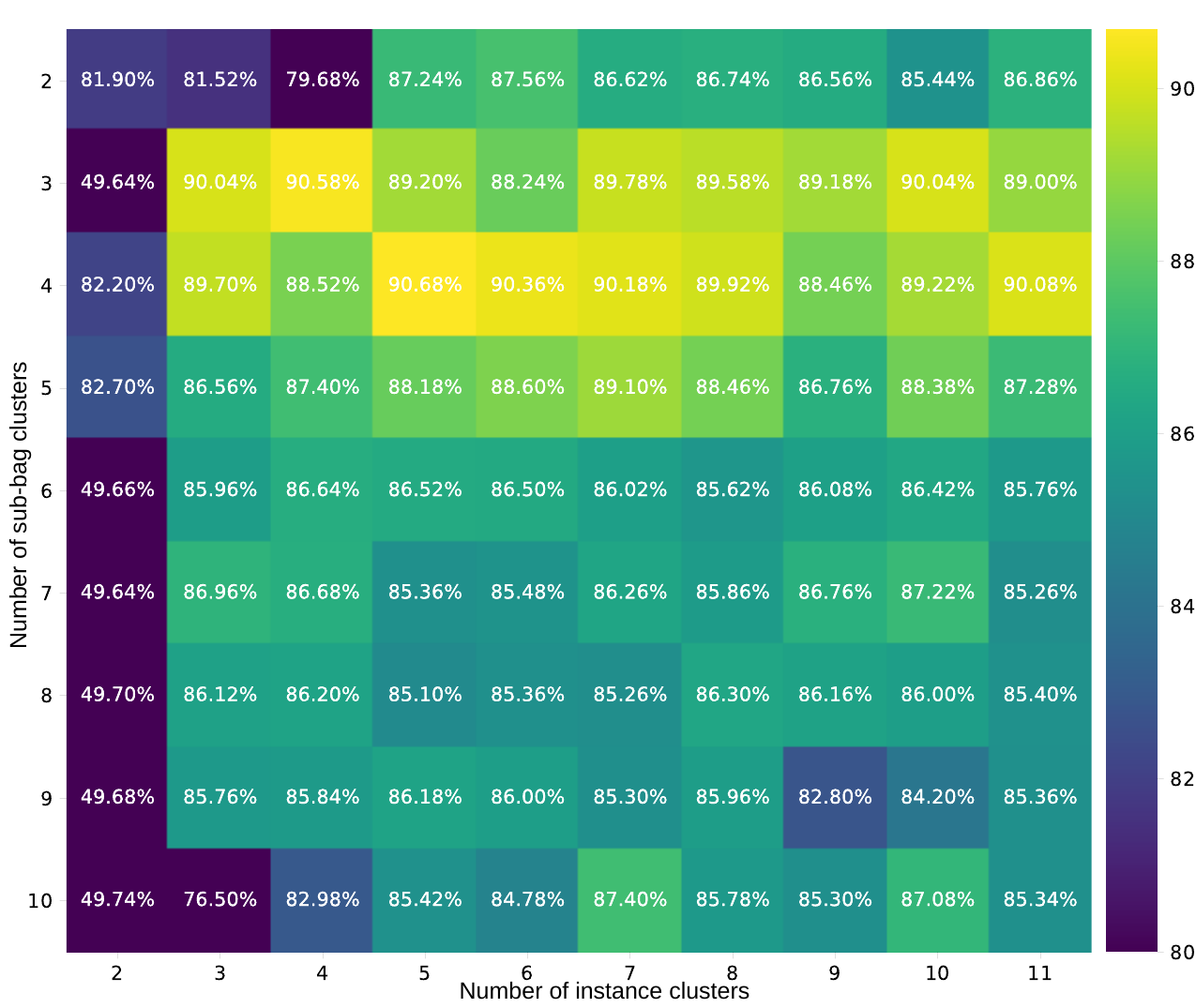}
  \caption{IMDB: Validation fidelity as a function of cluster sizes.}%
  \label{fig:imdb:faith}
\end{figure}

\section{Details for the Experiments on Citation Networks Data (Section~\ref{sec:cit-datasets})}
\label{apx:citation}

\begin{figure}[htb]
  \centering
  \includegraphics[width=.99\textwidth]{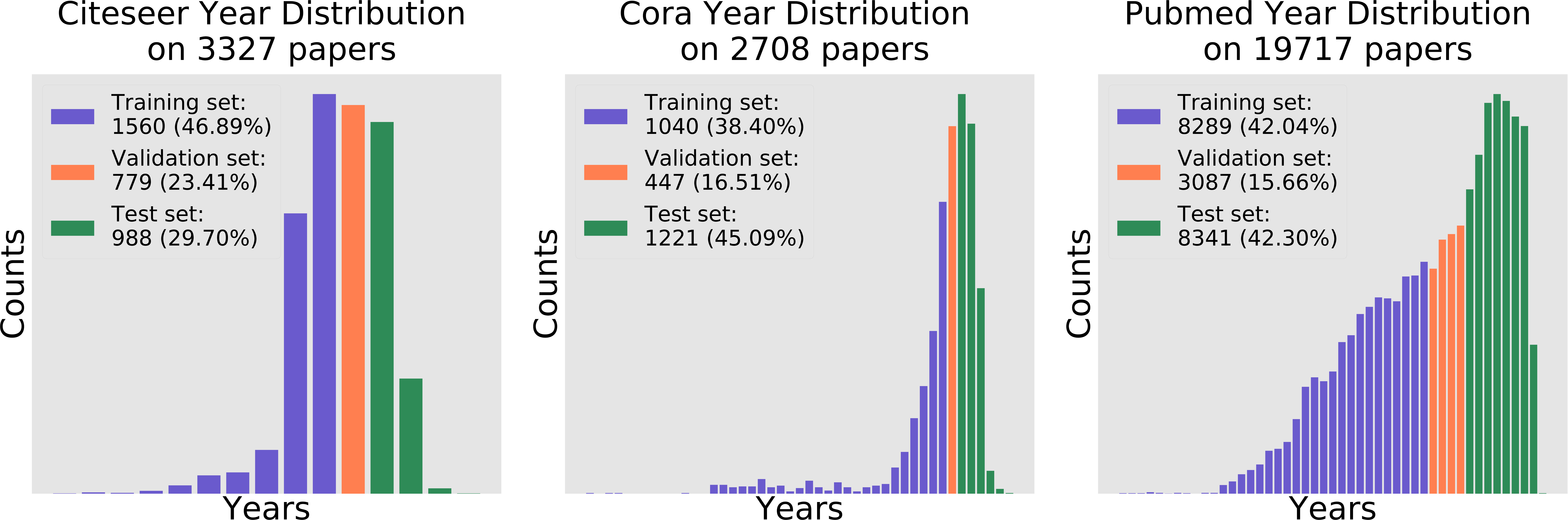}
  \caption{Distribution of papers over years for Citaseer, Cora, and PubMed. 
  }%
  \label{fig:citation:dist}
\end{figure}

\begin{figure}[ht]
  \centering
  \includegraphics[width=0.99\textwidth]{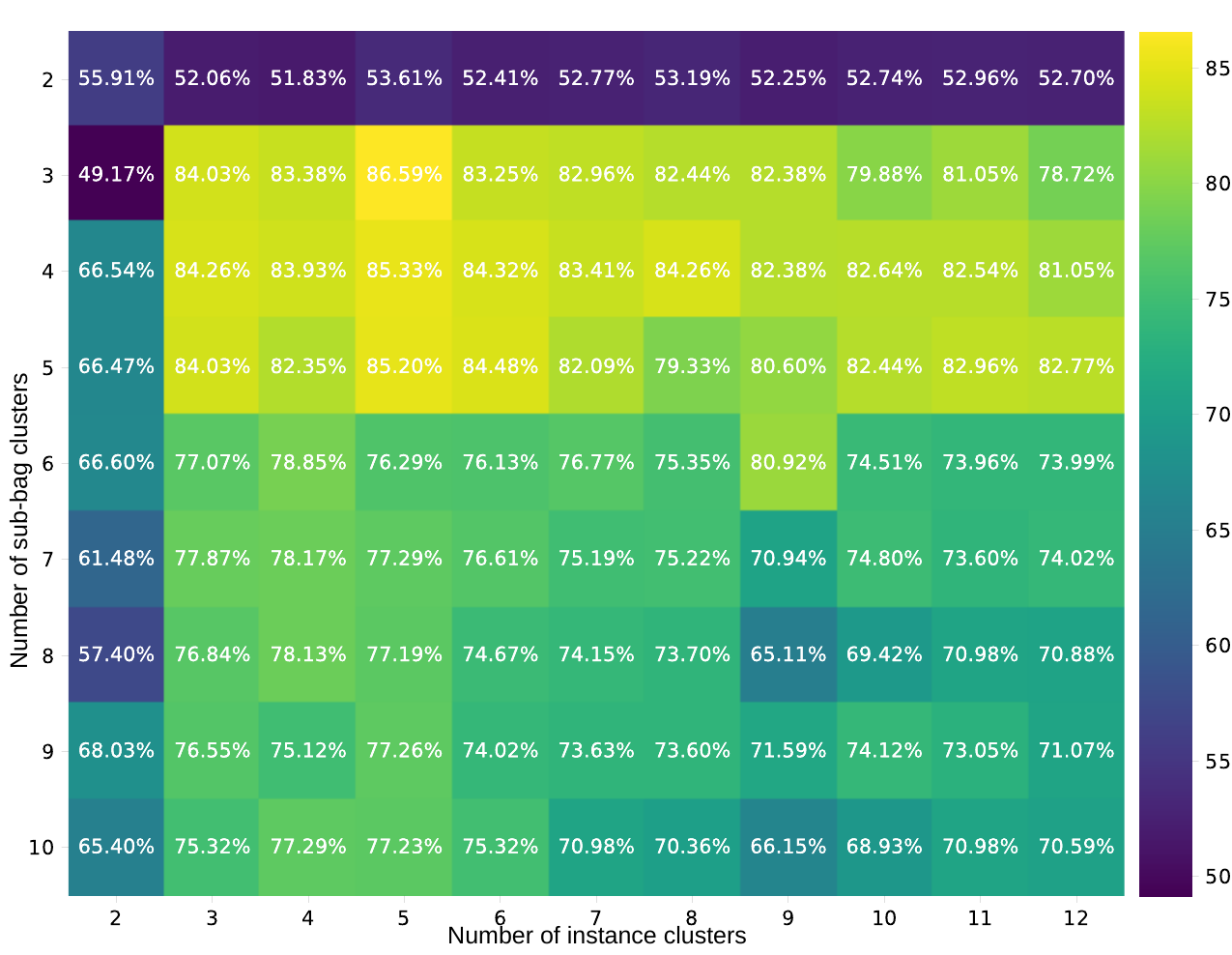}
  \caption{Pubmed: Validation fidelity as a function of cluster sizes.}%
  \label{fig:pubmed:faith}
\end{figure}

Here we interpret the MMIL-Mean and MIL-Mean models trained on the PubMed
citation dataset (Section~\ref{sec:cit-datasets}).  In the MMIL setting, the
optimal number of sub-bag and instance clusters in the validation set were
three and five, respectively (see Figure~\ref{fig:pubmed:faith}). In the MIL
setting the optimal number of instance clusters in the validation set was
three.

\begin{figure}[htb]
  \centering
  \includegraphics[width=0.99\textwidth]{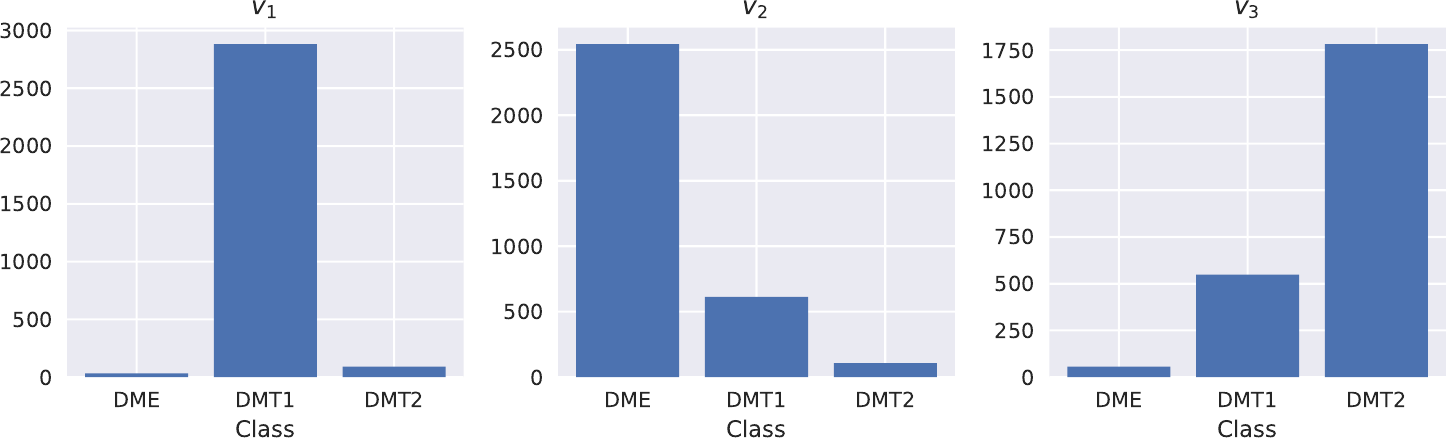}
  \caption{Correspondence between sub-bag clusters and actual paper class labels.}%
  \label{fig:pubmed:acts}
\end{figure}

Sub-bags in the
MMIL decomposition also are papers, and therefore also are
labeled with the actual class label.  The number
of inferred sub-bag clusters matches the true number of
classes, and, as shown in Figure~\ref{fig:pubmed:acts}, clusters and classes strongly correlate.
Instance (words) clusters are interpreted using
the same approach as in Section~\ref{sec:exp:imdb}. Result are shown
in Table~\ref{tab:citation:words:mmi:pl} for the MMIL setting and in Table~\ref{tab:citation:words:mi:pl} for the MIL setting.

\begin{table}[htp]
  \caption{PubMed: MMIL rules for mapping instance cluster frequencies into a
    sub-bag cluster identifiers.  See the caption of Table~\ref{tab:imdb1mmi} for
    details on the syntax.}
       \label{tab:cit2mmi}
  $$
  \begin{array}{ll}
    \toprule
    1&v_1 \leftarrow \PAR{f_{u_2}\mathord{\le}44.53},\PAR{f_{u_3}\mathord{\le}22.70},\PAR{f_{u_5}\mathord{\le}6.47}.\\
    2&v_1 \leftarrow \PAR{f_{u_2}\mathord{\le}44.53},\PAR{f_{u_3}\mathord{\le}31.31},\PAR{f_{u_5}\mathord{>}6.47}.\\
    3&v_1 \leftarrow \PAR{f_{u_2}\mathord{>}44.53},\PAR{f_{u_4}\mathord{\le}14.22},\PAR{f_{u_5}\mathord{>}14.83}.\\
    4&v_2 \leftarrow \PAR{f_{u_2}\mathord{>}44.53},\PAR{f_{u_3}\mathord{\le}22.60},\PAR{f_{u_4}\mathord{>}14.22}.\\
    5&v_2 \leftarrow \PAR{f_{u_2}\mathord{>}44.53},\PAR{f_{u_4}\mathord{\le}14.22},\PAR{f_{u_5}\mathord{\le}14.83}.\\
    6&v_3 \leftarrow \PAR{f_{u_2}\mathord{\le}44.53},\PAR{f_{u_3}\mathord{>}22.70}, \PAR{f_{u_5}\mathord{\le}6.47}.\\
    7&v_3 \leftarrow\PAR{f_{u_2}\mathord{\le}44.53}, \PAR{f_{u_3}\mathord{>}31.31},\PAR{f_{u_5}\mathord{>}6.47}.\\
    8&v_3 \leftarrow\PAR{f_{u_2}\mathord{>}44.53}, \PAR{f_{u_3}\mathord{>}22.60},\PAR{f_{u_4}\mathord{>}14.22}.\\
    \bottomrule
  \end{array}
  $$
\end{table}

Rules extracted by our procedure are reported in Tables~\ref{tab:cit2mmi},
\ref{tab:cit1mmi} (MMIL) and~\ref{tab:cit1mi} (MIL).  Test set accuracies of
the extracted rules were $76.88\%$ and $79.25\%$ in the MMIL and MIL setting,
respectively. The corresponding fidelities were $84.75\%$ and $87.99\%$,
respectively. Both of the results are still comparable and competitive with
the methods described in Table \ref{tab:citation:res}.  Thus, in this case the
interpretable MIL model outperforms the interpretable MMIL model in terms of
accuracy.  However, for explaining individual classifications, the MMIL model
can still have advantages due to the multi-level explanations it supports. As
we did for the IMDB experiment (Section~\ref{sec:exp:imdb}), one can first
explain the predicted label of a paper in terms of the citing/cited papers
assigned to the relevant clusters, and then refine this explanation by tracing
paper cluster assignments to word clusters. In the MIL model, on the other
hand, only word level explanations are possible.

\begin{table}[htp]
  \caption{PubMed: MMIL rules mapping sub-bag cluster frequencies
    into top-bag labels. See the caption of
    Table~\ref{tab:imdb1mmi} for details on the syntax.}
    \label{tab:cit1mmi}
  $$
  \begin{array}{ll}
    \toprule
    1&DME~ \leftarrow \PAR{f_{v_1}\mathord{\le}8.51},\PAR{f_{v_2}\mathord{>}63.96}.\\
    2&DMT1 \leftarrow \PAR{f_{v_1}\in (8.51,20.26]},\PAR{f_{v_2}\mathord{>}63.96}.\\
    3&DMT1 \leftarrow \PAR{f_{v_1}\mathord{>}20.26},\PAR{f_{v_3}\mathord{\le}55.49}.\\
    4&DMT2 \leftarrow \PAR{f_{v_1}\mathord{\le}20.26},\PAR{f_{v_2}\mathord{\le}63.96}.\\
    5&DMT2 \leftarrow \PAR{f_{v_1}\mathord{>}20.26},\PAR{f_{v_3}\mathord{>}55.49}.\\
    \bottomrule
  \end{array}
  $$
\end{table}

\begin{table}[htp]
  \caption{PubMed: MIL classification rules. See the caption of
    Table~\ref{tab:imdb1mmi} for details on the syntax.}
    \label{tab:cit1mi}
  $$
  \begin{array}{ll}
    \toprule
    1&DME~ \leftarrow \PAR{f_{u_1}\mathord{>}49.80}.\\
    2&DMT1 \leftarrow \PAR{f_{u_1}\mathord{\le}49.80}, \PAR{f_{u_2}\mathord{\le}30.60},\PAR{f_{u_3}\mathord{\le}30.65}.\\
    3&DMT1 \leftarrow \PAR{f_{u_1}\mathord{\le}49.80},\PAR{f_{u_2}\mathord{>}30.60},\PAR{f_{u_3}\mathord{\le}35.66}.\\
    4&DMT1 \leftarrow \PAR{f_{u_1}\mathord{\le}49.80}, \PAR{f_{u_2}\mathord{\le}30.60},\PAR{f_{u_3}\mathord{>}30.65}.\\
    5&DMT1 \leftarrow \PAR{f_{u_1}\mathord{\le}49.80},\PAR{f_{u_2}\mathord{>}30.60},\PAR{f_{u_3}\mathord{>}35.66}.\\
    \bottomrule
  \end{array}
  $$
\end{table}

\begin{table}[htp]
   \caption{PubMed: Clusters in the MIL case. Each column represents words belonging to the associated cluster. The percentage next to each cluster identifier refers to the number of
     of words associated with that cluster ($\approx 15$k). Words are ranked by intra-cluster distance in descending order.}
   \label{tab:citation:words:mi:pl}
\vskip 0.15in
\begin{center}
\begin{small}
%\begin{sc}
\begin{tabular}{lll}
\toprule
\multicolumn{1}{c}{ $u_1$ $(48.33\%)$} & \multicolumn{1}{c}{$u_2$ $(60.09\%)$} & \multicolumn{1}{c}{$u_3$ $(41.92\%)$ } \\
\midrule
     animals & children & non \\
     induction & juvenile & subjects \\
     induced & multiplex & patients \\
     experimental & hla & indians \\
     rats & childhood & fasting \\
     rat & adolescents & obesity \\
     dogs & conventional & pima \\
     caused & girls & american \\
     days & ascertainment & mexican \\
     strains & autoimmune & indian \\
     bl & dr & mody \\
     experiment & infusion & oral \\
     untreated & child & bmi \\
     wk & siblings & obese \\
     sz & intensive & men \\
     restored & healthy & prevalence \\
     sciatic & paediatric & resistance \\
     experimentally & spk & tolerance \\
     sprague & boys & mutations \\
     partially & sharing & igt\\
\bottomrule
\end{tabular}
%\end{sc}
\end{small}
\end{center}
\vskip -0.1in
\end{table}

\begin{table}[htp]
  \caption{PubMed: Clusters in the MMIL case. Each column represents words
    belonging to the associated cluster. The percentage next to the cluster
    identifier refers to the number of of words associated with that cluster
    ($\approx 15$k).  Words in cluster $u_1$ are grayed out since that cluster is never used
    for constructing the rules. Words are ranked by intra-cluster
    distance in descending order.}
   \label{tab:citation:words:mmi:pl}
\vskip 0.15in
\begin{center}
\begin{small}
%\begin{sc}
\begin{tabular}{lllll}
\toprule
\multicolumn{1}{c}{ $u_1$ $(21.28\%)$} & \multicolumn{1}{c}{$u_2$ $(28.76\%)$} & \multicolumn{1}{c}{$u_3$ $(27.25\%)$ } & \multicolumn{1}{c}{$u_4$ $(12.84\%)$} & \multicolumn{1}{c}{$u_5$ $(9.87\%)$ }\\
\midrule
     {\leavevmode\dimmed  normalization} & animals & non & subjects & children \\
     {\leavevmode\dimmed  greatly} & experimental & indians & patients & multiplex \\
     {\leavevmode\dimmed  susceptibility} & induced & pima & patient & ascertainment \\
     {\leavevmode\dimmed  lymphocytes} & induction & obesity & individuals & conventional \\
     {\leavevmode\dimmed  pregnant} & rats & oral & type & juvenile \\
     {\leavevmode\dimmed  always} & dogs & fasting & analysis & girls \\
     {\leavevmode\dimmed  organ} & made & mexican & sample & night \\
     {\leavevmode\dimmed  destruction} & rat & obese & cascade & childhood \\
     {\leavevmode\dimmed  tx} & strains & medication & otsuka & pittsburgh \\
     {\leavevmode\dimmed  contraction} & bl & bmi & forearm & adolescents \\
     {\leavevmode\dimmed  antibodies} & caused & mody & gdr & infusion \\
     {\leavevmode\dimmed  sequential} & wk & indian & reported & denmark \\
     {\leavevmode\dimmed  tract} & counteracted & tolerance & mmol & intensified \\
     {\leavevmode\dimmed  decarboxylase} & partially & look & age & child \\
     {\leavevmode\dimmed  recipients} & rabbits & index & gox & beef \\
     {\leavevmode\dimmed  livers} & days & agents & dependent & sharing \\
     {\leavevmode\dimmed  mt} & conscious & resistance & isoforms & knowing \\
     {\leavevmode\dimmed  cyclosporin} & sciatic & maturity & meals & paediatric \\
     {\leavevmode\dimmed  lv} & tubules & gk & score & unawareness \\
     {\leavevmode\dimmed  laboratories} & myo & ii & affinities & pubert \\
\bottomrule
\end{tabular}
%\end{sc}
\end{small}
\end{center}
\vskip -0.1in
\end{table}

\clearpage

\section{Details for the Experiments on Point Clouds (Section~\ref{ex:point:clouds})}\label{apx:point:cloud}
\begin{figure}[ht]
  \centering
  \includegraphics[width=0.99\textwidth]{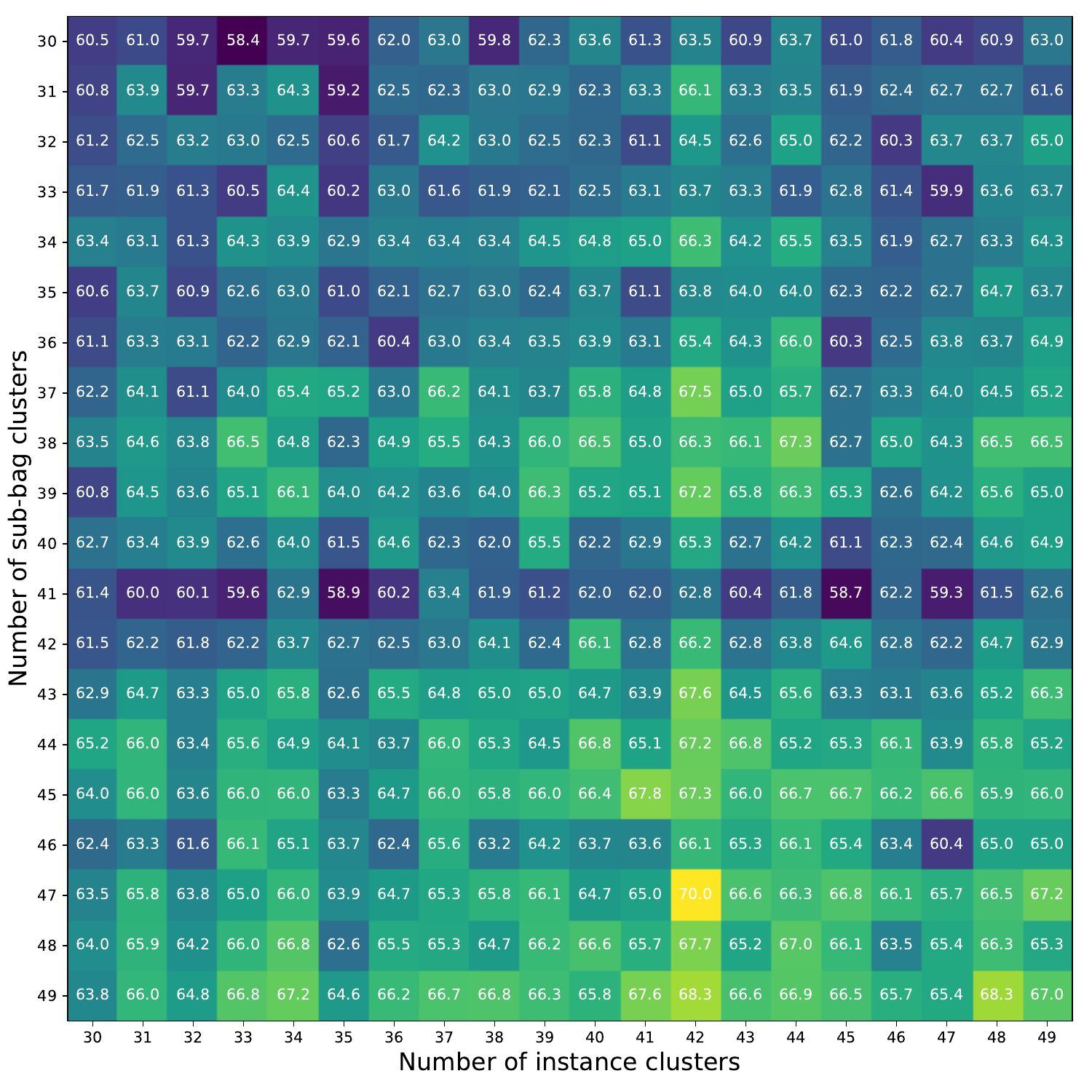}
  \caption{PointCloud: Validation fidelity as a function of cluster sizes.}%
  \label{fig:point:cloud:faith}
\end{figure}

Like in Section~\ref{sec:semi-mnist}, we derived interpreting rules in the
MMIL setting on the P100 dataset.
Using
2,000 point clouds as a validation set, we obtained 47 and 42
clusters
for sub-bags and instances, respectively.
For the rules mapping instance cluster identifiers to subbag cluster identifiers the
decision tree was used as propositional learner (as the aggregation function
in the first Bag-Layer is the $\max$), while for the rules mapping subbag
cluster identifiers to topbag labels the decision tree considered the counts of
subbag identifiers (as the aggregation function in the second Bag-Layer is the
sum).  Full grid search results on the validation set are reported in
Figure~\ref{fig:point:cloud:faith}. Accuracy
using rules was $59.12\%$, corresponding to a fidelity of $61.83\%$.

\begin{figure}[ht]
  \centering
  \includegraphics[width=0.99\textwidth]{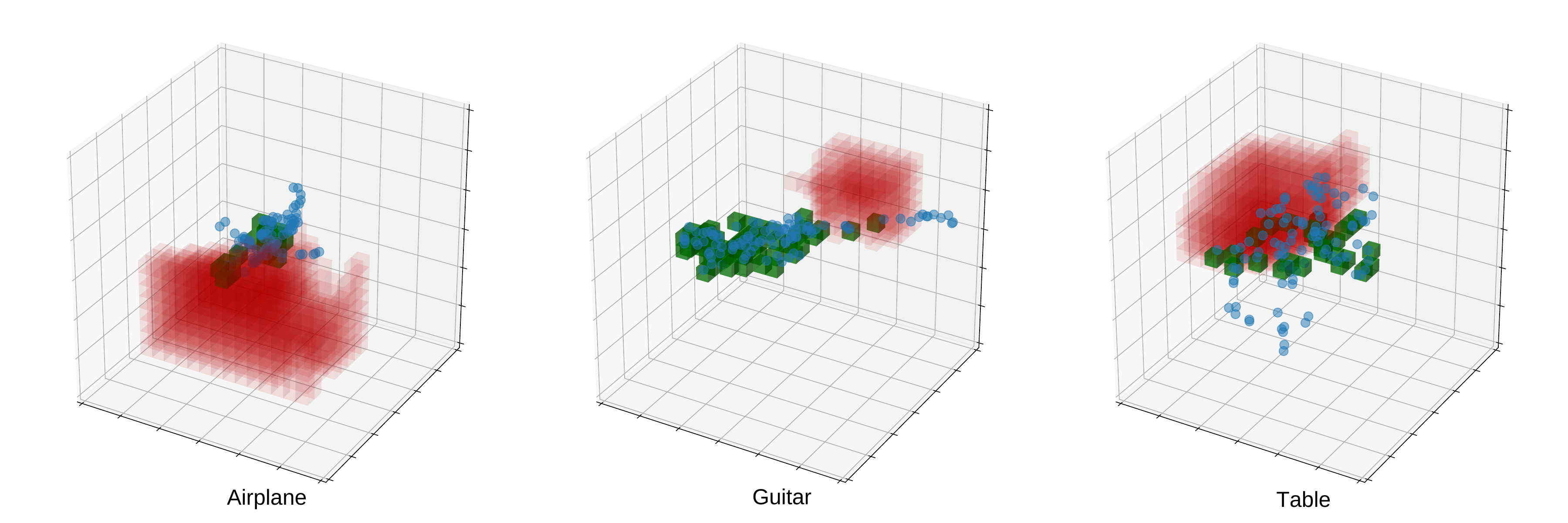}
  \caption{Three point clouds examples. The green and red boxes represents the active and inactive regions, respectively for the corresponding clusters.}%
  \label{fig:point_cloud_3_examples}
\end{figure}

Some examples are shown
in Figure~\ref{fig:point_cloud_3_examples}.
Subbag cluster identifiers (see Table~\ref{tab:rule:inst:to:sub:point:cloud})
$v_5$, $v_{31}$, $v_4$ correspond to \textit{airplane}, the \textit{guitar}
and the \textit{table}, respectively. The blue points represent the original
point clouds. For readability, in Figure~\ref{fig:point_cloud_3_examples} we only distinguish between regions
with active (green) and inactive (red) instance
clusters in the rules of
Table~\ref{tab:rule:inst:to:sub:point:cloud}.

\begin{figure}[ht]
  \centering
  \includegraphics[width=0.94\textwidth]{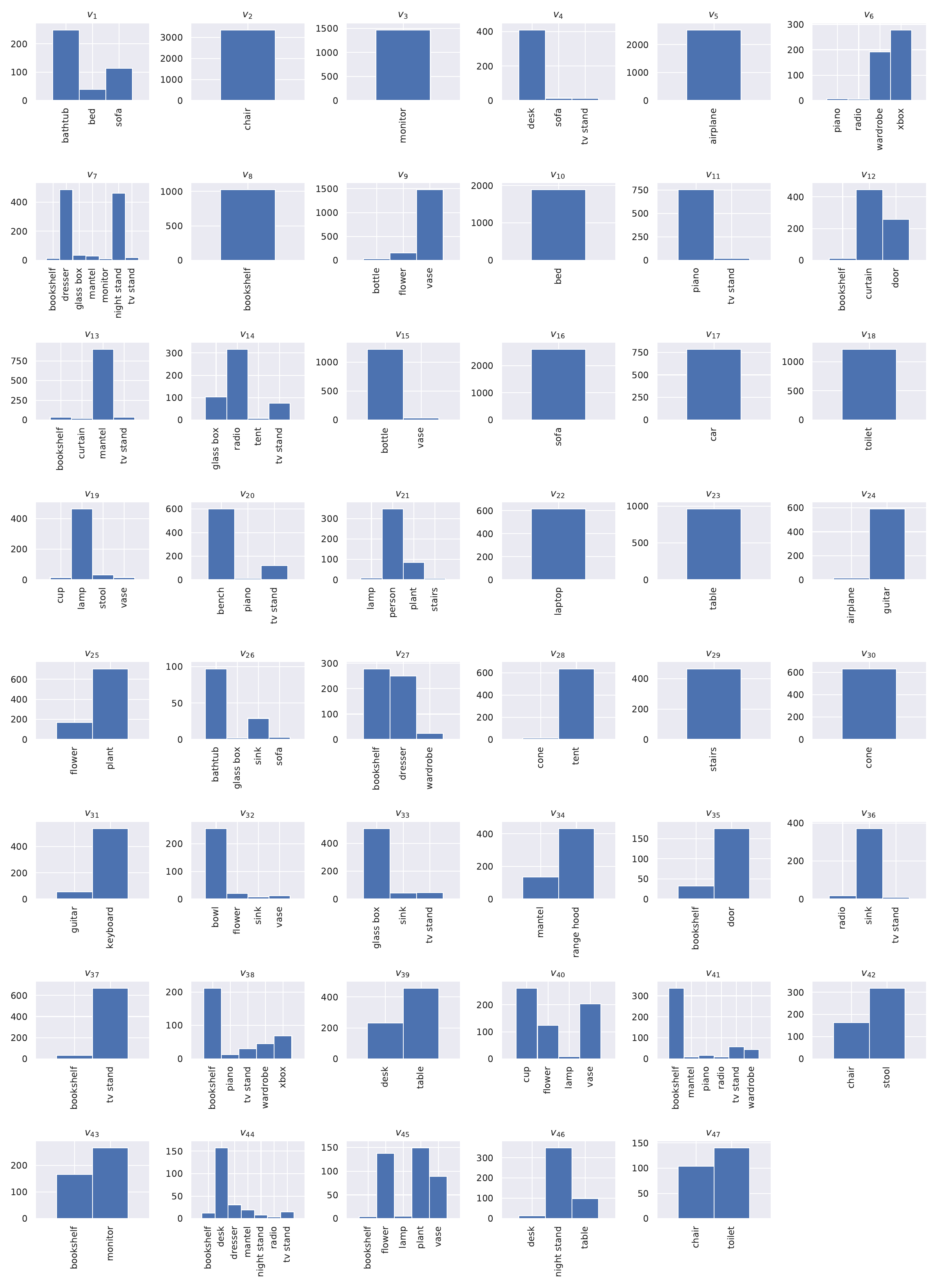}
  \caption{Correspondence between subbag cluster identifiers and actual point cloud class labels. Class labels without correspondence are omitted.}%
  \label{fig:point:cloud:examples:subbag:pl}
\end{figure}

\paragraph{MMIL rules.} Using a decision tree learner taking
 binary vectors $(f_{u_1},\ldots,f_{u_{42}})$ as inputs,
we obtained 959 rules for mapping
a bag $\SubBag{j}$ of instance cluster identifiers to a sub-bag cluster identifiers and
85 rules for mapping sub-bag cluster identifiers to the top-bag class labels.
We do not report the full sets of rules here. However we report in Table~\ref{tab:rule:inst:to:sub:point:cloud} the rules corresponding to the three point clouds depicted in Figure~\ref{fig:point_cloud_3_examples}. In particular $v_5$, $v_{31}$ and $v_4$ are the sub-bag clusters associated with classes \textit{airplane}, \textit{guitar}, and \textit{table}, respectively.
\begin{table}[t]
  \caption{Rules extracted from the MMIL network
    for mapping instance cluster identifiers into a sub-bag
    cluster identifiers. }
  \label{tab:rule:inst:to:sub:point:cloud}
\vskip 0.15in
\begin{center}
\begin{small}
%\begin{sc}
\begin{tabular}{llll}
\toprule
\multirow{2}[1]{*}{$1$} & \multirow{2}[1]{*}{$v_{ 5 }$} & \multirow{2}[1]{*}{$\leftarrow$} & $f_{ u_{ 9 } } = 1, \ f_{ u_{ 13 } } = 1, \ f_{ u_{ 14 } } = 1, \ f_{ u_{ 18 } } = 1, \ f_{ u_{ 15 } } = 0,$\\
& & & $f_{ u_{ 16 } } = 0, \ f_{ u_{ 21 } } = 0, \ f_{ u_{ 35 } } = 0, \ f_{ u_{ 39 } } = 0, \ f_{ u_{ 41 } } = 0.$\\
\cmidrule{1-4}
\multirow{2}[1]{*}{$2$} & \multirow{2}[1]{*}{$v_{ 31 }$} & \multirow{2}[1]{*}{$\leftarrow$} & $f_{ u_{ 13 } } = 1, \ f_{ u_{ 15 } } = 1, f_{ u_{ 32 } } = 1, \ f_{ u_{ 33 } } = 1, f_{ u_{ 4 } } = 0,$\\
& & & $f_{ u_{ 5 } } = 1, \ f_{ u_{ 9 } } = 0,  \ f_{ u_{ 18 } } = 0, \ f_{ u_{ 31 } } = 0.$\\
\cmidrule{1-4}
\multirow{2}[1]{*}{$3$} & \multirow{2}[1]{*}{$v_{ 4 }$} & \multirow{2}[1]{*}{$\leftarrow$} & $f_{ u_{ 12 } } = 1, \ f_{ u_{ 13 } } = 1, \ f_{ u_{ 15 } } = 1, \ f_{ u_{ 41 } } = 1, \ f_{ u_{ 30 } } = 1,$\\
& &  & $f_{ u_{ 31 } } = 0, \ f_{ u_{ 32 } } = 0, \ f_{ u_{ 35 } } = 1, \ f_{ u_{ 37 } } = 0.$\\
\bottomrule
\end{tabular}
%\end{sc}
\end{small}
\end{center}
\vskip -0.1in
\end{table}
For the left example (\textit{airplane}) in Figure~\ref{fig:point_cloud_3_examples}, all subbags are in cluster $v_5$. For the middle example (\textit{guitar}) two subbag are in $v_{31}$ and three in $v_{24}$. For the right example (\textit{table}) all subbags are in $v_4$. By inspecting Figure~\ref{fig:point:cloud:examples:subbag:pl} we can have an immediate intuition for the predicted topbag labels. In fact, $v_5$ exclusively correlates with \textit{airplanes}, $v_{31}$ correlates mainly with \textit{keyboards} and sometimes with \textit{guitars}, $v_{24}$ correlates most exclusively with \textit{guitars}, and $v_4$ correlates most exclusively with \textit{desks}. We would then expect that the left and middle examples are correctly classified as \textit{airplane} and \textit{guitar}, respectively, while the right example is misclassified as  \textit{desk} (the correct label was  \textit{table}). Finally, Table~\ref{tab:rule:sub:to:top:point:cloud} shows the rules (which confirm the intuition) that connects the subbag cluster identifiers to topbag labels for the examples depicted in Figure~\ref{fig:point_cloud_3_examples}.

\begin{table}[t]
  \caption{Rules extracted from the MMIL network
    for mapping subbag cluster identifiers into the top-bag
    label. }
  \label{tab:rule:sub:to:top:point:cloud}
\vskip 0.15in
\begin{center}
\begin{small}
%\begin{sc}
\begin{tabular}{llll}
\toprule
$1$ & $l_{ airplane }$ & $\leftarrow$ & $f_{ v_{5} } > 3, f_{ v_{2} } \le 1, f_{ v_{16} } \le 2.$ \\
\cmidrule{1-4}
\multirow{2}[1]{*}{$2$} & \multirow{2}[1]{*}{$l_{ guitar }$} & \multirow{2}[1]{*}{$\leftarrow$} & $f_{ v_{24} } > 2, f_{ v_{2} } \le 1, f_{ v_{3} } \le 1, f_{ v_{5} } \le 2, f_{ v_{9} } \le 1, f_{ v_{x10} } \le 1, f_{ v_{11} } \le 1, f_{ v_{13} } \le 1, f_{ v_{15} } \le 2, $\\
& & &$f_{ v_{16} } \le 2,f_{ v_{17} } \le 3, f_{ v_{18} } \le 1, f_{ v_{8} } = 0, f_{ v_{23} } = 0, f_{ v_{25} } = 0, f_{ v_{28} } = 0, f_{ v_{37} } = 0.$\\
\cmidrule{1-4}
\multirow{3}[1]{*}{$3$} & \multirow{3}[1]{*}{$l_{ desk }$} & \multirow{3}[1]{*}{$\leftarrow$} & $f_{ v_{4} } > 1, f_{ v_{2} } \le 1, f_{ v_{3} } \le 1, f_{ v_{5} } \le 2, f_{ v_{9} } \le 1, f_{ v_{11} } \le 1, f_{ v_{13} } \le 1, f_{ v_{14} } \le 1,$\\
& & & $f_{ v_{15} } \le 2, f_{ v_{16} } \le 1, f_{ v_{17} } \le 3, f_{ v_{18} } \le 1, f_{ v_{20} } \le 2, f_{ v_{22} } \le 2, f_{ v_{24} } \le 1, f_{ v_{28} } \le 1,$\\
& & & $f_{ v_{30} } \le 1, f_{ v_{36} } \le 2, f_{ v_{8} } = 0, f_{ v_{10} } = 0, f_{ v_{23} } = 0, f_{ v_{25} } = 0, f_{ v_{29} } = 0, f_{ v_{37} } = 0.$\\
\bottomrule
\end{tabular}
%\end{sc}
\end{small}
\end{center}
\vskip -0.1in
\end{table}

\clearpage

\section{Notation}

\begin{table}[ht]
\caption{Notation}
\label{tab:notation}
\vskip 0.15in
\begin{center}
  \begin{small}
    % \begin{sc}
    \begin{tabular}{ll}
      \toprule
      Symbol & Description \\
      \midrule
      $\mathcal{M}(A)$ & set of multisets of elements from $A$ \\
      $\TopBag$ & top-bag \\
      $\SubBag{j}$ & sub-bag \\
      $\Instance{j}{\ell}$ & is a instance \\
      $\CardinalityTopBag$ & cardinality of top-bag $\TopBag$ \\
      $\CardinalitySubBag{j}$ & cardinality of sub-bag $\SubBag{j}$ \\
      $\InstanceSpace$ & instance space \\
      $\SetOfInstanceLabels$ & instance label space \\
      $\SetOfSubBagLabels$ & sub-bag label space \\
      $\SetOfTopBagLabels$ & the top-bag label space \\
      $y_{j,\ell}$ & label associated with  $\Instance{j}{\ell}$ \\
      $y_j$ & label associated with $\SubBag{j}$. \\
      $y$ & label associated with $\TopBag$  \\
      $g$ & is the bag-layer function \\
      $F$ & is the MMIL network \\
      $\ApproxSetOfInstanceLabels$ & approximated instance label space \\
      $\ApproxSetOfSubBagLabels$ & approximated sub-bag label space \\
      $\CardinalityApproxInstanceLabels$ & cardinality of $\ApproxSetOfInstanceLabels$  \\
      $\CardinalityApproxSubBagLabels$ & is the cardinality of $\ApproxSetOfSubBagLabels$ \\
      $\InstanceReprSet$ & multi-set of the instance representations \\
      $\SubBagReprSet$ & multi-set of the sub-bag representations \\
      $\SurrogateIntanceLabelPredictor$ &  function mapping elements in $\InstanceSpace$ to elements in $\ApproxSetOfInstanceLabels$ \\
      $\DecisionTreeInstToSub$ & function mapping elements in $\ApproxSetOfInstanceLabels$ to elements in  $\ApproxSetOfSubBagLabels$\\
      $\DecisionTreeSubToTop$ & function mapping elements in $\ApproxSetOfSubBagLabels$ to elements in  $\SetOfInstanceLabels$\\
      $\phi_j$ & feature representation corresponding to $\SubBag{j}$   \\
      $\phi$ & feature representation corresponding to $\TopBag$   \\
      $\rho_i$ & activation of the $i$-th node of the bag-layer before performing the aggregation\\

      \bottomrule
    \end{tabular}
    % \end{sc}
  \end{small}
\end{center}
\vskip -0.1in
\end{table}

% BibTeX users please use one of
%\bibliographystyle{spbasic}      % basic style, author-year citations
% \bibliographystyle{spmpsci}      % mathematics and physical sciences
%\bibliographystyle{spphys}       % APS-like style for physics
\bibliographystyle{plain}       % APS-like style for physics
\bibliography{paper.bbl}

\begin{thebibliography}{78}
\providecommand{\natexlab}[1]{#1}
\providecommand{\url}[1]{\texttt{#1}}
\expandafter\ifx\csname urlstyle\endcsname\relax
  \providecommand{\doi}[1]{doi: #1}\else
  \providecommand{\doi}{doi: \begingroup \urlstyle{rm}\Url}\fi

\bibitem[geo()]{geoflora}
Flora von deutschland (phanerogamen).
\newblock URL \url{https://doi.org/10.15468/0fxsox}.
\newblock GBIF Occurrence Download https://doi.org/10.15468/dl.gj34x1.

\bibitem[Andrews et~al.(2002)Andrews, Tsochantaridis, and
  Hofmann]{andrews_support_2002}
Stuart Andrews, Ioannis Tsochantaridis, and Thomas Hofmann.
\newblock Support vector machines for multiple-instance learning.
\newblock In \emph{Advances in neural information processing systems}, pages
  561--568, 2002.

\bibitem[Arbelaez et~al.(2011)Arbelaez, Maire, Fowlkes, and
  Malik]{arbelaez_contour_2011}
Pablo Arbelaez, Michael Maire, Charless Fowlkes, and Jitendra Malik.
\newblock Contour {Detection} and {Hierarchical} {Image} {Segmentation}.
\newblock \emph{IEEE Transactions on Pattern Analysis and Machine
  Intelligence}, 33\penalty0 (5):\penalty0 898--916, May 2011.
\newblock \doi{10.1109/TPAMI.2010.161}.

\bibitem[Arora et~al.(2016)Arora, Liang, and Ma]{arora2016simple}
Sanjeev Arora, Yingyu Liang, and Tengyu Ma.
\newblock A simple but tough-to-beat baseline for sentence embeddings.
\newblock In \emph{International Conference on Learning Representations
  (ICLR)}, 2016.

\bibitem[Atwood and Towsley(2016)]{atwood_diffusion-convolutional_2016}
James Atwood and Don Towsley.
\newblock Diffusion-convolutional neural networks.
\newblock In \emph{Advances in {Neural} {Information} {Processing} {Systems}},
  pages 1993--2001, 2016.

\bibitem[Bach et~al.(2015)Bach, Binder, Montavon, Klauschen, M{\"u}ller, and
  Samek]{bach2015pixel}
Sebastian Bach, Alexander Binder, Gr{\'e}goire Montavon, Frederick Klauschen,
  Klaus-Robert M{\"u}ller, and Wojciech Samek.
\newblock On pixel-wise explanations for non-linear classifier decisions by
  layer-wise relevance propagation.
\newblock \emph{PloS one}, 10\penalty0 (7), 2015.

\bibitem[Blei et~al.(2003)Blei, Ng, and Jordan]{blei_latent_2003}
David~M. Blei, Andrew~Y. Ng, and Michael~I. Jordan.
\newblock Latent dirichlet allocation.
\newblock \emph{Journal of Machine Learning Research}, 3:\penalty0 993--1022,
  2003.
\newblock URL \url{http://dl.acm.org/citation.cfm?id=944937}.

\bibitem[Costa and De~Grave(2010)]{costa_fast_2010}
Fabrizio Costa and Kurt De~Grave.
\newblock Fast neighborhood subgraph pairwise distance kernel.
\newblock In \emph{Proceedings of the 26th {International} {Conference} on
  {Machine} {Learning}}, pages 255--262. Omnipress, 2010.

\bibitem[De~Raedt et~al.(2008{\natexlab{a}})De~Raedt, Demoen, Fierens, Gutmann,
  Janssens, Kimmig, Landwehr, Mantadelis, Meert, Rocha,
  et~al.]{De-Raedt:2008:Towards-digesting-the-alphabet-soup}
L.~De~Raedt, B.~Demoen, D.~Fierens, B.~Gutmann, G.~Janssens, A.~Kimmig,
  N.~Landwehr, T.~Mantadelis, W.~Meert, R.~Rocha, et~al.
\newblock {Towards digesting the alphabet-soup of statistical relational
  learning}, 2008{\natexlab{a}}.

\bibitem[De~Raedt et~al.(2008{\natexlab{b}})De~Raedt, Frasconi, Kersting, and
  Muggleton]{2008:Probabilistic-inductive-logic}
Luc De~Raedt, Paolo Frasconi, Kristian Kersting, and Stephen Muggleton.
\newblock Probabilistic inductive logic programming: theory and applications,
  2008{\natexlab{b}}.

\bibitem[Dietterich(2000)]{dietterich_ensemble_2000}
Thomas~G. Dietterich.
\newblock Ensemble {Methods} in {Machine} {Learning}.
\newblock In \emph{Multiple {Classifier} {Systems}}, number 1857 in Lecture
  {Notes} in {Computer} {Science}, pages 1--15. Springer Berlin Heidelberg,
  June 2000.
\newblock ISBN 978-3-540-67704-8 978-3-540-45014-6.

\bibitem[Dietterich et~al.(1997)Dietterich, Lathrop, and
  Lozano-P\'{e}rez]{dietterich_solving_1997}
Thomas~G. Dietterich, Richard~H. Lathrop, and Tom\'{a}s Lozano-P\'{e}rez.
\newblock Solving the multiple instance problem with axis-parallel rectangles.
\newblock \emph{Artificial Intelligence}, 89\penalty0 (1–2):\penalty0 31--71,
  January 1997.
\newblock \doi{10.1016/S0004-3702(96)00034-3}.

\bibitem[Duvenaud et~al.(2015)Duvenaud, Maclaurin, Iparraguirre, Bombarell,
  Hirzel, Aspuru-Guzik, and Adams]{duvenaud_convolutional_2015}
David~K Duvenaud, Dougal Maclaurin, Jorge Iparraguirre, Rafael Bombarell,
  Timothy Hirzel, Al{\'a}n Aspuru-Guzik, and Ryan~P Adams.
\newblock Convolutional networks on graphs for learning molecular fingerprints.
\newblock In \emph{Advances in neural information processing systems}, pages
  2224--2232, 2015.

\bibitem[Foulds and Frank(2010)]{foulds_review_2010}
James Foulds and Eibe Frank.
\newblock A review of multi-instance learning assumptions.
\newblock \emph{The Knowledge Engineering Review}, 25\penalty0 (01):\penalty0
  1, March 2010.
\newblock \doi{10.1017/S026988890999035X}.

\bibitem[Frasconi et~al.(1998)Frasconi, Gori, and Sperduti]{Frasconi98}
P.~Frasconi, M.~Gori, and A.~Sperduti.
\newblock A general framework for adaptive processing of data structures.
\newblock \emph{IEEE Trans. on Neural Networks}, 9:\penalty0 768--786, 1998.

\bibitem[Fukushima(1980)]{fukushima_neocognitron:_1980}
Kunihiko Fukushima.
\newblock Neocognitron: {A} self-organizing neural network model for a
  mechanism of pattern recognition unaffected by shift in position.
\newblock \emph{Biological cybernetics}, 36\penalty0 (4):\penalty0 193--202,
  1980.

\bibitem[Gardner et~al.(2018)Gardner, Pleiss, Bindel, Weinberger, and
  Wilson]{gardner2018gpytorch}
Jacob~R Gardner, Geoff Pleiss, David Bindel, Kilian~Q Weinberger, and
  Andrew~Gordon Wilson.
\newblock Gpytorch: Blackbox matrix-matrix gaussian process inference with gpu
  acceleration.
\newblock In \emph{Advances in Neural Information Processing Systems}, 2018.

\bibitem[G{\"a}rtner et~al.(2004)G{\"a}rtner, Lloyd, and
  Flach]{gartner2004kernels}
Thomas G{\"a}rtner, John~W Lloyd, and Peter~A Flach.
\newblock Kernels and distances for structured data.
\newblock \emph{Machine Learning}, 57\penalty0 (3):\penalty0 205--232, 2004.

\bibitem[Getoor and
  Taskar(2007)]{Getoor:2007:Introduction-to-statistical-relational}
Lise Getoor and Ben Taskar.
\newblock \emph{Introduction to statistical relational learning}.
\newblock MIT Press, Cambridge, Mass., 2007.

\bibitem[Gilmer et~al.(2017)Gilmer, Schoenholz, Riley, Vinyals, and
  Dahl]{DBLP:conf/icml/GilmerSRVD17}
Justin Gilmer, Samuel~S. Schoenholz, Patrick~F. Riley, Oriol Vinyals, and
  George~E. Dahl.
\newblock Neural message passing for quantum chemistry.
\newblock In \emph{Proceedings of the 34th International Conference on Machine
  Learning, {ICML} 2017}, pages 1263--1272, 2017.
\newblock URL \url{http://proceedings.mlr.press/v70/gilmer17a.html}.

\bibitem[Gori et~al.(2005)Gori, Monfardini, and Scarselli]{gori2005new}
Marco Gori, Gabriele Monfardini, and Franco Scarselli.
\newblock A new model for learning in graph domains.
\newblock In \emph{Neural Networks, 2005. IJCNN'05. Proceedings. 2005 IEEE
  International Joint Conference on}, volume~2, pages 729--734. IEEE, 2005.

\bibitem[Griffiths and Steyvers(2004)]{griffiths_2004_finding}
Thomas~L Griffiths and Mark Steyvers.
\newblock Finding scientific topics.
\newblock \emph{Proceedings of the National academy of Sciences}, 101\penalty0
  (suppl 1):\penalty0 5228--5235, 2004.

\bibitem[Hamilton et~al.(2017)Hamilton, Ying, and
  Leskovec]{hamilton2017inductive}
Will Hamilton, Zhitao Ying, and Jure Leskovec.
\newblock Inductive representation learning on large graphs.
\newblock In \emph{Advances in Neural Information Processing Systems}, pages
  1024--1034, 2017.

\bibitem[Haussler(1999)]{haussler_convolution_1999}
David Haussler.
\newblock Convolution kernels on discrete structures.
\newblock Technical Report 646, Department of Computer Science, University of
  California at Santa Cruz, 1999.

\bibitem[Hensman et~al.(2015)Hensman, Matthews, and
  Ghahramani]{hensman2015scalable}
James Hensman, Alexander Matthews, and Zoubin Ghahramani.
\newblock Scalable variational gaussian process classification.
\newblock 2015.

\bibitem[Hentschel and Sack(2015)]{hentschel2015image}
Christian Hentschel and Harald Sack.
\newblock What image classifiers really see--visualizing bag-of-visual words
  models.
\newblock In \emph{International Conference on Multimedia Modeling}, pages
  95--104. Springer, 2015.

\bibitem[Horv\'{a}th et~al.(2004)Horv\'{a}th, G\"{a}rtner, and
  Wrobel]{horvath_cyclic_2004}
Tam\'{a}s Horv\'{a}th, Thomas G\"{a}rtner, and Stefan Wrobel.
\newblock Cyclic pattern kernels for predictive graph mining.
\newblock In \emph{Proceedings of the tenth {ACM} {SIGKDD} international
  conference on {Knowledge} discovery and data mining}, pages 158--167. ACM,
  2004.

\bibitem[Hou et~al.(2015)Hou, Samaras, Kurc, Gao, Davis, and
  Saltz]{hou2015efficient}
Le~Hou, Dimitris Samaras, Tahsin~M Kurc, Yi~Gao, James~E Davis, and Joel~H
  Saltz.
\newblock Efficient multiple instance convolutional neural networks for
  gigapixel resolution image classification.
\newblock \emph{arXiv preprint arXiv:1504.07947}, 2015.

\bibitem[Jaeger(1997)]{jaeger_relational_1997}
M.~Jaeger.
\newblock Relational bayesian networks.
\newblock In Dan Geiger and Prakash~Pundalik Shenoy, editors, \emph{Proceedings
  of the 13th Conference of Uncertainty in Artificial Intelligence (UAI-13)},
  pages 266--273, Providence, USA, 1997. Morgan Kaufmann.

\bibitem[Kingma and Ba(2015)]{kingma2014adam}
Diederik Kingma and Jimmy Ba.
\newblock Adam: {A} {Method} for {Stochastic} {Optimization}.
\newblock In \emph{3rd International Conference for Learning Representations},
  San Diego, CA, 2015.
\newblock arXiv:1412.6980.

\bibitem[Kipf and Welling(2016)]{kipfsemi-supervised2016}
Thomas~N. Kipf and Max Welling.
\newblock Semi-supervised classification with graph convolutional networks.
\newblock \emph{arXiv preprint arXiv:1609.02907}, 2016.

\bibitem[Kondor and Jebara(2003)]{DBLP:conf/icml/KondorJ03}
Risi Kondor and Tony Jebara.
\newblock A kernel between sets of vectors.
\newblock In \emph{Machine Learning, Proceedings of the Twentieth International
  Conference {(ICML} 2003), August 21-24, 2003, Washington, DC, {USA}}, pages
  361--368, 2003.
\newblock URL \url{http://www.aaai.org/Library/ICML/2003/icml03-049.php}.

\bibitem[Landwehr et~al.(2010)Landwehr, Passerini, {De Raedt}, and
  Frasconi]{Landwehr:2010:Fast-learning-of-relational}
N.~Landwehr, A.~Passerini, L.~{De Raedt}, and P.~Frasconi.
\newblock {Fast learning of relational kernels}.
\newblock \emph{Machine learning}, 78\penalty0 (3):\penalty0 305--342, 2010.

\bibitem[Lapuschkin et~al.(2016)Lapuschkin, Binder, Montavon, M{\"u}ller, and
  Samek]{lapuschkin2016lrp}
Sebastian Lapuschkin, Alexander Binder, Gr{\'e}goire Montavon, Klaus-Robert
  M{\"u}ller, and Wojciech Samek.
\newblock The lrp toolbox for artificial neural networks.
\newblock \emph{The Journal of Machine Learning Research}, 17\penalty0
  (1):\penalty0 3938--3942, 2016.

\bibitem[LeCun et~al.(1989)LeCun, Boser, Denker, Henderson, Howard, Hubbard,
  and Jackel]{lecun_backpropagation_1989}
Yann LeCun, Bernhard Boser, John~S. Denker, Donnie Henderson, Richard~E.
  Howard, Wayne Hubbard, and Lawrence~D. Jackel.
\newblock Backpropagation applied to handwritten zip code recognition.
\newblock \emph{Neural computation}, 1\penalty0 (4):\penalty0 541--551, 1989.

\bibitem[Lusci et~al.(2013)Lusci, Pollastri, and Baldi]{Lusci_2013}
Alessandro Lusci, Gianluca Pollastri, and Pierre Baldi.
\newblock Deep architectures and deep learning in chemoinformatics: The
  prediction of aqueous solubility for drug-like molecules.
\newblock \emph{Journal of Chemical Information and Modeling}, 53\penalty0
  (7):\penalty0 1563–1575, Jul 2013.
\newblock ISSN 1549-960X.
\newblock \doi{10.1021/ci400187y}.
\newblock URL \url{http://dx.doi.org/10.1021/ci400187y}.

\bibitem[Maas et~al.(2011)Maas, Daly, Pham, Huang, Ng, and
  Potts]{maas-EtAl:2011:ACL-HLT2011}
Andrew~L. Maas, Raymond~E. Daly, Peter~T. Pham, Dan Huang, Andrew~Y. Ng, and
  Christopher Potts.
\newblock Learning word vectors for sentiment analysis.
\newblock In \emph{Proceedings of the 49th Annual Meeting of the Association
  for Computational Linguistics: Human Language Technologies}, pages 142--150,
  Portland, Oregon, USA, June 2011. Association for Computational Linguistics.

\bibitem[Maron and Lozano-Pérez(1998)]{maron_framework_1998}
Oded Maron and Tomás Lozano-Pérez.
\newblock A framework for multiple-instance learning.
\newblock \emph{Advances in neural information processing systems}, pages
  570--576, 1998.

\bibitem[Maron and Ratan(1998)]{maron1998multiple}
Oded Maron and Aparna~Lakshmi Ratan.
\newblock Multiple-instance learning for natural scene classification.
\newblock In \emph{ICML}, volume~98, pages 341--349, 1998.

\bibitem[Minsky and Papert(1988)]{Minsky_1988}
Marvin Minsky and Seymour~A. Papert.
\newblock \emph{Perceptrons, expanded edition}.
\newblock The MIT Press, 1988.

\bibitem[Miyato et~al.(2016)Miyato, Dai, and Goodfellow]{miyato2016virtual}
Takeru Miyato, Andrew~M Dai, and Ian Goodfellow.
\newblock Virtual adversarial training for semi-supervised text classification.
\newblock In \emph{International Conference on Learning Representations
  (ICLR)}, 2016.

\bibitem[Natarajan et~al.(2008)Natarajan, Tadepalli, Dietterich, and
  Fern]{natarajan_learning_2008}
Sriraam Natarajan, Prasad Tadepalli, Thomas~G. Dietterich, and Alan Fern.
\newblock Learning first-order probabilistic models with combining rules.
\newblock \emph{Annals of Mathematics and Artificial Intelligence}, 54\penalty0
  (1-3):\penalty0 223--256, 2008.
\newblock URL \url{http://link.springer.com/article/10.1007/s10472-009-9138-5}.

\bibitem[Neumann et~al.(2012)Neumann, Patricia, Garnett, and
  Kersting]{neumann_efficient_2012}
Marion Neumann, Novi Patricia, Roman Garnett, and Kristian Kersting.
\newblock Efficient graph kernels by randomization.
\newblock In \emph{Joint European Conference on Machine Learning and Knowledge
  Discovery in Databases}, pages 378--393. Springer, 2012.
\newblock URL
  \url{http://link.springer.com/chapter/10.1007/978-3-642-33460-3_30}.

\bibitem[Niepert et~al.(2016)Niepert, Ahmed, and
  Kutzkov]{niepert_learning_2016}
Mathias Niepert, Mohamed Ahmed, and Konstantin Kutzkov.
\newblock Learning {Convolutional} {Neural} {Networks} for {Graphs}.
\newblock In \emph{International conference on machine learning}, pages
  2014--2023, 2016.

\bibitem[Oliver and Webster(1990)]{oliver1990kriging}
Margaret~A Oliver and Richard Webster.
\newblock Kriging: a method of interpolation for geographical information
  systems.
\newblock \emph{International Journal of Geographical Information System},
  4\penalty0 (3):\penalty0 313--332, 1990.

\bibitem[Orsini et~al.(2015)Orsini, Frasconi, and De~Raedt]{orsini2015graph}
Francesco Orsini, Paolo Frasconi, and Luc De~Raedt.
\newblock Graph invariant kernels.
\newblock In \emph{Proceedings of the Twenty-fourth International Joint
  Conference on Artificial Intelligence}, pages 3756--3762, 2015.

\bibitem[Orsini et~al.(2018)Orsini, Baracchi, and Frasconi]{orsini2018shift}
Francesco Orsini, Daniele Baracchi, and Paolo Frasconi.
\newblock Shift aggregate extract networks.
\newblock \emph{Frontiers in Robotics and AI}, 5:\penalty0 42, 2018.

\bibitem[Passerini et~al.(2006)Passerini, Frasconi, and
  Raedt]{passerini2006kernels}
Andrea Passerini, Paolo Frasconi, and Luc~De Raedt.
\newblock Kernels on prolog proof trees: Statistical learning in the ilp
  setting.
\newblock \emph{Journal of Machine Learning Research}, 7\penalty0
  (Feb):\penalty0 307--342, 2006.

\bibitem[Pennington et~al.(2014)Pennington, Socher, and
  Manning]{pennington2014glove}
Jeffrey Pennington, Richard Socher, and Christopher~D. Manning.
\newblock Glove: Global vectors for word representation.
\newblock In \emph{Empirical Methods in Natural Language Processing (EMNLP)},
  pages 1532--1543, 2014.

\bibitem[Rahmani et~al.(2005)Rahmani, Goldman, Zhang, Krettek, and
  Fritts]{rahmani2005localized}
Rouhollah Rahmani, Sally~A Goldman, Hui Zhang, John Krettek, and Jason~E
  Fritts.
\newblock Localized content based image retrieval.
\newblock In \emph{Proceedings of the 7th ACM SIGMM international workshop on
  Multimedia information retrieval}, pages 227--236. ACM, 2005.

\bibitem[Ramon and De~Raedt(2000)]{ramon_multi_2000}
Jan Ramon and Luc De~Raedt.
\newblock Multi instance neural networks.
\newblock In \emph{Proceedings of the ICML-2000 workshop on attribute-value and
  relational learning}, 2000.

\bibitem[Ribeiro et~al.(2016)Ribeiro, Singh, and Guestrin]{ribeiro2016should}
Marco~Tulio Ribeiro, Sameer Singh, and Carlos Guestrin.
\newblock Why should i trust you?: Explaining the predictions of any
  classifier.
\newblock In \emph{Proceedings of the 22nd ACM SIGKDD international conference
  on knowledge discovery and data mining}, pages 1135--1144. ACM, 2016.

\bibitem[Richardson and Domingos(2006)]{Richardson:2006:Markov-logic-networks}
Matthew Richardson and Pedro Domingos.
\newblock Markov logic networks.
\newblock \emph{Machine Learning}, 62:\penalty0 107--136, 2006.

\bibitem[Rusu and Cousins(2011)]{rusu20113d}
Radu~Bogdan Rusu and Steve Cousins.
\newblock 3d is here: Point cloud library (pcl).
\newblock In \emph{2011 IEEE international conference on robotics and
  automation}, pages 1--4. IEEE, 2011.

\bibitem[Samek et~al.(2016)Samek, Montavon, Binder, Lapuschkin, and
  M{\"u}ller]{samek2016interpreting}
Wojciech Samek, Gr{\'e}goire Montavon, Alexander Binder, Sebastian Lapuschkin,
  and Klaus-Robert M{\"u}ller.
\newblock Interpreting the predictions of complex ml models by layer-wise
  relevance propagation.
\newblock In \emph{NIPS 2016 Workshop on Interpretable Machine Learning in
  Complex Systems}, 2016.
\newblock arXiv:1611.08191.

\bibitem[Scarselli et~al.(2009)Scarselli, Gori, Tsoi, Hagenbuchner, and
  Monfardini]{scarselli_graph_2009}
Franco Scarselli, Marco Gori, Ah~Chung Tsoi, Markus Hagenbuchner, and Gabriele
  Monfardini.
\newblock The graph neural network model.
\newblock \emph{IEEE Transactions on Neural Networks}, 20\penalty0
  (1):\penalty0 61--80, 2009.

\bibitem[Sen et~al.(2008)Sen, Namata, Bilgic, Getoor, Galligher, and
  Eliassi-Rad]{sen_collective_2008}
Prithviraj Sen, Galileo Namata, Mustafa Bilgic, Lise Getoor, Brian Galligher,
  and Tina Eliassi-Rad.
\newblock Collective classification in network data.
\newblock \emph{AI magazine}, 29\penalty0 (3):\penalty0 93, 2008.

\bibitem[{Shawe-Taylor}(1989)]{Shawe-Taylor-89}
J.~{Shawe-Taylor}.
\newblock Building symmetries into feedforward networks.
\newblock In \emph{First IEEE International Conference on Artificial Neural
  Networks, (Conf. Publ. No. 313)}, pages 158--162, 1989.

\bibitem[Shervashidze et~al.(2009)Shervashidze, Vishwanathan, Petri, Mehlhorn,
  and Borgwardt]{shervashidze_efficient_2009}
Nino Shervashidze, S.~V.~N. Vishwanathan, Tobias Petri, Kurt Mehlhorn, and
  Karsten~M. Borgwardt.
\newblock Efficient graphlet kernels for large graph comparison.
\newblock In \emph{{AISTATS}}, volume~5, pages 488--495, 2009.

\bibitem[Shervashidze et~al.(2011)Shervashidze, Schweitzer, Van~Leeuwen,
  Mehlhorn, and Borgwardt]{shervashidze_weisfeiler-lehman_2011}
Nino Shervashidze, Pascal Schweitzer, Erik~Jan Van~Leeuwen, Kurt Mehlhorn, and
  Karsten~M. Borgwardt.
\newblock Weisfeiler-lehman graph kernels.
\newblock \emph{The Journal of Machine Learning Research}, 12:\penalty0
  2539--2561, 2011.

\bibitem[Szegedy et~al.(2017)Szegedy, Ioffe, Vanhoucke, and
  Alemi]{szegedy_inception-v4_2016}
Christian Szegedy, Sergey Ioffe, Vincent Vanhoucke, and Alex Alemi.
\newblock Inception-v4, {Inception}-{ResNet} and the {Impact} of {Residual}
  {Connections} on {Learning}.
\newblock In \emph{Proc. of AAAI}, 2017.
\newblock arXiv: 1602.07261.

\bibitem[Tibo et~al.(2017)Tibo, Frasconi, and Jaeger]{tibo2017network}
Alessandro Tibo, Paolo Frasconi, and Manfred Jaeger.
\newblock A network architecture for multi-multi-instance learning.
\newblock In \emph{Joint European Conference on Machine Learning and Knowledge
  Discovery in Databases}, pages 737--752. Springer, 2017.

\bibitem[Uijlings et~al.(2012)Uijlings, Smeulders, and
  Scha]{uijlings2012visual}
Jasper~RR Uijlings, Arnold~WM Smeulders, and Remko~JH Scha.
\newblock The visual extent of an object.
\newblock \emph{International journal of computer vision}, 96\penalty0
  (1):\penalty0 46--63, 2012.

\bibitem[Velickovic et~al.(2018)Velickovic, Cucurull, Casanova, Romero,
  Li{\`{o}}, and Bengio]{DBLP:conf/iclr/VelickovicCCRLB18}
Petar Velickovic, Guillem Cucurull, Arantxa Casanova, Adriana Romero, Pietro
  Li{\`{o}}, and Yoshua Bengio.
\newblock Graph attention networks.
\newblock In \emph{6th International Conference on Learning Representations,
  {ICLR}}, 2018.
\newblock URL \url{https://openreview.net/forum?id=rJXMpikCZ}.

\bibitem[Verstrepen et~al.(2017)Verstrepen, Bhaduriy, Cule, and
  Goethals]{verstrepen2017collaborative}
Koen Verstrepen, Kanishka Bhaduriy, Boris Cule, and Bart Goethals.
\newblock Collaborative filtering for binary, positiveonly data.
\newblock \emph{ACM SIGKDD Explorations Newsletter}, 19\penalty0 (1):\penalty0
  1--21, 2017.

\bibitem[Vinyals et~al.(2016)Vinyals, Bengio, and
  Kudlur]{DBLP:journals/corr/VinyalsBK15}
Oriol Vinyals, Samy Bengio, and Manjunath Kudlur.
\newblock Order matters: Sequence to sequence for sets.
\newblock In \emph{4th International Conference on Learning Representations,
  {ICLR} 2016}, 2016.
\newblock URL \url{http://arxiv.org/abs/1511.06391}.

\bibitem[Wang and Zucker(2000)]{wang_solving_2000}
Jun Wang and Jean-Daniel Zucker.
\newblock Solving multiple-instance problem: {A} lazy learning approach.
\newblock In \emph{Proceedings of the Seventeenth International Conference on
  Machine Learning}, 2000.

\bibitem[Williams and Rasmussen(2006)]{williams2006gaussian}
Christopher~KI Williams and Carl~Edward Rasmussen.
\newblock \emph{Gaussian processes for machine learning}, volume~2.
\newblock MIT press Cambridge, MA, 2006.

\bibitem[Wu et~al.(2015)Wu, Song, Khosla, Yu, Zhang, Tang, and Xiao]{wu20153d}
Zhirong Wu, Shuran Song, Aditya Khosla, Fisher Yu, Linguang Zhang, Xiaoou Tang,
  and Jianxiong Xiao.
\newblock 3d shapenets: A deep representation for volumetric shapes.
\newblock In \emph{Proceedings of the IEEE conference on computer vision and
  pattern recognition}, pages 1912--1920, 2015.

\bibitem[Yan et~al.(2016)Yan, Zhan, Peng, Liao, Shinagawa, Zhang, Metaxas, and
  Zhou]{yan2016multi}
Zhennan Yan, Yiqiang Zhan, Zhigang Peng, Shu Liao, Yoshihisa Shinagawa,
  Shaoting Zhang, Dimitris~N Metaxas, and Xiang~Sean Zhou.
\newblock Multi-instance deep learning: Discover discriminative local anatomies
  for bodypart recognition.
\newblock \emph{IEEE transactions on medical imaging}, 35\penalty0
  (5):\penalty0 1332--1343, 2016.

\bibitem[Yanardag and Vishwanathan(2015)]{yanardag2015deep}
Pinar Yanardag and SVN Vishwanathan.
\newblock Deep graph kernels.
\newblock In \emph{Proceedings of the 21th ACM SIGKDD International Conference
  on Knowledge Discovery and Data Mining}, pages 1365--1374. ACM, 2015.

\bibitem[Yang et~al.(2006)Yang, Dong, and Hua]{yang2006region}
Changbo Yang, Ming Dong, and Jing Hua.
\newblock Region-based image annotation using asymmetrical support vector
  machine-based multiple-instance learning.
\newblock In \emph{Computer Vision and Pattern Recognition, 2006 IEEE Computer
  Society Conference on}, volume~2, pages 2057--2063. IEEE, 2006.

\bibitem[Yang and Lozano-Perez(2000)]{yang2000image}
Cheng Yang and Tomas Lozano-Perez.
\newblock Image database retrieval with multiple-instance learning techniques.
\newblock In \emph{Data Engineering, 2000. Proceedings. 16th International
  Conference on}, pages 233--243. IEEE, 2000.

\bibitem[Zaheer et~al.(2017)Zaheer, Kottur, Ravanbakhsh, Poczos, Salakhutdinov,
  and Smola]{zaheer2017deep}
Manzil Zaheer, Satwik Kottur, Siamak Ravanbakhsh, Barnabas Poczos, Russ~R
  Salakhutdinov, and Alexander~J Smola.
\newblock Deep sets.
\newblock In \emph{Advances in neural information processing systems}, pages
  3391--3401, 2017.

\bibitem[Zha et~al.(2008)Zha, Hua, Mei, Wang, Qi, and Wang]{zha2008joint}
Zheng-Jun Zha, Xian-Sheng Hua, Tao Mei, Jingdong Wang, Guo-Jun Qi, and Zengfu
  Wang.
\newblock Joint multi-label multi-instance learning for image classification.
\newblock In \emph{Computer Vision and Pattern Recognition, 2008. CVPR 2008.
  IEEE Conference on}, pages 1--8. IEEE, 2008.

\bibitem[Zhou et~al.(2008)Zhou, Wilkinson, Schreiber, and Pan]{zhou2008large}
Yunhong Zhou, Dennis Wilkinson, Robert Schreiber, and Rong Pan.
\newblock Large-scale parallel collaborative filtering for the netflix prize.
\newblock In \emph{International conference on algorithmic applications in
  management}, pages 337--348. Springer, 2008.

\bibitem[Zhou et~al.(2005)Zhou, Jiang, and Li]{zhou2005multi}
Zhi-Hua Zhou, Kai Jiang, and Ming Li.
\newblock Multi-instance learning based {W}eb mining.
\newblock \emph{Applied Intelligence}, 22\penalty0 (2):\penalty0 135--147,
  2005.

\bibitem[Zhou et~al.(2012)Zhou, Zhang, Huang, and Li]{zhou_multi-instance_2012}
Zhi-Hua Zhou, Min-Ling Zhang, Sheng-Jun Huang, and Yu-Feng Li.
\newblock Multi-instance multi-label learning.
\newblock \emph{Artificial Intelligence}, 176\penalty0 (1):\penalty0
  2291--2320, January 2012.
\newblock \doi{10.1016/j.artint.2011.10.002}.

\end{thebibliography}
\end{document}